\documentclass[10pt,journal,compsoc]{IEEEtran}
\ifCLASSOPTIONcompsoc
  \usepackage[nocompress]{cite}
\else
  \usepackage{cite}
\fi
\ifCLASSINFOpdf
\else
\fi

\usepackage{url}            
\usepackage{nicefrac}       
\usepackage{microtype}      
\usepackage{floatrow}
\newfloatcommand{capbtabbox}{table}[][\FBwidth]
\usepackage{xcolor}

\usepackage[utf8]{inputenc} 
\usepackage[T1]{fontenc}    
\usepackage{url}            
\usepackage{amsfonts}       
\usepackage{nicefrac}       
\usepackage{microtype}      
\usepackage{xcolor}         

\usepackage{algorithm}
\usepackage{algpseudocode}

\usepackage{graphicx}
\graphicspath{{./}{../}}
\usepackage{booktabs}
\usepackage{times}
\usepackage{epsfig}
\usepackage{amsmath}
\usepackage{amssymb}
\usepackage{enumitem}
\usepackage[font=small]{caption}
\usepackage{subcaption}
\usepackage{array}
\usepackage[bookmarks=false]{hyperref}  
\hypersetup{                           
    colorlinks = true,                 
    citecolor  = blue,                 
    linkcolor  = blue,                 
    urlcolor   = blue,                 
} 
\usepackage{cleveref}
\Crefformat{figure}{Fig.~#2#1#3}                           
\Crefname{subfigure}{Fig.}{Figs.}
\Crefname{figure}{Fig.}{Figs.}
\Crefformat{table}{TABLE~#2#1#3}                           
\usepackage[skip=5pt]{caption}            
\renewcommand{\vec}[1]{\boldsymbol{#1}}
\renewcommand{\L}{\vec{L}}
\newcommand{\R}{\vec{R}}
\newcommand{\I}{\vec{I}}
\newcommand{\minisection}[1]{\vspace{.1in}\noindent{\textbf{#1}}.}

\begin{document}

\title{
    Low-Light Video Enhancement with An Effective Spatial-Temporal Decomposition Paradigm
}

\author{
    Xiaogang Xu, Kun Zhou, Tao Hu, Jiafei Wu, Ruixing Wang, Hao Peng and Bei Yu 
    \thanks{Xiaogang Xu and Bei Yu are with the Department of Computer Science and Engineering, The Chinese University of Hong Kong, E-mail: xiaogangxu00@gmail.com, byu@cse.cuhk.edu.hk}
    \thanks{Kun Zhou is with the Shenzhen University, E-mail:  kunzhou@link.cuhk.edu.cn}
    \thanks{Tao Hu is with Pico MR division Bytedance, E-mail: yihouxiang@gmail.com}
    \thanks{Jiafei Wu is with The University of Hong Kong, E-mail: jcjiafeiwu@gmail.com}
    \thanks{Ruixing Wang is with the camera group of DJI, E-mail: ruixingw@hustunique.com}
    \thanks{Hao Peng is with Zhejiang Normal University, E-mail: hpeng@zjun.edu.cn}
}

\markboth{submission to IEEE Transactions on Pattern Analysis and Machine Intelligence}
{Shell \MakeLowercase{\textit{et al.}}: Bare Demo of IEEEtran.cls for Computer Society Journals}

\IEEEtitleabstractindextext{%
\begin{abstract}
Low-Light Video Enhancement (LLVE) seeks to restore dynamic or static scenes plagued by severe invisibility and noise. In this paper, we present an innovative video decomposition strategy that incorporates view-independent and view-dependent components to enhance the performance of LLVE. The framework is called View-aware Low-light Video Enhancement (VLLVE). We leverage dynamic cross-frame correspondences for the view-independent term (which primarily captures intrinsic appearance) and impose a scene-level continuity constraint on the view-dependent term (which mainly describes the shading condition) to achieve consistent and satisfactory decomposition results. To further ensure consistent decomposition, we introduce a dual-structure enhancement network featuring a cross-frame interaction mechanism. By supervising different frames simultaneously, this network encourages them to exhibit matching decomposition features. This mechanism can seamlessly integrate with encoder-decoder single-frame networks, incurring minimal additional parameter costs.  Building upon VLLVE, we propose a more comprehensive decomposition strategy by introducing an additive residual term, resulting in VLLVE++. This residual term can simulate scene-adaptive degradations, which are difficult to model using a decomposition formulation for common scenes, thereby further enhancing the ability to capture the overall content of videos. In addition, VLLVE++ enables bidirectional learning for both enhancement and degradation-aware correspondence refinement (end-to-end manner), effectively increasing reliable correspondences while filtering out incorrect ones. This leads to significantly improved performance. Notably, VLLVE++ demonstrates strong capability in handling challenging cases, such as real-world scenes and videos with high dynamics.
Extensive experiments are conducted on widely recognized LLVE benchmarks, covering diverse scenarios. Our framework of VLLVE and VLLVE++ consistently outperforms existing methods, establishing a new state-of-the-art (SOTA) performance.
\end{abstract}

}

\maketitle
\thispagestyle{plain}
\pagestyle{plain}

\IEEEdisplaynontitleabstractindextext

%
\IEEEpeerreviewmaketitle

\IEEEraisesectionheading{\section{Introduction}\label{sec:intro}}


\IEEEPARstart{L}{ow-light} enhancement aims to enhance underexposed images and videos captured in low-light conditions~\cite{xu2022snr,wang2021sdsd}, improving their visual quality. This is a longstanding and important topic in computer vision, with wide-ranging applications, such as improving portrait photography on mobile devices~\cite{ignatov2017dslr,hasinoff2016burst} and supporting various downstream tasks, including nighttime face recognition~\cite{ma2022toward,wang2022unsupervised} and vehicle detection~\cite{fu2023dancing,wu2022edge}.

Numerous methods have been proposed to enhance underexposed images using supervised learning~\cite{wang2021real,xu2022snr,xu2023low,cai2023retinexformer,wang2022low}, yielding impressive results. However, these methods still struggle to provide satisfactory performance when applied to videos.
The key challenge when enhancing videos is the need for consistent enhancement results across corresponding locations in different frames.
Sufficient ground truth can help to ensure this consistency.
Unfortunately, there is a scarcity of real-world, high-quality, spatially aligned video pairs for dynamic scenes, resulting in limited generalization capabilities for unseen videos.

\begin{figure}[tb!]
    \centering
    \includegraphics[width=1.0\linewidth]{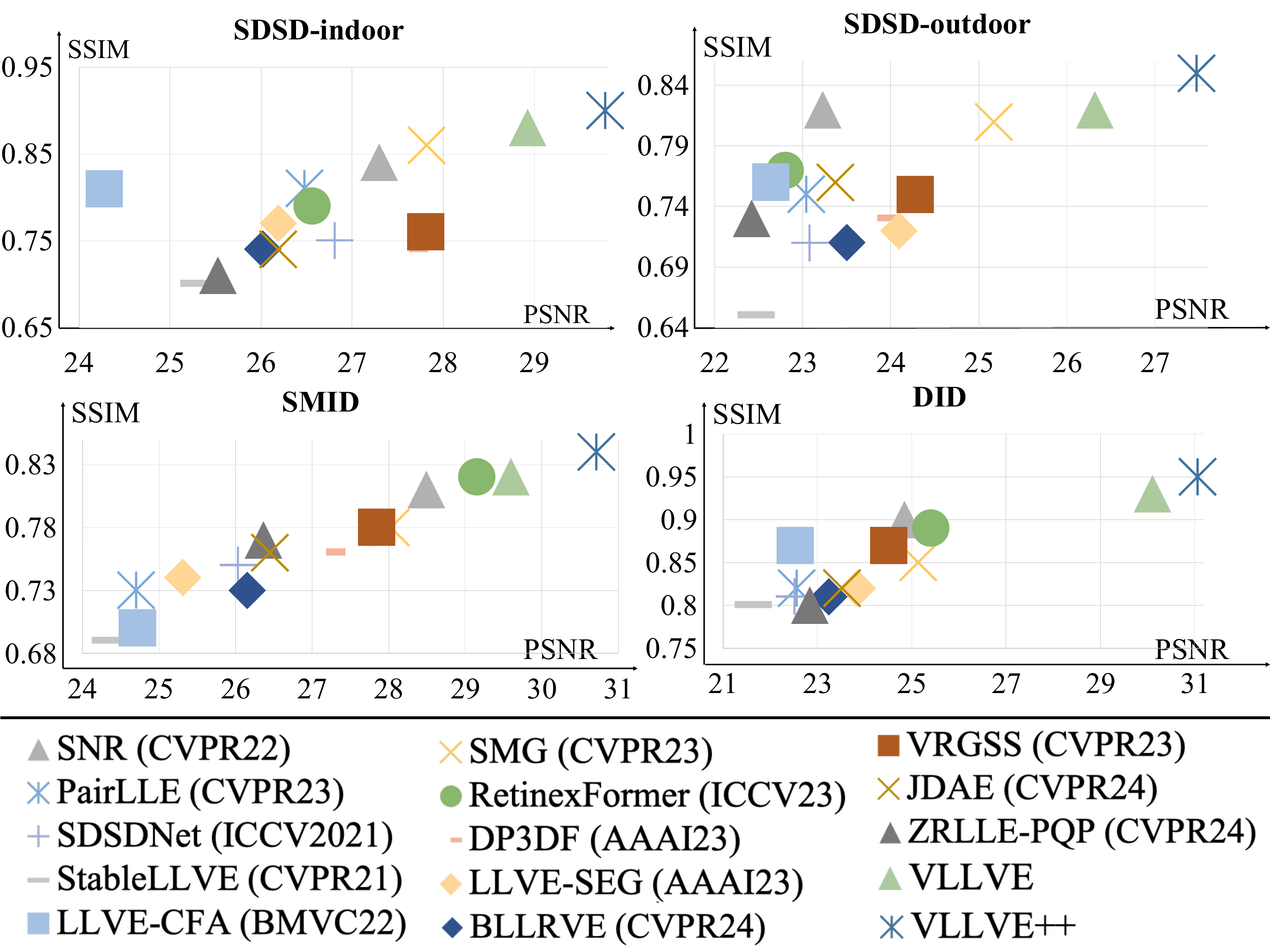}
    \vspace{-0.1in}
    \caption{
        Our proposed LLVE method {\em consistently\/} achieves SOTA performance on {\em different LLVE datasets involving various scenes\/} with the same network architecture. 
    }
    \label{fig:teaser-com}
    \vspace{-0.2in}
\end{figure}

One solution is to incorporate parameterized enhancement models with physical significance into the enhancement process, reducing the reliance on training data.
Towards high-quality enhancement, we propose View-Aware Low-Light Video Enhancement (VLLVE), which is inspired by the linear Lambertian model for intrinsic image decomposition.
According to the linear Lambertian model for intrinsic image decomposition, an observed image $\I$ can be expressed as the element-wise product ($\otimes$) of its albedo component $\boldsymbol{A}$ and shading component $\boldsymbol{S}$, formulated as $\I = \boldsymbol{A} \otimes \boldsymbol{S}$.
$\boldsymbol{A}$ is normally view-independent, capturing the intrinsic appearance, whereas $\boldsymbol{S}$ is commonly view-dependent, influenced by lighting conditions and the surface normal's direction.
Inspired by the Lambertian model, we propose that an image under normal illumination in the LLVE task can similarly be decomposed as $\I = \L \otimes \R$, where $\L$ represents the view-dependent term, and $\R$ corresponds to the view-independent component.
To implement the decomposition in VLLVE, there are two primary components: (1) we leverage cross-frame correspondences and real-world physical continuity constraints to achieve this decomposition without the need for explicit supervision. (2) Furthermore, to enhance the consistency of the decomposition, we develop an efficient dual-structure enhancement network featuring a novel cross-frame interaction mechanism, named as Cross-Frame Interaction Module (CFIM).
Our proposed decomposition strategy and CFIM can be seamlessly integrated into any encoder-decoder single-image enhancement architecture (i.e., backbone) by inserting CFIM at the deepest hidden layer and applying supervision for decomposition at the pixel level. 
Consequently, VLLVE shows SOTA performance on various widely recognized video enhancement datasets~\cite{wang2021sdsd,chen2019seeing,fu2023dancing,zhang2021learning}, as well as a large-scale user study.

However, two key aspects warrant further investigation: (1) multiplication-based decomposition may not adequately model all real-world scenes, potentially leading to failure in enhancing scenes with complex textures or illumination; (2) while VLLVE has demonstrated robustness to correspondence priors, a refinement strategy can be designed to improve correspondence quality during training to further enhance performance.

Thus, we further extend VLLVE to VLLVE++ by introducing an advanced decomposition strategy, coupled with a newly designed correspondence refinement method that filters low-quality correspondences and incorporates high-quality ones during training. This enhanced strategy not only achieves superior frame-level quality but also improves temporal consistency.

To formulate a more flexible decomposition that better captures video content,
we redefine our decomposition as $\I = \L \otimes \R+ \boldsymbol{B}$, where $\L$ and $\R$ still represent the view-dependent and view-independent parts, respectively, and $\boldsymbol{B}$ denotes the remaining component that may not be expressed by $\L$ and $\R$, e.g., some scenario-adaptive intricate lighting components.
Our approach aims to predict $\R$, $\L$, and $\boldsymbol{B}$ of normal-light images from the provided low-light inputs.
\textit{It is important to note that the decomposition of these three terms is not unique, and our goal is to implement a suitable decomposition that enhances the performance of LLVE}.
Building on the trained VLLVE, which provides an initial decomposition of $\L$ and $\R$, we further learn the term $\boldsymbol{B}$ from an initially zero variable, guided by the reconstruction error across multiple frames. To ensure that the learned $\boldsymbol{B}$ adheres to physical principles, we assume it satisfies continuity across different views, i.e., it represents a variable that varies smoothly rather than erratically. 
This newly designed decomposition network aligns more closely with real-world video characteristics, reducing the optimization burden on the view-dependent and view-independent components in the original VLLVE. As a result, it enhances performance across both frame-level and video-level metrics.

Moreover, the original VLLVE framework effectively utilizes input correspondences predicted by existing correspondence estimation networks and has demonstrated satisfactory robustness to variations in correspondence quality and sources (see the experimental section). However, we observe that the enhancement performance can be further improved when more reliable correspondences are available during training. This highlights the need to move beyond the fixed influence of pretrained correspondence networks and instead enable joint learning of video enhancement and correspondence refinement.
To address this, we propose a bidirectional strategy for dual-target learning. Given the initial correspondences, we introduce a correspondence refinement network that learns during training to output a residual correction to the input correspondence maps. The refinement process is guided by two principles: (1) enforcing matched points to share similar view-independent features, (2) view-independent features are also optimized accordingly, creating a positive feedback loop. This refinement mechanism allows for the removal of low-quality correspondences and the inclusion of high-quality ones, ultimately leading to improved enhancement performance, since more accurate correspondences contribute to improved decomposition.

In summary, our contributions are three-fold.
\begin{itemize}
	\item \textbf{Innovative video decomposition strategy for LLVE.} We introduce a novel canonical form (from VLLVE to VLLVE++) for LLVE that predicts spatial-temporal consistent decomposition. 
    This innovative decomposition strategy leverages spatial-temporal correspondences and continuity. Moreover, we facilitate interaction among the features of different frames, leading to a consistent decomposition. 
    \item \textbf{Bidirectional Learning between enhancement and correspondence refinement.}
    In our proposed VLLVE++, we introduce a correspondence refinement network during training, supervised using features from the VLLVE network as guidance. The refined correspondences further improve the low-light enhancement, facilitating bidirectional learning.
\item \textbf{Effective LLVE performance.}
We conduct extensive experiments on public datasets, demonstrating the effectiveness of our framework (VLLVE/VLLVE++) across diverse real-world scenes with varying categories and motion patterns. Furthermore, we showcase the potential of VLLVE++ to be extended to enhancement tasks in more challenging scenarios.

\end{itemize}

A preliminary version of our work, VLLVE, has been accepted to the International Joint Conferences on Artificial Intelligence (IJCAI) 2025~\cite{xu2025low}. In this extended version, we present several key contributions and advancements. First, we enhance the decomposition strategy by reformulating it into three terms: view-independent, view-dependent, and residual terms, which better align with the requirements of real-world modeling. Second, we introduce a bidirectional learning scheme that enables simultaneous correspondence refinement and video enhancement. In this scheme, correspondences are refined using the decomposed features as references, and the decomposed features, in turn, benefit from the refined correspondences. 
These improvements result in VLLVE++ achieving superior performance. 
Moreover, we conduct more extensive experiments to analyze VLLVE and VLLVE++, highlighting essential properties such as robustness, thereby helping the research community better understand the effects of core contributions.
Furthermore, we demonstrate that VLLVE++ can be easily applied to more challenging scenes, including real-world scenes with higher dynamics, the RAW domain, and 3D-related scenes.

\section{Related Work}
\label{sec:formatting}

\subsection{Low-Light Image Enhancement.}
To enhance the quality of a low-light video, image enhancement methods can be applied on a frame-by-frame basis. In recent years, learning-based Low-Light Image Enhancement (LLIE) techniques~\cite{yan2014learning, yan2016automatic, lore2017llnet, cai2018learning,zamir2020learning,xu2020learning,zeng2020learning,kim2021representative,zhao2021deep,zheng2021adaptive,wang2021real,liu2021retinex,yang2021sparse,jiang2021enlightengan,yang2021band} have made significant advancements, with a primary focus on supervised approaches thanks to the abundance of image pairs available for training. Xu et.al.~\cite{xu2022snr} introduced spatial-varying operations, considering Signal-to-Noise Ratio (SNR) as a prior factor.
Among these supervised approaches, Retinex-based methods have proven to be effective due to their adherence to fundamental physical principles. Fu et.al.~\cite{fu2023learning} employed illumination augmentation to a pair of images, achieving self-supervised Retinex decomposition learning.
However, when applying image enhancement algorithms to individual frames, the issue of flickering often arises.

\subsection{Low-Light Video Enhancement.}
In addition to the need for LLIE, there is a growing demand for video enhancement, considering the widespread use of videos as a popular data format on the internet and in photographic equipment. Various approaches have been proposed in this context~\cite{chen2019seeing,triantafyllidou2020low,ye2023spatio,lv2023unsupervised,xu2023deep,fu2023dancing,lin2025geometric}. Danai et.al.~\cite{triantafyllidou2020low} introduced a data synthesis mechanism to generate dynamic video pairs from SID~\cite{chen2019seeing}. Wang et.al.~\cite{wang2021sdsd} proposed a multi-branch network that simultaneously estimates noise and illumination, suitable for videos with severe noise conditions. Liu et.al.~\cite{liu2023low} and Liang et.al.~\cite{liang2023coherent} used prior event information to learn enhancement mapping for brightening videos. Xu et.al.~\cite{xu2023deep} designed a parametric 3D filter tailored for enhancing and sharpening low-light videos. Recently, Fu et.al.~\cite{fu2023dancing} introduced a video enhancement method called LAN, which iteratively refines illumination and adaptively adjusts it. However, it's important to note that LAN lacks an explicit constraint for maintaining consistent reflectance and illumination decomposition.

Several datasets for video enhancement, encompassing both static~\cite{chen2019seeing,jiang2019learning,wang2019enhancing,triantafyllidou2020low} and dynamic motions~\cite{wang2021sdsd,fu2023dancing}, have been introduced. In this paper, we introduce a novel LLVE method, designed to enhance effects on these datasets. Our decomposition method explicitly and consistently models view-dependent and -independent for all frames, with the guidance of videos themselves.

\section{Method: VLLVE}

\begin{figure*}[tb]
    \centering
    \includegraphics[width=.818\linewidth]{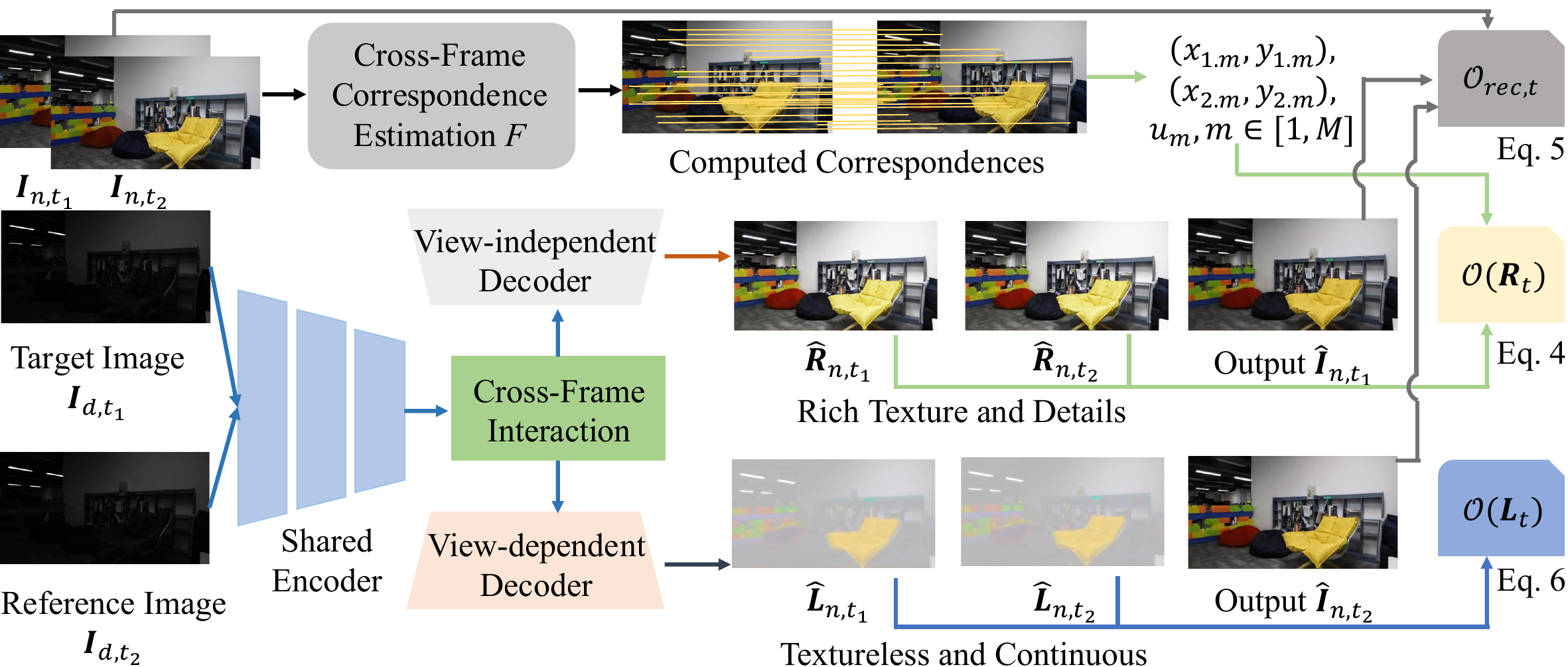}
    \caption{
        Our framework offers a comprehensive solution that explicitly and consistently models the view-independent and view-dependent decomposition of enhanced normal-light outputs across different frames.
        To achieve this, we enforce consistent features in the view-independent terms across different frames by leveraging computed correspondences in the temporal dimension of videos ($\mathcal{O}(\R_t)$). Simultaneously, we ensure that the view-dependent terms exhibit a spatially continuous distribution ($\mathcal{O}(\L_t)$), aligning with real-world scenarios. Furthermore, our network incorporates cross-frame interaction and simultaneous supervision of different frames within a video, encouraging consistent features for these frames derived from one video. For a more detailed visual representation, please refer to \Cref{fig:framework2}. 
    }
    \label{fig:framework}
    \vspace{-0.1in}
\end{figure*}

\subsection{Decomposition Model}
\label{sec:1}

\minisection{Motivation} 
According to the linear Lambertian model, an observed image $\I$ can be formulated as the element-wise product of its albedo component $\boldsymbol{A}$ and shading part $\boldsymbol{S}$, as $\I = \boldsymbol{A} \otimes \boldsymbol{S}$. Here, $\boldsymbol{A}$ is view-independent, capturing the intrinsic appearance, while $\boldsymbol{S}$ is view-dependent.
Inspired by this, when given an image $\I$ (in this paper, we denote the low-light image as $\I_d$ and the normal-light image as $\I_n$, with $d$ and $n$ serving as subscript abbreviations),
we assume its decomposition into view-dependent part $\L$ and view-independent part $\R$, as $\I=\L \otimes \R$,
where $\otimes$ is the element-wise multiplication, $\L$ and $\R$ can be formulated for different channels, i.e., the channel number is 3 for the sRGB domain.
In this context, $\L$ mainly describes the light intensity of objects, which is expected to exhibit piece-wise continuity.
On the other hand, $\R$ primarily represents the physical properties of the objects.
If we obtain $\L$ and $\R$ in the normal-light conditions for a given input low-light image, the target normal-light image can be derived accordingly.
The common objective is summarized as
\begin{equation}
	\mathcal{O}=\Vert \I-\L\otimes \R \Vert + \mathcal{O}(\L)+\mathcal{O}(\R),
	\label{eq:image}
\end{equation}
where $\mathcal{O}(\L)$ and $\mathcal{O}(\R)$ denote the constraints for $\L$ and $\R$.

Some prior approaches (such as the Retinex theory) have been developed to achieve similar decompositions by imposing various constraints on the two components. However, these methods have not incorporated a view-related constraint, which distinguishes our method.
Nevertheless, there is a scarcity of satisfactory decomposition strategy tailored for sequential video data. The objective concerning the decomposition of video data is 
\begin{equation}
	\begin{aligned}
	&\mathcal{O}_v=\mathbb{E}_{t=1:T} [\mathcal{O}_{rec,t} + \mathcal{O}(\L_t)+\mathcal{O}(\R_t)],\\
	&\mathcal{O}_{rec,t}=\Vert \I_t-\L_t\otimes \R_t \Vert,
	\end{aligned}
	\label{eq:video}
\end{equation}
where $\mathbb{E}$ is the average operation, $T$ is the frame number. 

\minisection{Video Data Guide the Decomposition by Themselves}
By establishing correspondences among different frames, we can impose specific constraints: the $\R_t$ of each frame should faithfully represent the intrinsic texture of objects within the target scene, remaining consistent regardless of changes in viewpoint. Likewise, the $\L_t$ of each frame should exhibit the desired continuity consistent with real-world physical properties.

\minisection{Problem Formulation}
Given a clip of low-light data $\I_{d, t}, t\in[1, T]$, there is a paired normal-light data $\I_{n, t}, t\in[1, T]$. 
\textit{We obtain the decomposition of $\I_{n, t}$ from $\I_{d, t}$ using network $f$}, as
\begin{equation}
	\hat{\L}_{n, t}, \hat{\R}_{n, t}= f(\I_{d, t}), \hat{\I}_{n, t}=\hat{\L}_{n, t}\otimes \hat{\R}_{n, t},
    \label{eq:first}
\end{equation}
where $\hat{\L}_{n, t}$ and $\hat{\R}_{n, t}$ are the estimated targets,
and $\hat{\I}_{n, t}$ is the enhancement result.
Our framework is shown in \Cref{fig:framework}.

\subsection{Spatial-Temporal Consistent Decomposition}
\label{sec:2}

\minisection{Motivation} 
Video can be conceptualized as static/dynamic 3D multi-view data~\cite{wang2023tracking,wang2023lighting}. Prior research has demonstrated that the multi-view data itself can be harnessed to decompose the view-dependent and view-independent elements, which respectively encapsulate the intrinsic texture and view-altering illuminations~\cite{wang2023lighting}. Thus, once we establish correspondence relationships among $\I_{d,t}$ for all $t$, we can apply the view-independent constraint to derive $\hat{\R}_{n, t}$, as it represents the intrinsic properties of the target scene. 
Subsequently, obtaining $\hat{\L}_{n, t}$ becomes achievable through the utilization of the inherent image reconstruction loss, in conjunction with adherence to the continuity assumption.

\minisection{Implementation of Correspondences}
Obtaining correspondences for $\I_{d,t}$, $t\in[1, T]$, can be a challenging task due to the presence of various degradations, including visibility issues and noise. Fortunately, we have access to corresponding normal-light data, which significantly aids in establishing these correspondences, as illustrated in \Cref{fig:framework}, where $\I_{d,t}$ and $\I_{n,t}$ are pixel-wise aligned.

Given two frames $\I_{n,t_1}$ and $\I_{n,t_2}$, correspondences can be determined using the prediction network $F$~\cite{edstedt2023dkm}. We assume the detection of $M$ correspondences, denoted as ${ c_m=(x_{1,m}, y_{1,m}, x_{2,m}, y_{2,m}), m\in[1, M]}$. Each correspondence, represented by $c_m$, consists of four coordinates: $(x_{1,m}, y_{1,m})$ represents the pixel coordinate in $\I_{n,t_1}$, while $(x_{2,m}, y_{2,m})$ signifies the coordinate in $\I_{n,t_2}$. Furthermore, each correspondence is associated with an uncertainty value $u_m$, determined through a data-driven approach.

\minisection{Constraints Formulation}
To satisfy the constraint that corresponding pixels in $\R_t$ are matched, we formulate $\mathcal{O}(\R_t)$ as
\begin{equation}
	\begin{aligned}
		\mathcal{O}(\R_{t_1,t_2})=\mathbb{E}_{m\in[1, M]} u_m\Vert &\hat{\R}_{n, t_1}[x_{1,m}, y_{1,m}]- \\ &\hat{\R}_{n, t_2}[x_{2,m}, y_{2,m}] \Vert .
	\end{aligned}
	\label{eq:1}
\end{equation}
The reconstruction item in \Cref{eq:video} can be denoted as 
\begin{equation}
	\mathcal{O}_{rec,t}=\Vert \I_{n, t}-\hat{\L}_{n, t}\otimes \hat{\R}_{n, t} \Vert.
	\label{eq:2}
\end{equation}
Although the view-dependent part can be obtained via the reconstruction constraint, 
We further incorporate a continuity constraint to align with real-world properties. 
For a pixel in the video frame, denoted as $p$, we introduce a spatial continuity objective on the predicted $\hat{\L}_{n,t}$ as the following loss function
	\begin{equation}
		\mathcal{O}(\L_t)=\mathbb{E}_t [v_t^p \times [\partial_x \hat{\L}_{n,t}(p)]^2 + u_t^p \times [\partial_y \hat{\L}_{n,t}(p)]^2],
		\label{eq:3}
	\end{equation}
	where $\partial_x$ and $\partial_y$ are partial derivatives in horizontal and vertical directions, respectively. $v_t^p$ and $u_t^q$ are spatially-varying smoothness weights, calculated as
	\begin{equation}
 \small
		v_t^p=(\Vert \partial_x \boldsymbol{U}_{t}(p) \Vert ^{1.2} + \Delta )^{-1}, u_t^p=(\Vert \partial_y \boldsymbol{U}_t(p) \Vert ^{1.2}+ \Delta)^{-1},
		\label{eq:smooth}
	\end{equation}
	where $\boldsymbol{U}_t$ represents the logarithmic transformation of $\I_{d,t}$, and $\Delta$ is a small constant (set to 0.0001) used to avoid division by zero. 
	The number ``1.2'' is chosen empirically.
	The diagram for the decomposition implementation is shown in \Cref{fig:framework}.

\minisection{The Influence of Correspondences Quality}
In practice, there are situations where the computation of correspondences can be influenced by factors such as occlusions. We emphasize that we only employ high-quality correspondences with low uncertainty values. The performance of our strategy remains robust to certain perturbations in these filtered correspondences (please view the experimental section for the empirical verification).

 \begin{figure}[tb!]
     \centering
     \includegraphics[width=1\linewidth]{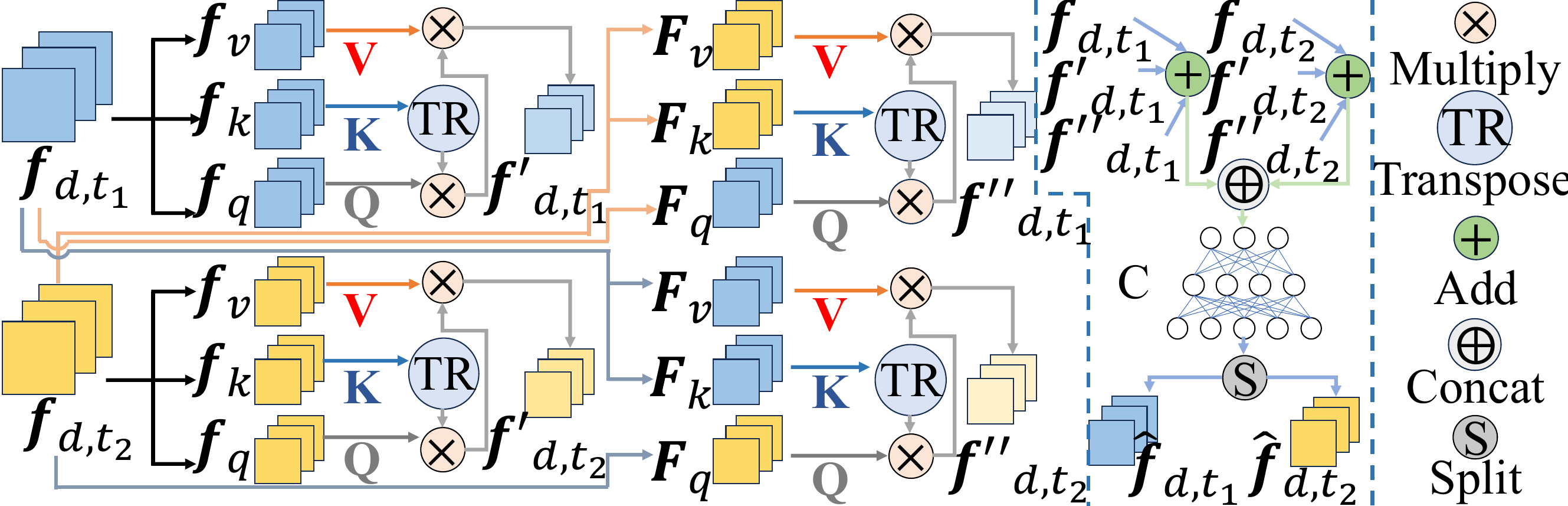}
     \caption{The lightweight Cross-Frame Interaction Mechanism (CFIM) propagates different frames' features, along with cross-frame attention and spatial-channel fusion. 
     Cross-frame interaction can be employed in the deep feature space of arbitrary single-image encoder-decoder frameworks, and we choose U-Net here.
     }
     \label{fig:framework2}
\end{figure}

	\subsection{Dual Network for Consistent Decomposition}
	\label{sec:3}

\vspace{-0.1in}
	\minisection{Overview and Motivation} 
	To enhance the consistency of decomposition, it's crucial for various frames within a video to mutually share features. By incorporating shared features and simultaneous supervision, the features used for synthesizing decomposition parameters across different frames can exhibit greater consistency. When the video is enhanced frame by frame without feature propagation, the decomposition results tend to be suboptimal due to the inherently challenging nature of maintaining consistency. 
    Current LLVE methods primarily leverage multiple-frame inputs to facilitate propagation~\cite{wang2021sdsd,fu2023dancing}. However, these strategies come with a notable cost in terms of aligning features.

	In this paper, we introduce an efficient strategy for propagation.
    This approach entails loading two frames, specifically target frame $\I_{d, t_1}$ and reference frame $\I_{d, t_2}$ (where $\I_{d, t_2}$ can be randomly selected from the temporal neighbors of $\I_{d, t_1}$ during training, and is set as the closest frame of $\I_{d, t_1}$ during inference),
    propagating features at the deepest layer of the network through our cross-frame interaction (as depicted in \Cref{fig:framework2}), and concurrently supervising these dual outputs to guarantee consistency.

\vspace{-0.1in}
 \minisection{Feature propagation via spatial-varying fusion.}
	Suppose the feature of $\I_{d, t_1}$ and $\I_{d, t_2}$ are denoted as $\boldsymbol{f}_{d, t_1}$ and $\boldsymbol{f}_{d, t_2}$ which is extracted from the same encoder $M_E$ in the network $f$.
    To complete the propagation, we initially employ a long-range cross-frame attention operation in the feature space. A traditional attention operation~\cite{vaswani2017attention} typically consists of the query vector $\boldsymbol{Q}$, the key vector $\boldsymbol{K}$, and the value vector $\boldsymbol{V}$. The attention relationship is established using $A(\boldsymbol{Q},\boldsymbol{K},\boldsymbol{V})=\text{softmax}(\boldsymbol{Q}\times \boldsymbol{K}^\top)\times \boldsymbol{V}$, where $\times$ represents matrix multiplication.
	To enable cross-frame attention, we process the feature via dual paths, as
	\begin{equation}
		\small
		\begin{aligned}
			&\boldsymbol{f}'_{d, t_1}=A(\boldsymbol{f}_{d, t_1},\boldsymbol{f}_{d, t_1},\boldsymbol{f}_{d, t_1}), \boldsymbol{f}''_{d, t_1}=A(\boldsymbol{f}_{d, t_1},\boldsymbol{f}_{d, t_2},\boldsymbol{f}_{d, t_2}),\\
			&\boldsymbol{f}'_{d, t_2}=A(\boldsymbol{f}_{d, t_2},\boldsymbol{f}_{d, t_2},\boldsymbol{f}_{d, t_2}), \boldsymbol{f}''_{d, t_2}=A(\boldsymbol{f}_{d, t_2},\boldsymbol{f}_{d, t_1},\boldsymbol{f}_{d, t_1}).
		\end{aligned}
        \label{eq:fusion1}
	\end{equation}
	Moreover, a short-range fusion operation is set as the refinement operation to the propagation at both spatial and channel levels, as
	\begin{equation}
		\small
		\begin{aligned}
		\hat{\boldsymbol{f}}_{d, t_1},\hat{\boldsymbol{f}}_{d, t_2}=S(C(&(\boldsymbol{f}_{d, t_1}+\boldsymbol{f}'_{d, t_1}+\boldsymbol{f}''_{d, t_1})\oplus\\ &(\boldsymbol{f}_{d, t_2}+\boldsymbol{f}'_{d, t_2}+\boldsymbol{f}''_{d, t_2}))),
		\end{aligned}
        \label{eq:fusion2}
	\end{equation}
	where $\oplus$ represents channel concatenation, $C$ refers to the convolution network, and $S$ signifies channel dimension splitting. The decomposition results for $\I_{d,t_1}$ and $\I_{d,t_2}$ are obtained by processing $\hat{\boldsymbol{f}}_{d, t_1}$ and $\hat{\boldsymbol{f}}_{d, t_2}$ through the decoder $M_D$.

 \minisection{The role of the cross-attention in CFIM}
 The cross-attention mechanism facilitates information propagation across different frames, aligning with our training procedure that requires spatial consistency for $\R$. 
Moreover, during training, CFIM brings varying filtered features from randomly selected reference images ($\I_{d,t_2}$) into the backbone (i.e., the part without feature propagation), where the input is the image at the current time step $\I_{d,t_1}$. This setup enables CFIM to guide the backbone in learning how to adaptively leverage diverse knowledge from reference images, resulting in robust and effective feature spaces within the backbone, without the need for expensive temporal alignment. In other words, the backbone performs strongly without a reference image during inference (though, of course, having a suitable reference image to provide additional useful information generally improves performance), as confirmed by the ablation study.
Additionally, CFIM can leverage useful cross-frame information when it complements the features in the backbone, further enhancing the decomposition process.
	
	\subsection{Overall Objective to Compute Loss Function}
	During the training, we find it is better to adopt the dual training strategy, i.e., for each $\I_{d, t_1}$, we sample its neighboring reference $\I_{d, t_2}$ and constrain their outputs simultaneously.
	Thus, the loss function can be written as
	\begin{equation}
		\begin{aligned}
					\mathcal{O}_v=\mathbb{E}_{t_1, t_2} [ &\mathcal{O}_{rec, t_1} + \mathcal{O}_{rec, t_2} +\\ & \lambda_1 (\mathcal{O}(\L_{t_1})+\mathcal{O}(\L_{t_2}))+\lambda_2\mathcal{O}(\R_{t_1,t_2})],
		\end{aligned}
		\label{eq:video2}
	\end{equation}
	where $\lambda_1$ and $\lambda_2$ are the loss weights, and each loss term is defined in \Cref{eq:1}, \Cref{eq:2} and \Cref{eq:3}.

	\subsection{The Differences Between Existing Methods}
	We want to emphasize the differences between our method and existing Retinex-based image restoration method.
    Besides the difference of basic decomposition assumption,
	some existing strategies apply Retinex theory for image enhancement~\cite{liu2021retinex,cai2023retinexformer,fu2023learning}, while our framework is tailored for video enhancement. Implementing decomposition for video data presents unique challenges and differences compared to image data, highlighting the novelty of our approach. 1) We introduce novel loss terms for learning $\R$ and $\L$. We establish new spatial-temporal constraints for $\R$ by leveraging pixel-level correspondences and their uncertainties, a method not proposed by previous works. Additionally, we design spatial-continuous terms for $\L$. 2) Moreover, we propose utilizing a cross-frame mechanism with randomly selected reference frames to align with our newly designed training procedure.
	
	Moreover, the proposed CFIM also differs from existing cross-view interaction in image enhancement.
	The cross-view interaction presented in \cite{zheng2023decoupled} differs from our cross-attention mechanism. While \cite{zheng2023decoupled} focuses solely on horizontal correlations between two views, our attention computation addresses both horizontal and vertical directions. Additionally, our cross-attention approach contrasts with the view transfer modules in \cite{huang2022low} and \cite{zhang2024stereo}, which use parallax attention maps that are computed mutually between two features. In contrast, our method employs the attention strategy utilized in the transformer, which needs query, key, and value feature maps.

The overall training algorithm for VLLVE is summarized in Alg.~\ref{alg:1}.

\begin{algorithm}[t]
\caption{The training procedure of VLLVE}
\label{alg:1}
\begin{algorithmic}
\renewcommand{\algorithmicrequire}{\textbf{Input:}}
\renewcommand{\algorithmicensure}{\textbf{Output:}}

\algnewcommand\algorithmicforeach{\textbf{for each}}
\algdef{S}[FOR]{ForEach}[1]{\algorithmicforeach\ #1\ \algorithmicdo}
    \Require{Training dataset with pairs $\{(\I_{d,t}, \I_{n,t})\}$, initialized target enhancement network $G$, pre-trained correspondence estimation network $F$, current training iteration $T$, maximum training iteration $T_{max}$},

\While {$T < T_{max}$}
\State Sample training batch $(\I_{d,t_1}, \I_{n,t_1})$ and $(\I_{d,t_2}, \I_{n,t_2})$
\State Use the frozen correspondence estimation network $F$ to predict the correspondences ${ c_m=(x_{1,m}, y_{1,m}, x_{2,m}, y_{2,m}), m\in[1, M]}$
\State Feed the input of $\I_{d,t_1}$ and $\I_{d,t_2}$ into the enhancement network, obtain the hidden feature as $\boldsymbol{f}_{d,t_1}$ and $\boldsymbol{f}_{d,t_2}$
\State Conduct the interaction between these two features with Eqs.~\ref{eq:fusion1} and \ref{eq:fusion2}, obtaining $\hat{\boldsymbol{f}}_{d,t_1}$ and $\hat{\boldsymbol{f}}_{d,t_2}$
\State Obtain the enhancement results from $\hat{\boldsymbol{f}}_{d,t_1}$ and $\hat{\boldsymbol{f}}_{d,t_2}$ via the decoders as $\hat{\L}_{n,t_1}$, $\hat{\L}_{n,t_2}$, $\hat{\R}_{n,t_1}$, and $\hat{\R}_{n,t_2}$
\State Compute the loss term with Eq.~\ref{eq:video2} (including the loss terms in Eqs.~\ref{eq:1}, \ref{eq:2}, \ref{eq:3}, and \ref{eq:smooth})
\State Update the network of $G$, $T=T+1$
\EndWhile
\Ensure{Trained model $G$}
\end{algorithmic}
\end{algorithm}

 \begin{figure*}[tb!]
     \centering
     \includegraphics[width=.818\linewidth]{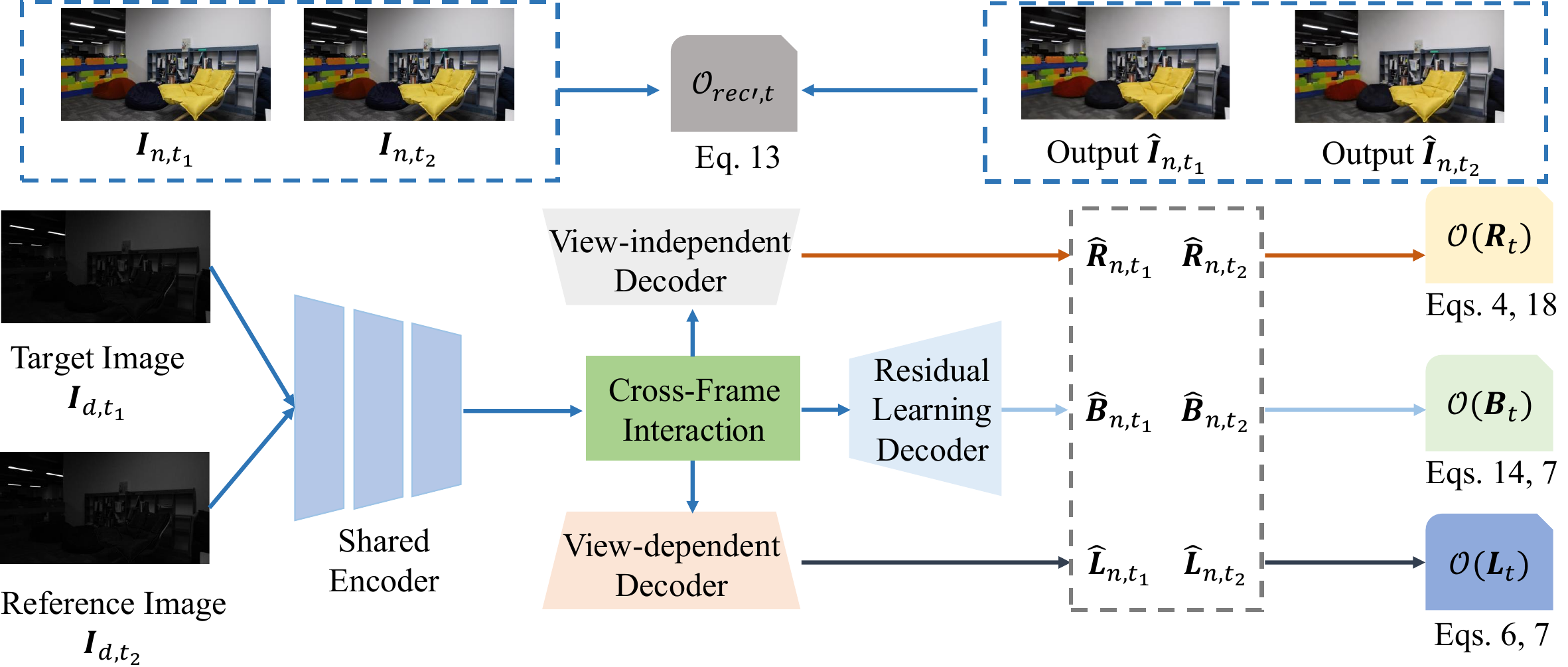}
     \vspace{-0.1in}
     \caption{The overview of VLLVE++ with new decomposition strategy, including the network structure and the training loss terms.
     }
     \label{fig:vllve++}
\end{figure*}

 \begin{figure}[tb!]
     \centering
     \includegraphics[width=1\linewidth]{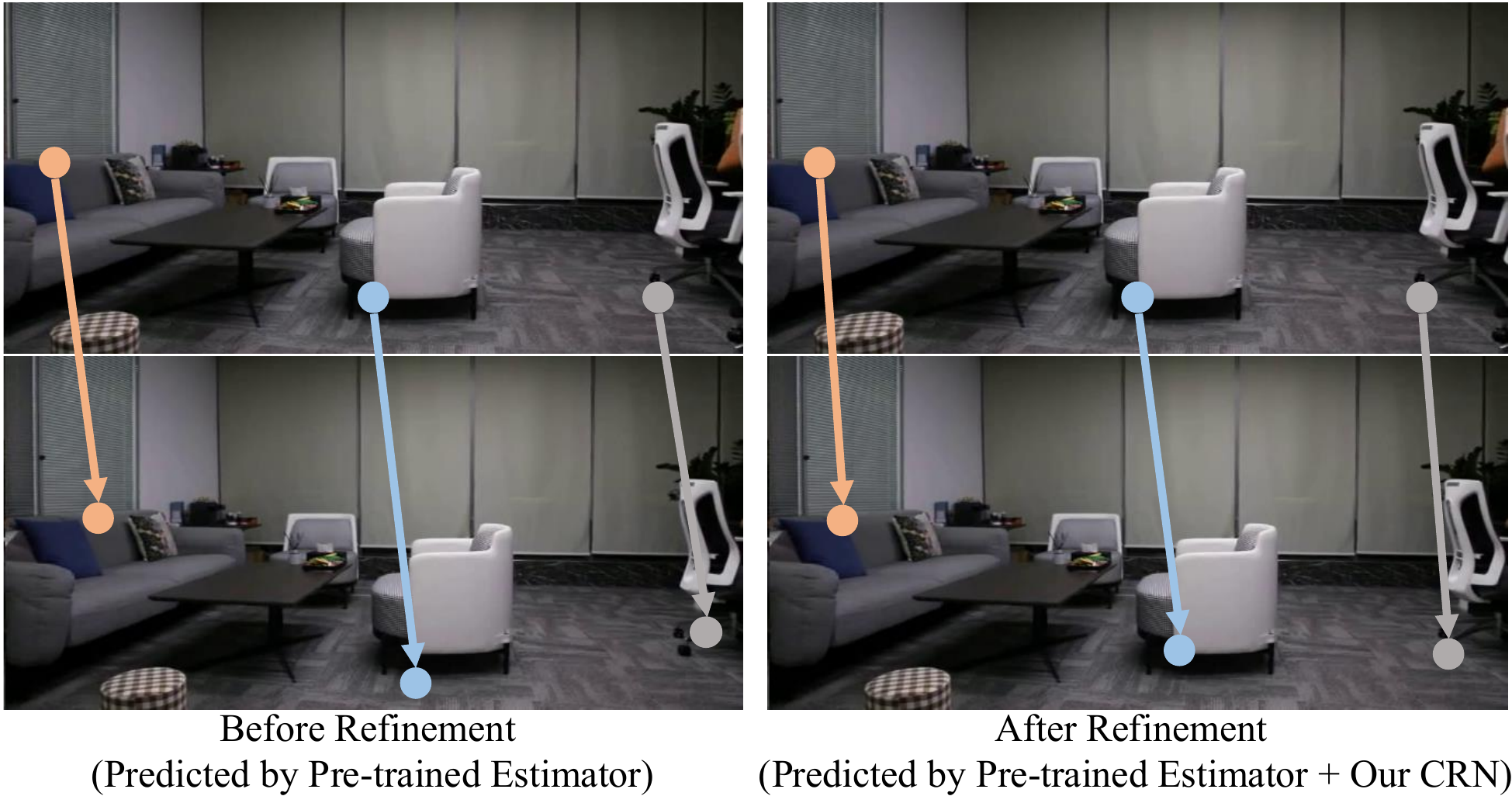}
     \vspace{-0.2in}
     \caption{Visualization of incorrect correspondences predicted by the pretrained estimator DKM (image from the LLVE test set). Although pretrained models have been optimized for generalization, they inevitably generate some erroneous correspondences because they are not fine-tuned on LLVE. These mismatched points often exhibit unduly distinct values. To this, we refine them by aligning matched points to a view-independent intrinsic value. While RGB values can also be used for alignment, their effect is weaker (see \Cref{comparison-vertify}). As for implementation, we introduce a correspondence refinement network trained jointly with the LLVE model. For illustration, we visualize three representative points from the predicted correspondences here.
     }
     \label{fig:refine-visual}
\end{figure}

\section{Method: VLLVE++}

Building upon VLLVE, we propose VLLVE++, which incorporates a novel decomposition strategy by introducing a residual term. In addition, we design a bidirectional learning mechanism that jointly enhances image quality and refines correspondences to further improve performance.

\subsection{Complex Decomposition with Residual Term}
In the VLLVE framework, decomposition is achieved through the multiplicative operation between $\hat{\L}_{n,t}$ and $\hat{\R}_{n,t}$ as shown in Eq.~\ref{eq:first}. However, relying solely on multiplication may constrain its reconstruction capacity due to the complexity of real-world data. A relevant example is the evolution of the Lambertian model: 
according to principles of optical imaging physics, the classical linear Lambertian model can be extended into a more comprehensive formulation by introducing a specular component as an additive residual term. 
The difference can be expressed as
\begin{equation}
    \begin{aligned}
        \text{old}:& \text{image} \; \I = \text{albedo} \;\boldsymbol{A} \otimes \text{shading} \; \boldsymbol{S},\\
        \text{new}:& \text{image} \; \I = \text{albedo} \;\boldsymbol{A} \otimes \text{shading} \;\boldsymbol{S}+\text{specular} \;\boldsymbol{P}.
    \end{aligned}
\end{equation}
Here, the specular term primarily accounts for the complex and variable components that cannot be fully represented by albedo and shading alone. 
We also think residual terms might help to capture additional physical factors, particularly those that are difficult to model explicitly due to their dependence on environmental conditions, material properties, and other influencing factors.

In the following sections, we present the formulation of our new decomposition equation and describe the loss terms designed to facilitate its learning.

\subsection{Decomposition Target and Networks}
First, we present the formulation of the newly designed decomposition, which preserves the original view-independent and view-dependent parts while introducing an additional residual term. The decomposition can be expressed as follows:
\begin{equation}
    \hat{\I}_{n, t}=\hat{\L}_{n, t}\otimes \hat{\R}_{n, t}+\hat{\boldsymbol{B}}_{n,t},
\end{equation}
where $\hat{\L}_{n, t}$ and $\hat{\R}_{n, t}$ represent the view-dependent and view-independent components, respectively, while $\hat{\boldsymbol{B}}_{n,t}$ denotes the residual component. The overall framework is illustrated in Fig.~\ref{fig:vllve++}, where separate decoders are assigned to $\hat{\L}_{n, t}$, $\hat{\R}_{n, t}$, and $\hat{\boldsymbol{B}}_{n,t}$ individually. All decoders share a common feature space extracted by the encoder, ensuring mutual relevance and mitigating potential conflicts that could arise from a lack of information interaction.

To ensure that the new residual component $\hat{\boldsymbol{B}}_{n,t}$ primarily captures information beyond what has already been learned by $\hat{\L}_{n, t}$ and $\hat{\R}_{n, t}$, we base the training process on a pre-trained VLLVE model and initialize the output of $\hat{\boldsymbol{B}}_{n,t}$ to zero. Since VLLVE is trained to reconstruct the target $\I_{n, t}$ using $\hat{\L}_{n, t}\otimes \hat{\R}_{n, t}$ as closely as possible, the learning of $\hat{\boldsymbol{B}}_{n,t}$ focuses on capturing additional knowledge not represented by the original decomposition.
Therefore, an additional reconstruction loss is applied as
\begin{equation}
	\mathcal{O}_{rec',t}=\Vert \I_{n, t}-(\hat{\L}_{n, t}\otimes \hat{\R}_{n, t}+\hat{\boldsymbol{B}}_{n,t}) \Vert.
	\label{eq:22}
\end{equation}

Moreover, similar to the loss terms for $\hat{\L}_{n, t}$ and $\hat{\R}_{n, t}$, we emphasize the importance of enforcing a suitable constraint on $\hat{\boldsymbol{B}}_{n,t}$. We note that $\hat{\boldsymbol{B}}_{n,t}$ represents physical factors not accurately captured by $\hat{\L}_{n, t}$ and $\hat{\R}_{n, t}$, and these factors are generally continuous and smoothly varying, rather than abrupt or mutational, across real-world temporal and spatial dimensions.
Therefore, a continuity loss function is defined as follows
\begin{equation}
		\mathcal{O}(\hat{\boldsymbol{B}}_t)=\mathbb{E}_t [v_t^p \times [\partial_x \hat{\boldsymbol{B}}_{n,t}(p)]^2 + u_t^p \times [\partial_y \hat{\boldsymbol{B}}_{n,t}(p)]^2],
		\label{eq:33}
\end{equation}
which is similar to Eq.~\ref{eq:3}, with the pixel-wise weights $v_t^p$ and $u_t^p$ defined as in Eq.~\ref{eq:smooth}.

Thus, the overall loss term can be set as
	\begin{equation}
		\begin{aligned}
					\mathcal{O}_v=\mathbb{E}_{t_1, t_2} [ &\mathcal{O}_{rec, t_1} + \mathcal{O}_{rec, t_2} +\\ 
                    &\mathcal{O}_{rec', t_1} + \mathcal{O}_{rec', t_2} +\\ 
                    & \lambda_1 (\mathcal{O}(\L_{t_1})+\mathcal{O}(\L_{t_2}))+\lambda_2\mathcal{O}(\R_{t_1,t_2})\\
                    & \lambda_3 (\mathcal{O}(\boldsymbol{B}_{t_1})+\mathcal{O}(\boldsymbol{B}_{t_2}))],
		\end{aligned}
		\label{eq:video22}
	\end{equation}
where $\lambda_1$, $\lambda_2$, and $\lambda_3$ represent the loss weights. The inclusion of $\mathcal{O}_{rec, t_1}$, $\mathcal{O}_{rec, t_2}$,  $\mathcal{O}(\L_{t_1})$, $\mathcal{O}(\L_{t_2})$, and $\mathcal{O}(\R_{t_1,t_2})$ is motivated by two reasons. First, they ensure that $\R_t$ and $\L_t$ continue to be learned according to their original objectives, preventing their representations from being mixed with $\boldsymbol{B}_t$. Second, jointly optimizing $\R_t$, $\L_t$, and $\boldsymbol{B}_t$ promotes consistency among these three components, which is reinforced by our dual-network architecture.

 \begin{figure*}[tb!]
     \centering
     \includegraphics[width=.818\linewidth]{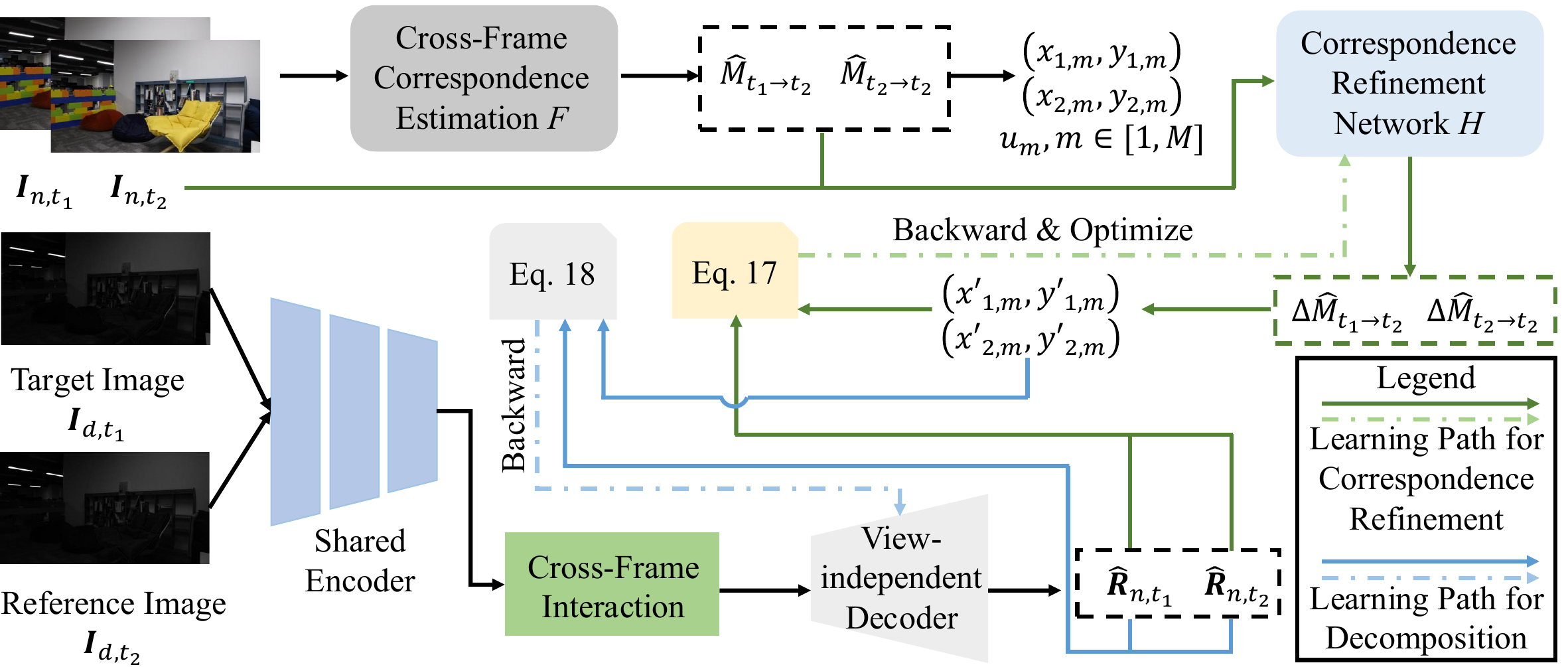}
     \vspace{-0.1in}
     \caption{The overview of our correspondences refinement strategy with a self-supervised manner. The learning of correspondences refinement and view-independent parts is bi-directional.
     }
     \label{fig:vllve++cr}
\end{figure*}

\vspace{-0.1in}
\subsection{Correspondences Refinement with Decomposition
}
\noindent\textbf{Motivation}. In addition to improve the decomposition strategy, we also incorporate an approach for correspondence refinement, as the initial estimated correspondences may contain errors as shown in Fig.~\ref{fig:refine-visual}. 
This issue arises because the pre-trained correspondence estimation networks have not been fine-tuned on the LLVE training datasets, due to the absence of ground truth correspondences in these datasets.

Relying solely on the generalization ability of correspondence estimation networks is reasonable while leaves room for improvement. Previous methods~\cite{edstedt2023dkm,zhu2023pmatch} have demonstrated strong generalization through evaluations on unseen real-world indoor and outdoor scenes using cross-dataset settings; however, errors still persist in the corresponding test datasets. Although several works have introduced various techniques to enhance robust generalization (e.g., incorporating pre-trained features from the foundational generalization model DINOv2~\cite{edstedt2024roma}), complete generalization cannot be fully guaranteed.

Therefore, to ensure strong performance in target scenes, we need a strategy to improve correspondence estimation on the corresponding datasets via finetuning. We observe that learning the view-independent component and refining correspondences are mutually beneficial processes. The view-independent component can partially assess whether matched correspondences share the same content, while improved correspondences, in turn, enhance the learning of the view-independent component. Motivated by this insight, we propose a self-supervised learning strategy.

\noindent\textbf{Overview of refinement strategy}.
Suppose we have utilized a pre-trained correspondence estimation network to train the VLLVE. 1) We first learn an initial view-independent component using the loss function in Eq.~\ref{eq:1}. 
2) Next, we use a similar loss term with a gradient stop operation to refine the correspondences. We predict the correspondence refinement by inputting the two frames and the initially estimated correspondence maps into the refinement network. Then, we apply the same formulation as Eq.~\ref{eq:1}, with a modified gradient flow, encouraging the refined matched correspondences to be closer in terms of the content represented by the view-independent components. 
3) Furthermore, these refined correspondences can be used to better supervise the learning of the view-independent components, creating a positive feedback loop between the learning of view-independent parts and correspondence refinement. The overall pipeline is illustrated in Fig.~\ref{fig:vllve++cr}.

\vspace{-0.1in}
\subsection{Implementation of Correspondence Refinement}
Here, we present the details of the correspondence refinement network $H$, including its architecture, inputs, and loss terms.

\noindent\textbf{Correspondence refinement network.}
The goal of the Correspondence Refinement Network (CRN) is to identify potential errors in the predictions produced by the pre-trained Correspondence Estimation Networks (CENs). Therefore, its inputs include the original two frames used by the pre-trained CENs, as well as the estimated correspondence outputs from these networks. This design allows the CRN to both observe the pre-trained network results and access the original information, which serves as the basis for refining the correspondence estimates.

Suppose the input frames are denoted as ${\I}_{n,t_1}$ and ${\I}_{n,t_2}$, and the correspondence maps as $\hat{\boldsymbol{M}}_{t_1 \rightarrow t_2}$ and $\hat{\boldsymbol{M}}_{t_2 \rightarrow t_1}$, which are used to obtain the correspondences $M$ in Eq.~\ref{eq:1}. The CRN takes as input the concatenation of ${\I}_{n,t_1}$, ${\I}_{n,t_2}$, $\hat{\boldsymbol{M}}_{t_1 \rightarrow t_2}$, and $\hat{\boldsymbol{M}}_{t_2 \rightarrow t_1}$, and outputs the refined correspondences as
\begin{equation}
    \resizebox{0.87\linewidth}{!}{$\Delta \hat{\boldsymbol{M}}_{t_1 \rightarrow t_2}, \; \Delta \hat{\boldsymbol{M}}_{t_2 \rightarrow t_1}=H({\I}_{n,t_1} \oplus {\I}_{n,t_2} \oplus \hat{\boldsymbol{M}}_{t_1 \rightarrow t_2} \oplus \hat{\boldsymbol{M}}_{t_2 \rightarrow t_1})$},
    \label{eq:refine}
\end{equation}
where $\Delta \hat{\boldsymbol{M}}_{t_1 \rightarrow t_2}$ and $\Delta \hat{\boldsymbol{M}}_{t_2 \rightarrow t_1}$ denote the predicted refinements for $\hat{\boldsymbol{M}}_{t_1 \rightarrow t_2}$ and $\hat{\boldsymbol{M}}_{t_2 \rightarrow t_1}$, respectively, both having the same shape as the original correspondence maps.

The correspondence refinement network, CRN, is implemented as a simple convolutional neural network with an encoder-decoder architecture following the UNet design (without Tanh or Sigmoid activation function at the output layer). 
It is important to note that CRN does not share parameters with either the LLVE enhancement network or the pre-trained correspondence estimation networks. Furthermore, CRN is only used during training and is not required during LLVE inference, similar to the pre-trained CENs. 

\noindent\textbf{The training loss of CRN.}
During training, the CRN produces refined correspondence maps, resulting in new correspondence maps denoted as $\hat{\boldsymbol{M}'}_{t_1 \rightarrow t_2}$ and $\hat{\boldsymbol{M}'}_{t_2 \rightarrow t_1}$. Similarly, we obtain the refined set of $M$ correspondences, represented as ${ c'_m=(x'_{1,m}, y'_{1,m}, x'_{2,m}, y'_{2,m}), m\in[1, M]}$. Each correspondence $c'_m$ consists of four coordinates: $(x'_{1,m}, y'_{1,m})$ indicates the pixel position in $\I_{n,t_1}$, while $(x'_{2,m}, y'_{2,m})$ corresponds to the pixel position in $\I_{n,t_2}$. We then define two new loss terms based on these correspondences to supervise the training of the CRN, which also indirectly benefits the learning of the enhancement network.

For supervising the CRN, we adopt a self-supervised approach. This is because the enhancement network is trained to predict view-independent components that preserve consistent content across different views. Therefore, if the predicted correspondences show differences in their view-independent components, it indicates potential errors. The CRN training builds upon the pretrained VLLVE network, which has already learned to extract view-independent features. Accordingly, the loss to train the CRN is
\begin{equation}
	\begin{aligned}
		\mathcal{O}(\R'_{t_1,t_2})=\mathbb{E}_{m\in[1, M]} u_m\Vert &\textrm{S.G.}(\hat{\R}_{n, t_1})[x'_{1,m}, y'_{1,m}]- \\ &\textrm{S.G.}(\hat{\R}_{n, t_2})[x'_{2,m}, y'_{2,m}] \Vert,
	\end{aligned}
	\label{eq:1-crn}
\end{equation}
where $\textrm{S.G.}$ denotes the stop-gradient operation, which prevents updates to the view-independent component during backpropagation. This ensures that only the parameters of the CRN are optimized to predict the new coordinate values ${ c'_m=(x'_{1,m}, y'_{1,m}, x'_{2,m}, y'_{2,m}), m\in[1, M]}$. Although this loss has a formulation similar to that of Eq.~\ref{eq:1}, its purpose is distinct.

On the other hand, the refined coordinate values are expected to be more accurate, and they can, in turn, be used to supervise the learning of the view-independent components. Since these components represent the same content across different frames, this supervision aligns with the fundamental definition of correspondences. The loss term can be expressed as
\begin{equation}
	\begin{aligned}
		\mathcal{O}(\R''_{t_1,t_2})=\mathbb{E}_{m\in[1, M]} u_m\Vert &\hat{\R}_{n, t_1}[\textrm{S.G.}(x'_{1,m}, y'_{1,m})]- \\ &\hat{\R}_{n, t_2}[\textrm{S.G.}(x'_{2,m}, y'_{2,m})] \Vert,
	\end{aligned}
	\label{eq:1-new}
\end{equation}
where the stop-gradient operation is applied to the refined coordinate values, preventing the CRN parameters from being updated through this loss term.
Combining Eq.~\ref{eq:1-crn} and Eq.~\ref{eq:1-new}, we establish a positive feedback cycle: first, the decomposition network supervises the refinement of correspondences, and then the refined correspondences, in turn, facilitate improved learning of the decomposition network.

In summary, the overall loss function for VLLVE++, which includes both the decomposition loss and the correspondence refinement loss, can be expressed as
	\begin{equation}
		\begin{aligned}
					\mathcal{O}_v=\mathbb{E}_{t_1, t_2} [ &\mathcal{O}_{rec, t_1} + \mathcal{O}_{rec, t_2} + \mathcal{O}_{rec', t_1} + \mathcal{O}_{rec', t_2} +\\ 
                    & \lambda_1 (\mathcal{O}(\L_{t_1})+\mathcal{O}(\L_{t_2}))+\\
                    &\lambda_2(\mathcal{O}(\R_{t_1,t_2})+\mathcal{O}(\R'_{t_1,t_2})+\mathcal{O}(\R''_{t_1,t_2}))\\
                    & \lambda_3 (\mathcal{O}(\boldsymbol{B}_{t_1})+\mathcal{O}(\boldsymbol{B}_{t_2}))].
		\end{aligned}
		\label{eq:video222}
	\end{equation}
The overall training algorithm is summarized in Alg.~\ref{alg:2}.

\subsection{Verification of Correspondence Refinement}
We verify whether the correspondence has actually been refined.
A visual example is shown in Fig.~\ref{fig:refine-visual}, and a quantitative analysis is presented below.
By utilizing computed correspondences among neighboring frames, we compute point trajectories across \textit{neighboring frames}. This aligns with standard temporal consistency evaluation strategies~\cite{lai2018learning}, with the difference that previous methods rely on dense optical flow,
whereas we use sparse trajectories. We measure the alignment error ($L_2$ norm) of points along these trajectories to assess consistency: 
whether refined correspondences identify points with identical content (matched content normally results in lower values for alignment errors).

The evaluation is conducted on two dynamic testing sets, SDSD and DID, where propagation errors are computed on the normal-light frames to avoid the influence of image enhancement artifacts. For VLLVE++, propagation errors are calculated using refined correspondences, whereas the baseline errors are derived from correspondences estimated by the pre-trained CEN.
As shown in Table~\ref{comparison-vertify}, these results (``After Refine'' v.s. ``Baseline'') show that using CRN achieves lower alignment error, indicating a reduction in incorrect correspondences.
Moreover, using the predicted RGB values also brings some improvement (as shown in Table~\ref{comparison-vertify}, under the setting ``After Refine (using RGB)'', by replacing $\hat{\R}_{n, t_1}$ and $\hat{\R}_{n, t_2}$ to $\hat{\I}_{n, t_1}$ and $\hat{\I}_{n, t_2}$ in Eq.~\ref{eq:1-crn}), since the CRN is still fine-tuned on the LLVE datasets and thus leverages their knowledge. However, the effect is weaker compared to using the predicted view-independent values as supervision. This is because the values of matched points in the view-independent component are more intrinsic and robust to viewpoint changes, making them more beneficial for correspondence estimation.

\begin{table}[tb!]
    \scriptsize
	\centering
	\huge
	\caption{Experiments verify that our correspondence refinement helps correct misalignments, reducing the value of alignment error. Incorrect correspondences typically result in larger alignment errors.} 
    \label{comparison-vertify}
    \vspace{-0.1in}
    \renewcommand{\arraystretch}{1.06}
    \resizebox{1.0\linewidth}{!}
    {
        \begin{tabular}{|l|p{4.5cm}<{\centering}|p{4.5cm}<{\centering}|p{4cm}<{\centering}|}
            \hline
            & SDSD-indoor &SDSD-outdoor&DID  \\
            \hline
            Baseline &0.014&0.011&0.017\\
            After Refine (using RGB) & 0.012 & 0.010 & 0.015\\
            After Refine &\textbf{0.010} & \textbf{0.008} & \textbf{0.012}\\
            \hline
        \end{tabular}
        }
\end{table}

\begin{algorithm}[t]
\caption{The training procedure of VLLVE++}
\label{alg:2}
\begin{algorithmic}
\renewcommand{\algorithmicrequire}{\textbf{Input:}}
\renewcommand{\algorithmicensure}{\textbf{Output:}}

\algnewcommand\algorithmicforeach{\textbf{for each}}
\algdef{S}[FOR]{ForEach}[1]{\algorithmicforeach\ #1\ \algorithmicdo}
    \Require{Training dataset with pairs $\{(\I_{d,t}, \I_{n,t})\}$, initialized target enhancement network $G$ and the correspondence refinement network $H$, pre-trained correspondence estimation network $F$, current training iteration $T$, maximum training iteration $T_{max}$},

\State \textbf{Pre-Train} $G$ with the decomposition strategy of VLLVE (Alg.~\ref{alg:1}), update the network besides the learning module for $\boldsymbol{B}$
\While {$T < T_{max}$}
\State Sample training batch $(\I_{d,t_1}, \I_{n,t_1})$ and $(\I_{d,t_2}, \I_{n,t_2})$ 
\State Use the frozen correspondence estimation network $F$ to obtain the correspondence maps $\hat{\boldsymbol{M}}_{t_1 \rightarrow t_2}$, and $\hat{\boldsymbol{M}}_{t_2 \rightarrow t_1}$, and use them to obtain the correspondences coordinates ${ c_m=(x_{1,m}, y_{1,m}, x_{2,m}, y_{2,m}), m\in[1, M]}$
\State Use the correspondence refinement network $H$ to obtain the refined correspondence maps $\hat{\boldsymbol{M}'}_{t_1 \rightarrow t_2}$, and $\hat{\boldsymbol{M}'}_{t_2 \rightarrow t_1}$, and use them to obtain the correspondences coordinates ${ c'_m=(x'_{1,m}, y'_{1,m}, x'_{2,m}, y'_{2,m}), m\in[1, M]}$ with Eq.~\ref{eq:refine}
\State Feed the input of $\I_{d,t_1}$ and $\I_{d,t_2}$ into the enhancement network, obtain the hidden feature as $\boldsymbol{f}_{d,t_1}$ and $\boldsymbol{f}_{d,t_2}$
\State Conduct the interaction between these two features with Eqs.~\ref{eq:fusion1} and \ref{eq:fusion2}, obtaining $\hat{\boldsymbol{f}}_{d,t_1}$ and $\hat{\boldsymbol{f}}_{d,t_2}$
\State Obtain the enhancement results from $\hat{\boldsymbol{f}}_{d,t_1}$ and $\hat{\boldsymbol{f}}_{d,t_2}$ via the decoder as $\hat{\L}_{n,t_1}$, $\hat{\L}_{n,t_2}$, $\hat{\R}_{n,t_1}$, $\hat{\R}_{n,t_2}$, and $\hat{\boldsymbol{B}}_{n,t_1}$, $\hat{\boldsymbol{B}}_{n,t_2}$
\State Compute the loss term with Eq.~\ref{eq:video222} (including the loss terms in Eqs.~\ref{eq:1}, \ref{eq:2}, \ref{eq:3}, \ref{eq:smooth}, \ref{eq:22}, \ref{eq:33}, and \ref{eq:1-new}), updating network $G$
\State Compute loss with Eq.~\ref{eq:1-crn} to update network $H$, $T=T+1$
\EndWhile
\Ensure{Trained model $G$}
\end{algorithmic}
\end{algorithm}

\section{Experiments}

\subsection{Datasets}
In this section, experiments are conducted on several public datasets (within the sRGB domain), which contain a wide range of real-world videos with various motion patterns and degradations.
\begin{itemize}
	\item (1) SMID~\cite{chen2019seeing}: This dataset consists of static videos with ground truths obtained through long exposure. 
	\item (2) SDSD~\cite{wang2021sdsd}: SDSD is a dynamic video dataset containing both indoor and outdoor subsets. 
	\item (3) DID~\cite{fu2023dancing}: DID is another dynamic video dataset. 
    We follow DID's division into a training set, a test set, and a validation set, maintaining a ratio of 3:1:1.
	\item (4) DAVIS~\cite{pont20172017}: DAVIS is a dataset that includes both scene dynamics and camera motions. To simulate low-light degradation in DAVIS, we followed the methodology outlined in \cite{zhang2021learning}, further incorporating noise degradation.
\end{itemize}

\begin{table}[t]
    \scriptsize
	\centering
	\caption{Architecture of image encoder $M_E$.}
	\label{tab:enc_image}
    \vspace{-0.1in}
    \renewcommand{\arraystretch}{1.26}
    {
		\begin{tabular}{|c|c|c|c|c|c|}
            \hline
			Layer Type 		  & Activation & Kernel & Stride & Padding & Output Size \\ \hline \hline
			Input Image       	     & -       & -      & -   &-   & $H \times W \times 3$ \\ 
			Convolution 	  & LeakyReLU    & 3      & 1   &1   & $H \times W \times 64$ \\ 
			Convolution       & LeakyReLU    & 4      & 2   &1   & $\frac{H}{2} \times \frac{W}{2} \times 64$ \\ 
			Convolution       & LeakyReLU    & 4      & 2   &1   & $\frac{H}{4} \times \frac{W}{4} \times 64$ \\ \hline
	\end{tabular}}
\end{table}

\begin{table}[tb!]
    \scriptsize
	\centering
	\caption{Architecture of image decoder $M_D$.}
    \vspace{-0.1in}
    \renewcommand{\arraystretch}{1.26}
    {
		\begin{tabular}{|c|c|c|c|c|c|}
            \hline
			Layer Type 		 & Activation & Kernel & Stride & Padding & Output Size \\ \hline \hline
			Input Feature       	     & -       & -      & -   &-   & $\frac{H}{4} \times \frac{W}{4} \times 64$ \\ 
			Convolution 	  & LeakyReLU    & 3      & 1   &1   & $\frac{H}{4} \times \frac{W}{4} \times 256$ \\ 
			Pixel Shuffle     & - & - & 2& -& $\frac{H}{2} \times \frac{W}{2} \times 64$   \\ 
			Convolution 	  & LeakyReLU    & 3      & 1   &1   & $\frac{H}{2} \times \frac{W}{2} \times 256$ \\ 
			Pixel Shuffle      & - & - & 2& -& $H \times W \times 64$   \\ 
			Convolution 	  & LeakyReLU    & 3      & 1   &1   & $H \times W \times 64$ \\ 
			Convolution 	  & LeakyReLU    & 3      & 1   &1   & $H \times W \times 3$ \\ 
            \hline
	\end{tabular}}
	\label{tab:dec_image}
\end{table}

\begin{table}[t]
    \scriptsize
	\centering
	\caption{Architecture of the correspondence refinement network $H$, where the residual block follows the structure of ``Input + Convolution(Input)'', where the ``Convolution'' module consists of two same convolutional layers.}
	\label{tab:corr_refine}
    \vspace{-0.1in}
    \renewcommand{\arraystretch}{1.26}
	\resizebox{1.0\linewidth}{!}
    {
		\begin{tabular}{|c|c|c|c|c|c|}
            \hline
			Layer Type 		  & Activation & Kernel & Stride & Padding & Output Size \\ \hline \hline
			Input Image       	     & -       & -      & -   &-   & $H \times W \times 10$ \\ 
			Convolution 	  & LeakyReLU    & 3      & 1   &1   & $H \times W \times 64$ \\ 
			Convolution       & LeakyReLU    & 4      & 2   &1   & $\frac{H}{2} \times \frac{W}{2} \times 128$ \\ 
			Convolution       & LeakyReLU    & 4      & 2   &1   & $\frac{H}{4} \times \frac{W}{4} \times 256$ \\ \hline
\hline
Residual Block  & ReLU       & 3      & 1   &1   & $\frac{H}{4} \times \frac{W}{4} \times 256$ \\ 
Convolution 	  & LeakyReLU    & 3      & 1   &1   & $\frac{H}{4} \times \frac{W}{4} \times 512$ \\ 
Pixel Shuffle     & - & - & 2& -& $\frac{H}{2} \times \frac{W}{2} \times 128$   \\ 
Residual Block  & ReLU       & 3      & 1   &1   & $\frac{H}{2} \times \frac{W}{2} \times 128$ \\ 
Convolution 	  & LeakyReLU    & 3      & 1   &1   & $\frac{H}{2} \times \frac{W}{2} \times 256$ \\ 
Pixel Shuffle      & - & - & 2& -& $H \times W \times 64$   \\ 
Residual Block  & ReLU       & 3      & 1   &1   & $H \times W \times 64$ \\ 
Convolution 	  & LeakyReLU    & 3      & 1   &1   & $H \times W \times 64$ \\ 
Residual Block  & ReLU       & 3      & 1   &1   & $H \times W \times 64$ \\ 
Convolution 	  & LeakyReLU    & 3      & 1   &1   & $H \times W \times 4$ \\ 
            \hline
            
	\end{tabular}}
\end{table}

\begin{table}[t]
	\centering
	\caption{Quantitative comparison on SDSD, SMID, and DID datasets. Our method performs the best consistently.} 
    \vspace{-0.1in}
	\resizebox{1.0\linewidth}{!}
    {
        \begin{tabular}{|l|cc|cc|cc|cc|}
            \hline
			& \multicolumn{2}{c|}{SDSD-indoor} &\multicolumn{2}{c|}{SDSD-outdoor}& \multicolumn{2}{c|}{SMID} &\multicolumn{2}{c|}{DID}  \\
            \hline
			Methods & PSNR & SSIM& PSNR & SSIM & PSNR & SSIM& PSNR & SSIM\\
			\hline \hline
			SNR~\cite{xu2022snr}    &27.30 &0.84 &23.23 & 0.82& 28.49 &0.81 &24.85 &0.90 \\
			SMG~\cite{xu2023low}    &27.82 &0.86 & 25.17&0.81 &28.03 &0.78 &25.14 &0.85 \\
			PairLLE~\cite{fu2023learning}    &23.48 &0.71 &20.04 &0.65 &22.70 &0.63 &22.56 &0.82 \\
			RetinexFormer~\cite{cai2023retinexformer}     & 26.56& 0.79 &22.80 &0.77 &29.15 &0.82 & 25.40&0.89 \\
			\hline
			MBLLEN~\cite{lv2018mbllen} & 22.17 & 0.66 &  21.41 & 0.63 & 22.67 & 0.68 & 24.22 & 0.86\\
			SMID~\cite{chen2019seeing} & 24.84 & 0.72 & 23.30&0.67 & 24.78 &0.72 & 22.28&0.84\\
			SMOID~\cite{jiang2019learning} & 24.63 & 0.70 & 22.25 & 0.68 & 23.64 & 0.71&22.13 &0.85\\
			SDSDNet~\cite{wang2021sdsd} & 26.81 & 0.75 & 23.08 & 0.71 & 26.03 &0.75 & 22.52 & 0.81\\
			DP3DF~\cite{xu2023deep} & 27.63 & 0.74 & 23.85 & 0.73 & 27.19 & 0.76 & 22.39 & 0.88\\
			StableLLVE~\cite{zhang2021learning} & 25.32 & 0.70 & 22.47 & 0.65 & 24.37 & 0.69 & 21.64 & 0.80\\
			LLVE-SEG~\cite{liu2023low} & 26.19 & 0.77 & 24.09 & 0.72 & 25.31 & 0.74 & 23.85 & 0.82\\
			LLVE-CFA~\cite{chhirolya2022low} & 24.28 &  0.81 & 22.64 & 0.76 & 24.72 & 0.70 & 22.53 & 0.87\\
			\hline
			BLLRVE~\cite{zhang2024binarized}&26.02 & 0.74&23.50&0.71&26.15&0.73	&23.25& 0.81\\
			VRGSS~\cite{li2023simple}&27.81 &0.76&24.28& 0.75&27.84&0.78&	24.51&0.87\\
			ZRLLE-PQP~\cite{wang2024zero}&25.53& 0.71&22.42&0.73&26.36&0.77	&22.84&0.80\\
			JDAE~\cite{shi2024zero}&26.19&0.74&23.37&0.76&26.45&0.76	&23.53&0.82\\
			\hline  
            VLLVE~\cite{xu2025low}& \underline{28.93}&\underline{0.88} &\underline{26.32} & \underline{0.82}& \underline{29.60} &\underline{0.82} & \underline{30.10}&\underline{0.93} \\
            VLLVE++ &\textbf{29.78} &\textbf{0.90} &\textbf{27.47} &\textbf{0.85} &\textbf{30.71} &\textbf{0.84} &\textbf{31.06} &\textbf{0.95}\\
            \hline
	\end{tabular}}
	\label{comparison5}
\end{table}

\vspace{-0.1in}
\subsection{Implementation Details}

\vspace{-0.1in}
\minisection{Our Framework's Structure}
Our framework consists of frame encoder $M_E$, frame decoder $M_D$ (the decoder for view-dependent branch, view-independent, and residual learning branch is the same), and the proposed CFIM.
The encoder follows a structure of convolution and downsampling, while the decoder consists of convolution and upsampling. 
Suppose the size of the input frame is $H\times W$, $M_E$ consists of different convolutional layers, as shown in \Cref{tab:enc_image}; $M_D$ contains convolutional layers and uses pixel-shuffle for upsampling, as exhibited in \Cref{tab:dec_image}.

CFIM consists of one transformer block for self-attention computation, one transformer block for cross-attention, and one ResNet block to conduct spatial-temporal short-range fusion.

Moreover, the correspondence refinement network $H$ is implemented as a simple convolutional encoder-decoder architecture for efficiency. Its input consists of two frames and the initial bidirectional correspondence maps (share the same spatial size as the frames). 
Therefore, the input dimension is $3 \times 2 + 2 \times 2 = 10$, and the output dimension is $2\times 2=4$. The detailed architecture is provided in Table~\ref{tab:corr_refine}.

\vspace{-0.1in}
\minisection{Experimental Details}
We conducted experiments on all datasets using the same network structure. 
$T$ is set as 5.
All modules were trained end-to-end, with the learning rate initialized at $4e^{-4}$ for all layers, adapted by a cosine learning scheduler. The batch size used was 4. Correspondences were computed using DKM~\cite{edstedt2023dkm}. 
The experiments in the ablation study would employ other correspondence estimation networks, showing that our method is not sensitive to the choice of correspondence estimation models.
Additionally, no data augmentation was utilized.
Our method was implemented using Python 3.7.16 and PyTorch 1.9.1~\cite{paszke2019pytorch}, and all experiments were executed on a single NVidia A100 GPU. 
We employed Kaiming Initialization~\cite{he2015delving} to initialize the weights and used the Adam optimizer~\cite{kingma2014adam} for training with a momentum value of 0.9. 
We set hyperparameters $\lambda_1=1$, $\lambda_2=1$, and $\lambda_3=1$ for experiments on all datasets.
Furthermore, all components in VLLVE++ are trained end-to-end using identical hyper-parameters across the overall framework.
\textbf{The code and trained models of VLLVE++ will be released upon publication}.

\begin{figure*}[t]
	\centering
	\begin{subfigure}[c]{0.16\textwidth}
		\centering
		\includegraphics[width=1.15in]{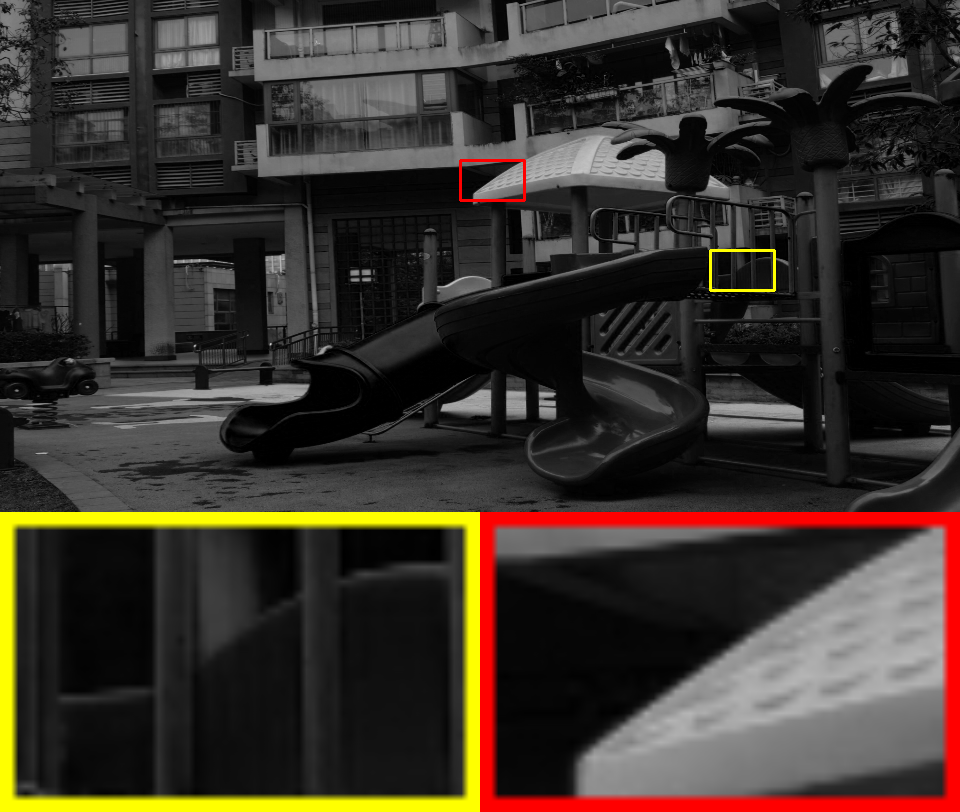}
		\caption*{Input}
	\end{subfigure}
	\begin{subfigure}[c]{0.16\textwidth}
		\centering
		\includegraphics[width=1.15in]{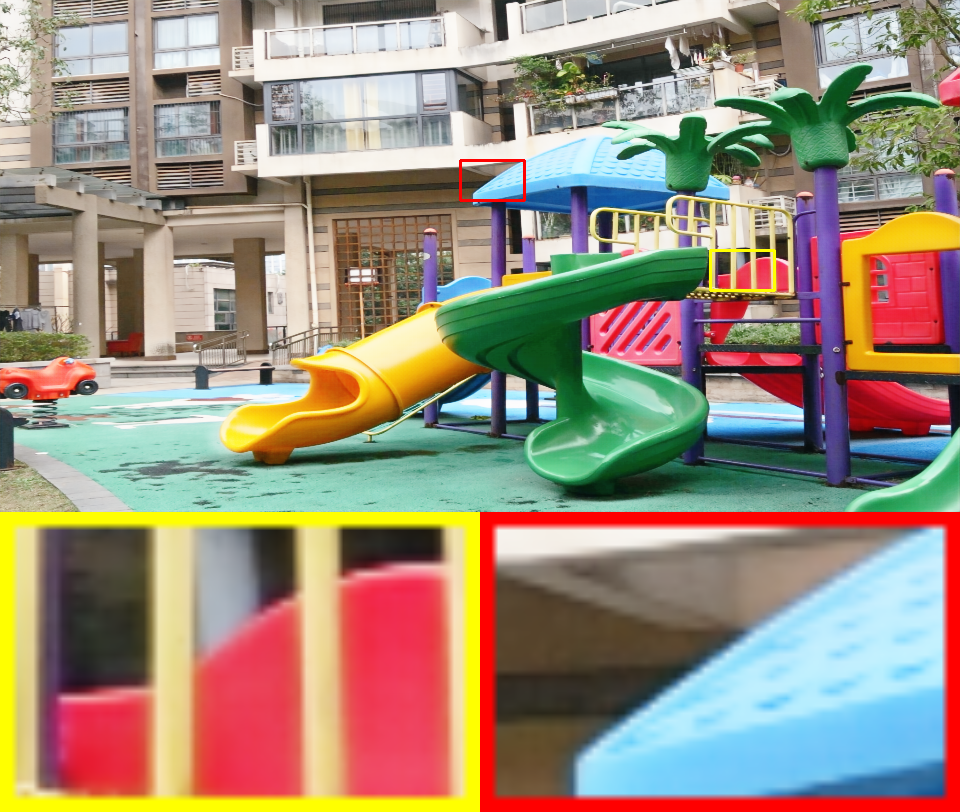}
		\caption*{RetinexFormer}
	\end{subfigure}
	\begin{subfigure}[c]{0.16\textwidth}
		\centering
		\includegraphics[width=1.15in]{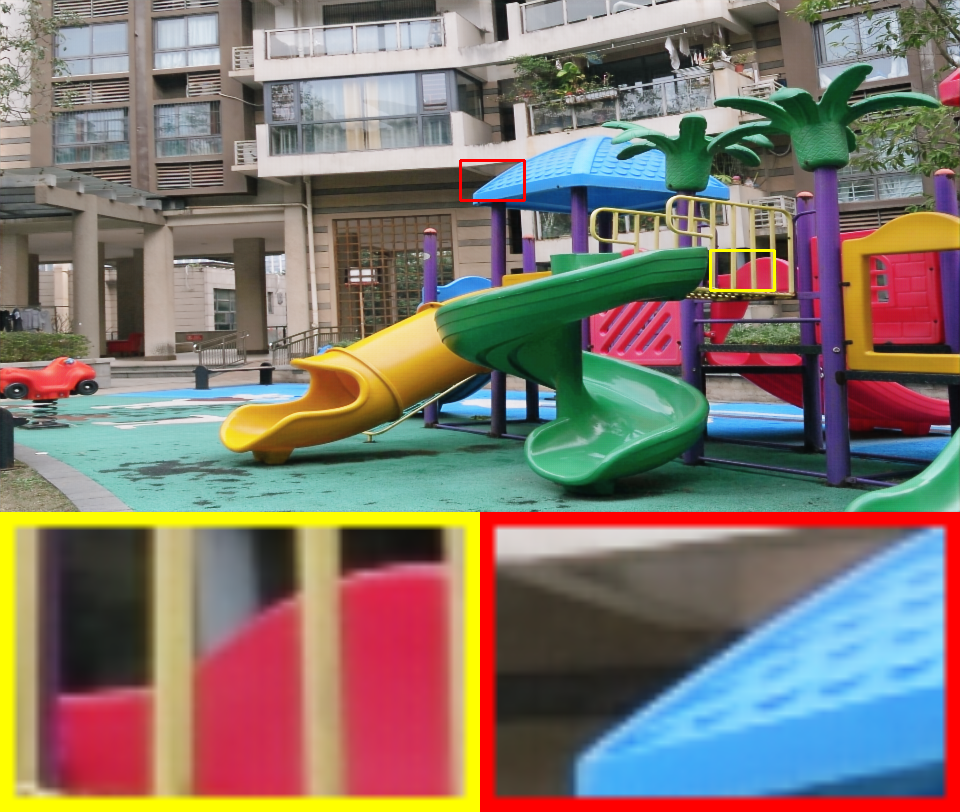}
		\caption*{DP3DF}
	\end{subfigure}
    	\begin{subfigure}[c]{0.16\textwidth}
		\centering
		\includegraphics[width=1.15in]{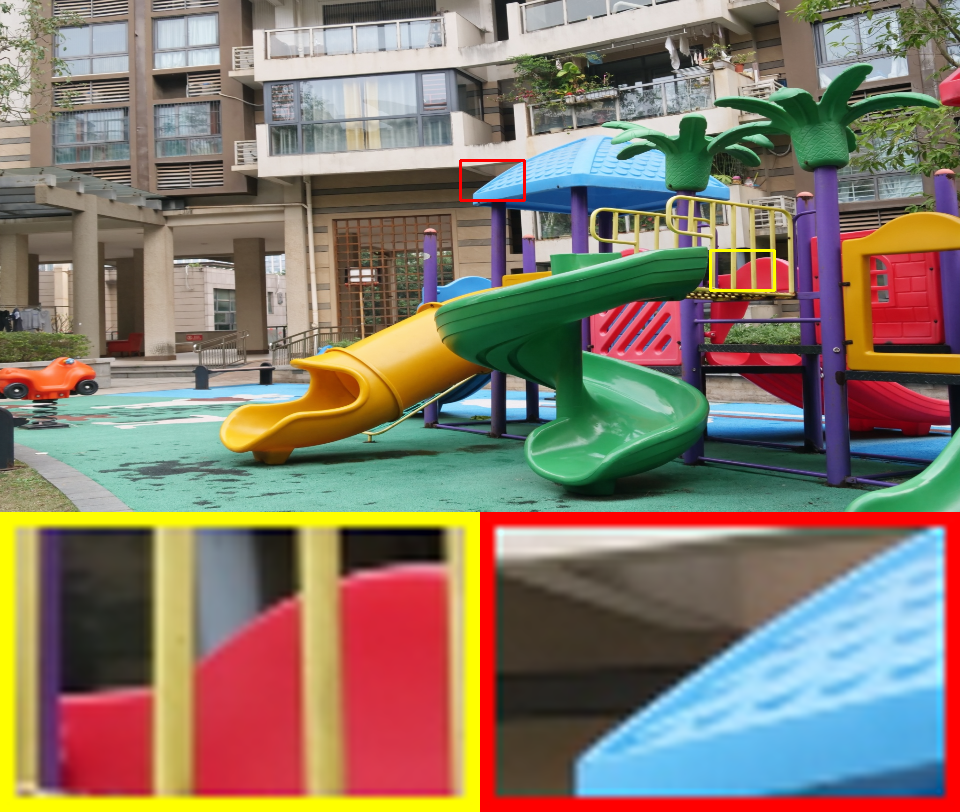}
		\caption*{VLLVE}
	\end{subfigure}
	\begin{subfigure}[c]{0.16\textwidth}
		\centering
		\includegraphics[width=1.15in]{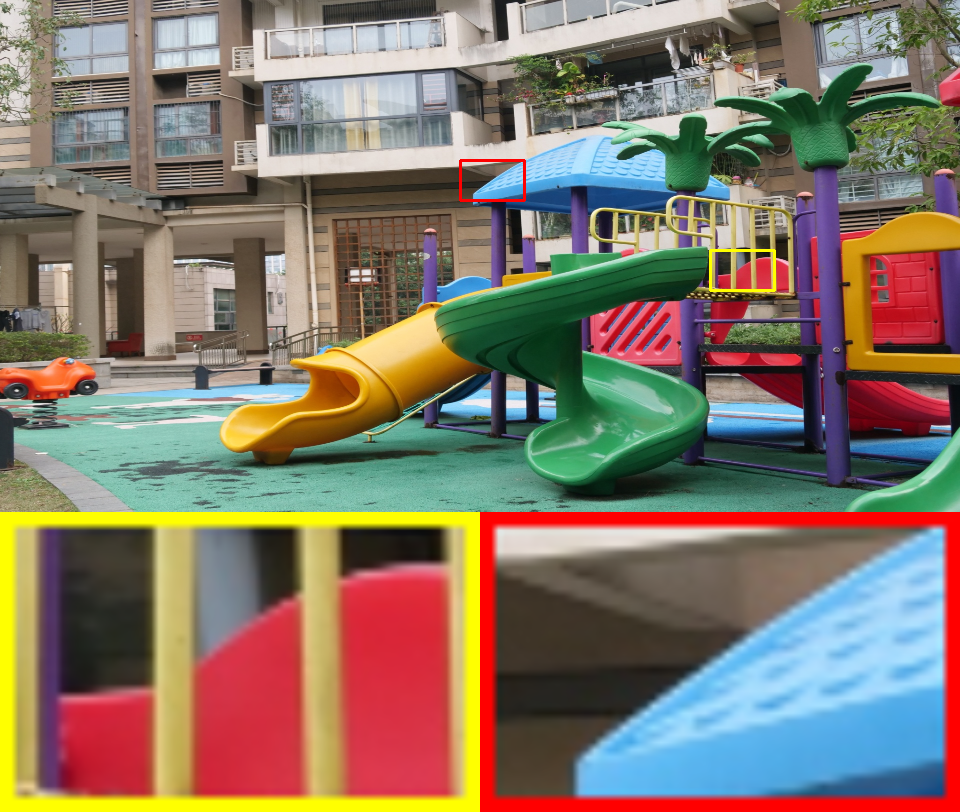}
		\caption*{VLLVE++}
	\end{subfigure} 
	\begin{subfigure}[c]{0.16\textwidth}
		\centering
		\includegraphics[width=1.15in]{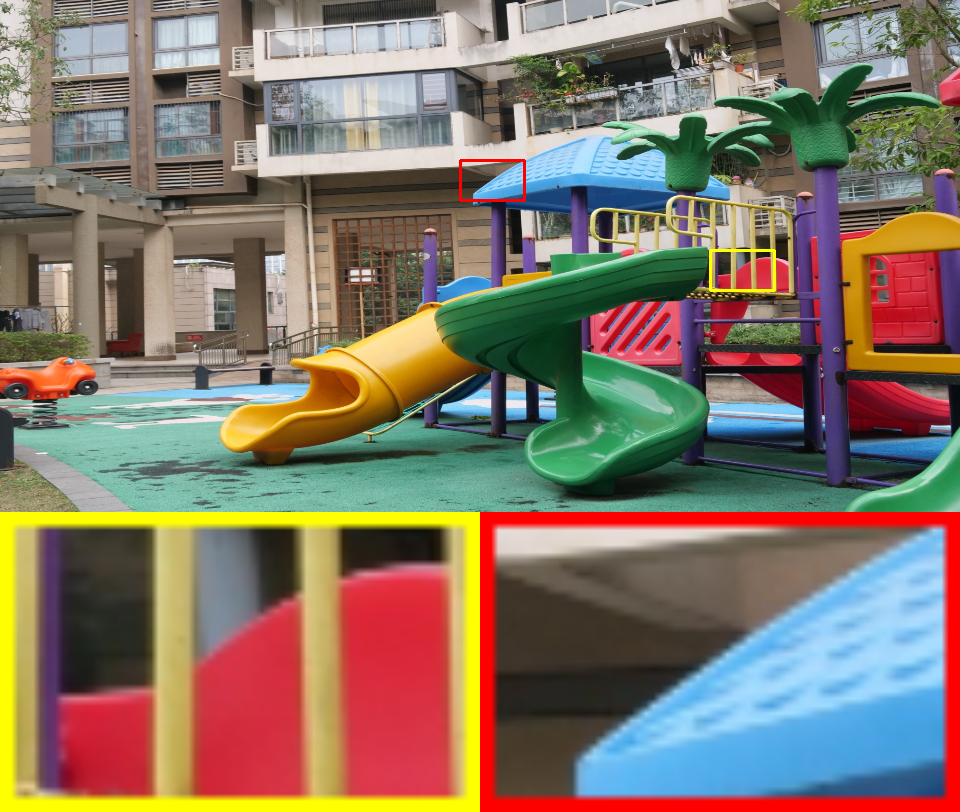}
		\caption*{GT}
	\end{subfigure} \vspace{0.5em}  \\

	\begin{subfigure}[c]{0.16\textwidth}
		\centering
		\includegraphics[width=1.15in]{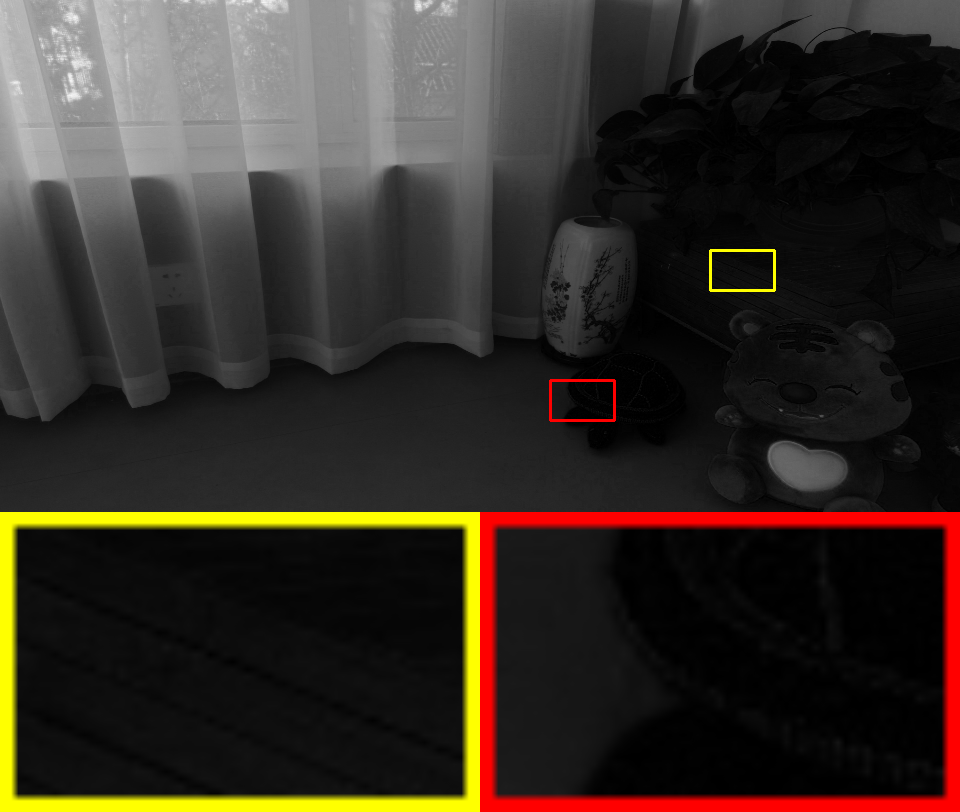}
		\caption*{Input}
	\end{subfigure}
	\begin{subfigure}[c]{0.16\textwidth}
		\centering
		\includegraphics[width=1.15in]{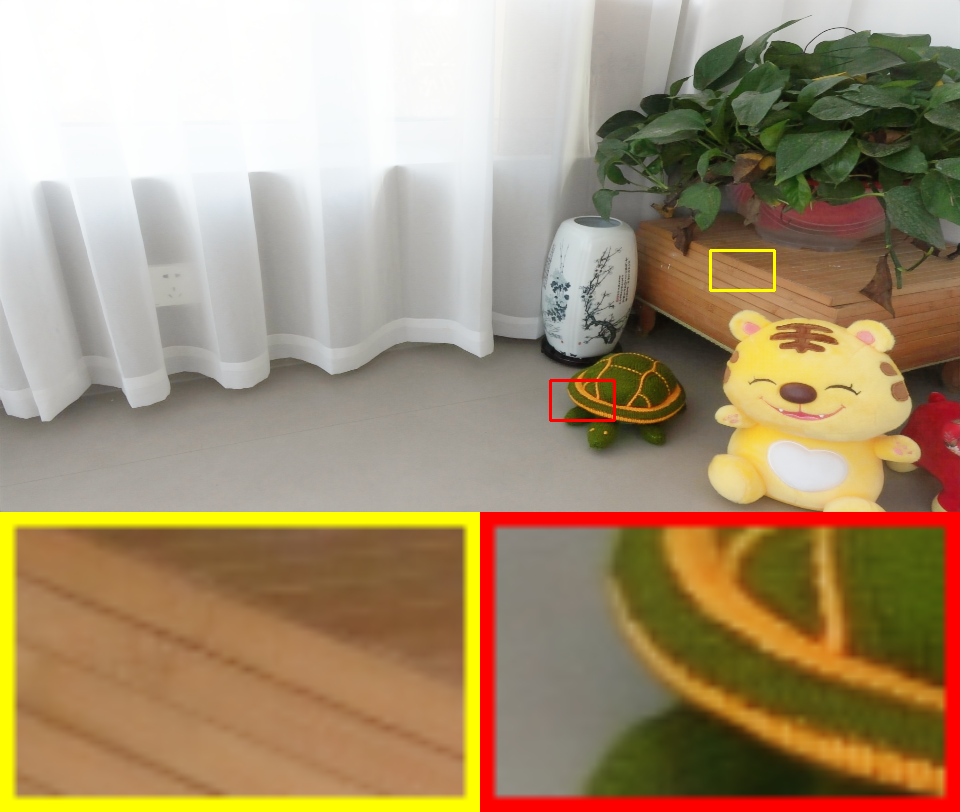}
		\caption*{RetinexFormer}
	\end{subfigure}
	\begin{subfigure}[c]{0.16\textwidth}
		\centering
		\includegraphics[width=1.15in]{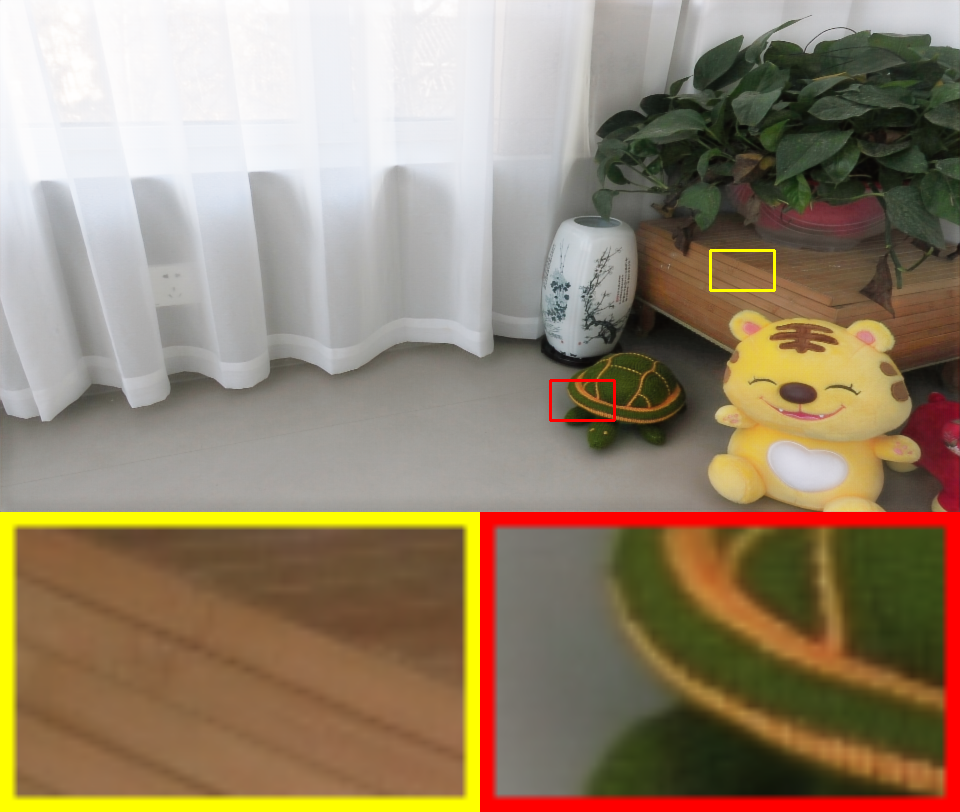}
		\caption*{DP3DF}
	\end{subfigure}
    	\begin{subfigure}[c]{0.16\textwidth}
		\centering
		\includegraphics[width=1.15in]{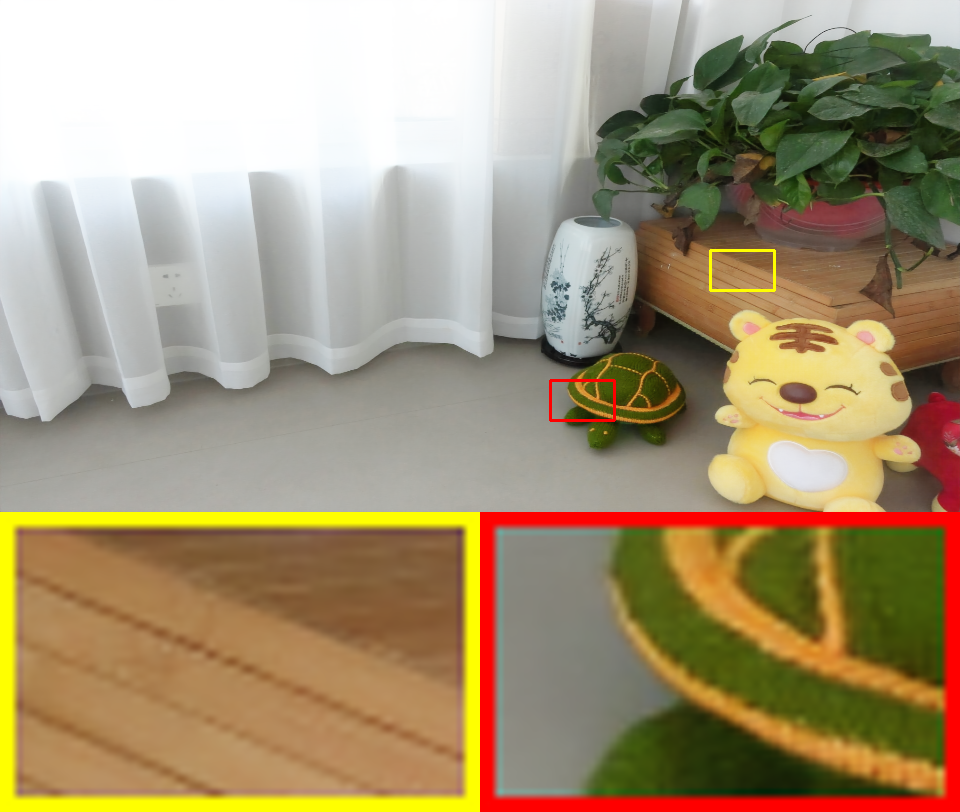}
		\caption*{VLLVE}
	\end{subfigure}
	\begin{subfigure}[c]{0.16\textwidth}
		\centering
		\includegraphics[width=1.15in]{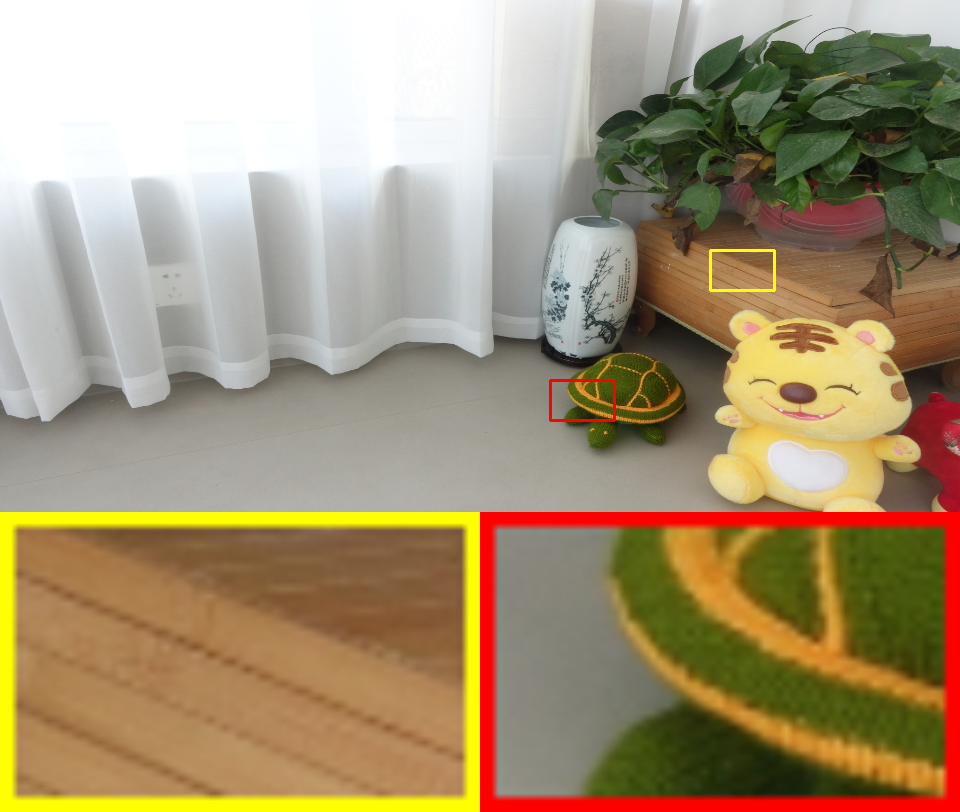}
		\caption*{VLLVE++}
	\end{subfigure} 
	\begin{subfigure}[c]{0.16\textwidth}
		\centering
		\includegraphics[width=1.15in]{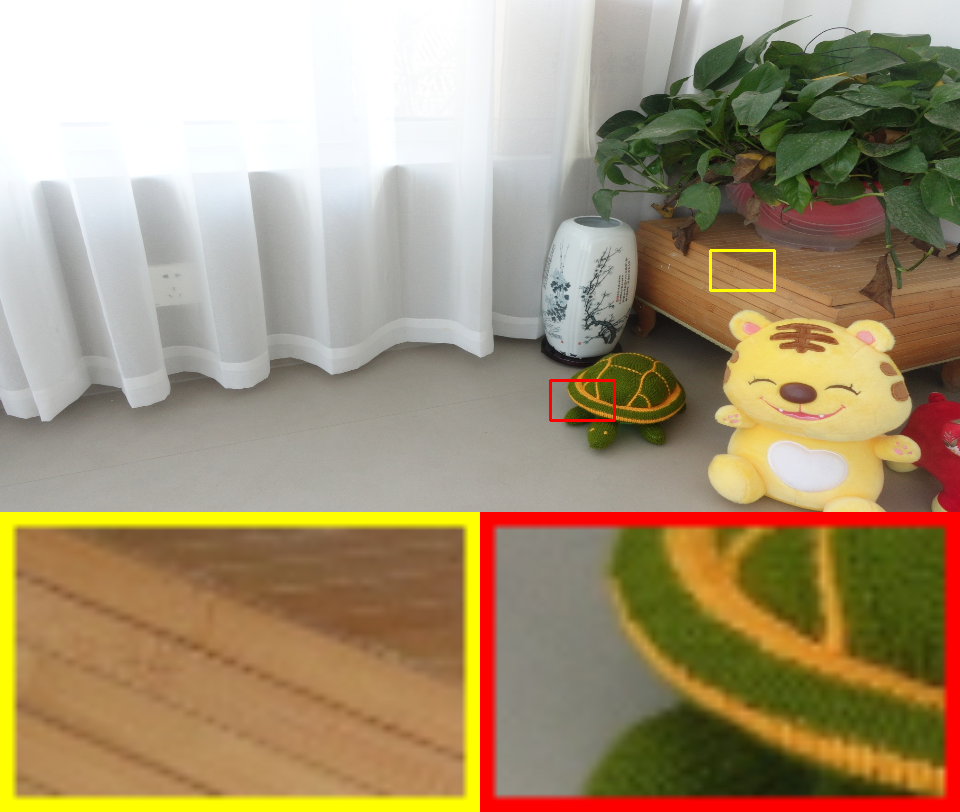}
		\caption*{GT}
	\end{subfigure}  \vspace{0.5em} \\

	\begin{subfigure}[c]{0.16\textwidth}
		\centering
		\includegraphics[width=1.15in]{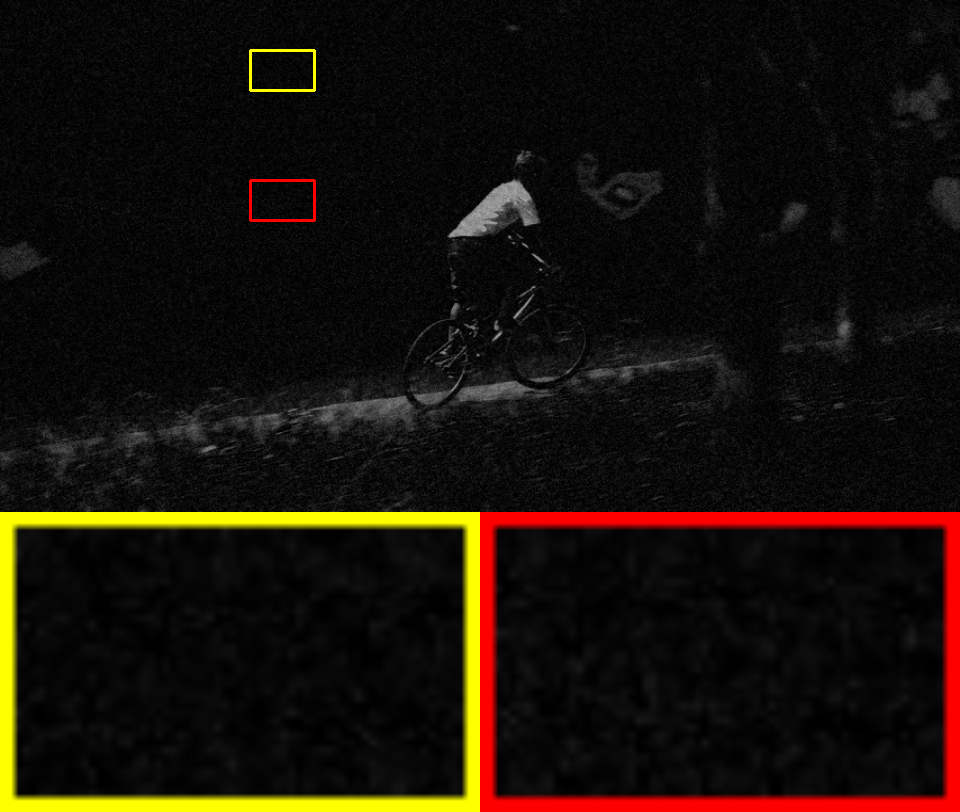}
		\caption*{Input}
	\end{subfigure}
	\begin{subfigure}[c]{0.16\textwidth}
		\centering
		\includegraphics[width=1.15in]{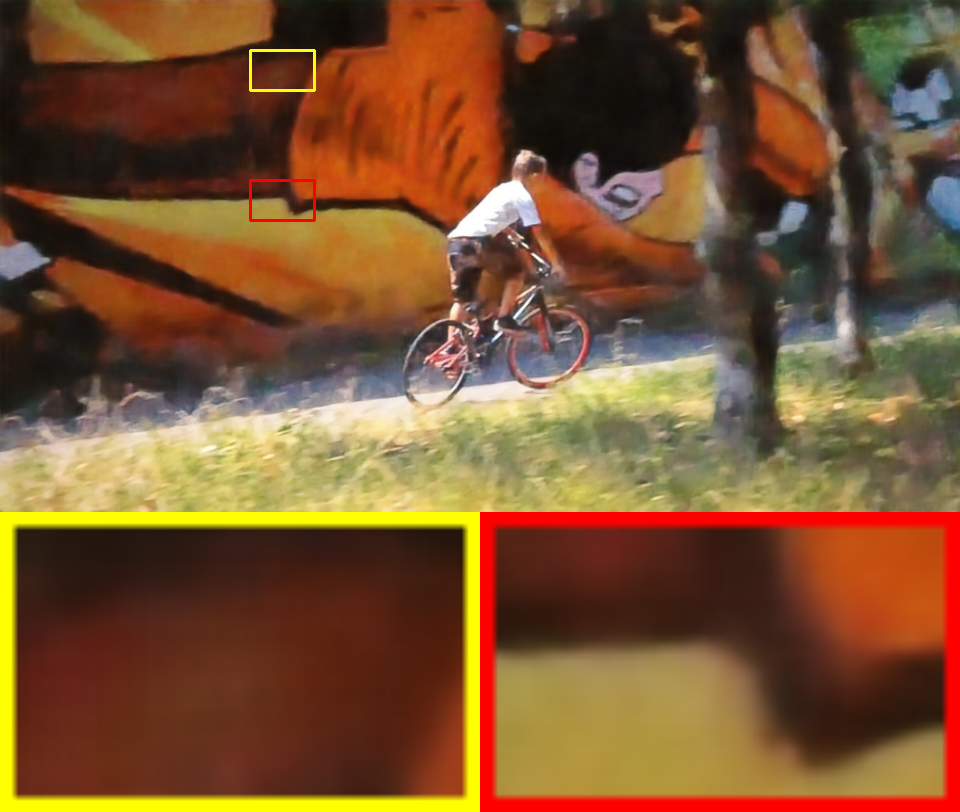}
		\caption*{RetinexFormer}
	\end{subfigure}
	\begin{subfigure}[c]{0.16\textwidth}
		\centering
		\includegraphics[width=1.15in]{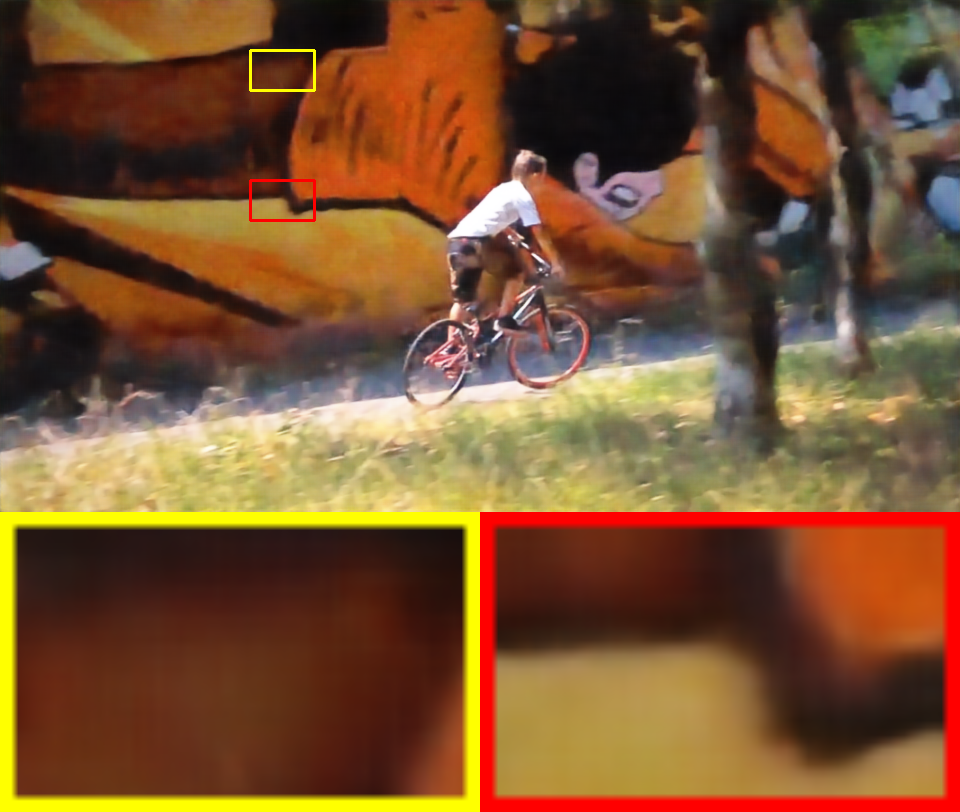}
		\caption*{DP3DF}
	\end{subfigure}
    	\begin{subfigure}[c]{0.16\textwidth}
		\centering
		\includegraphics[width=1.15in]{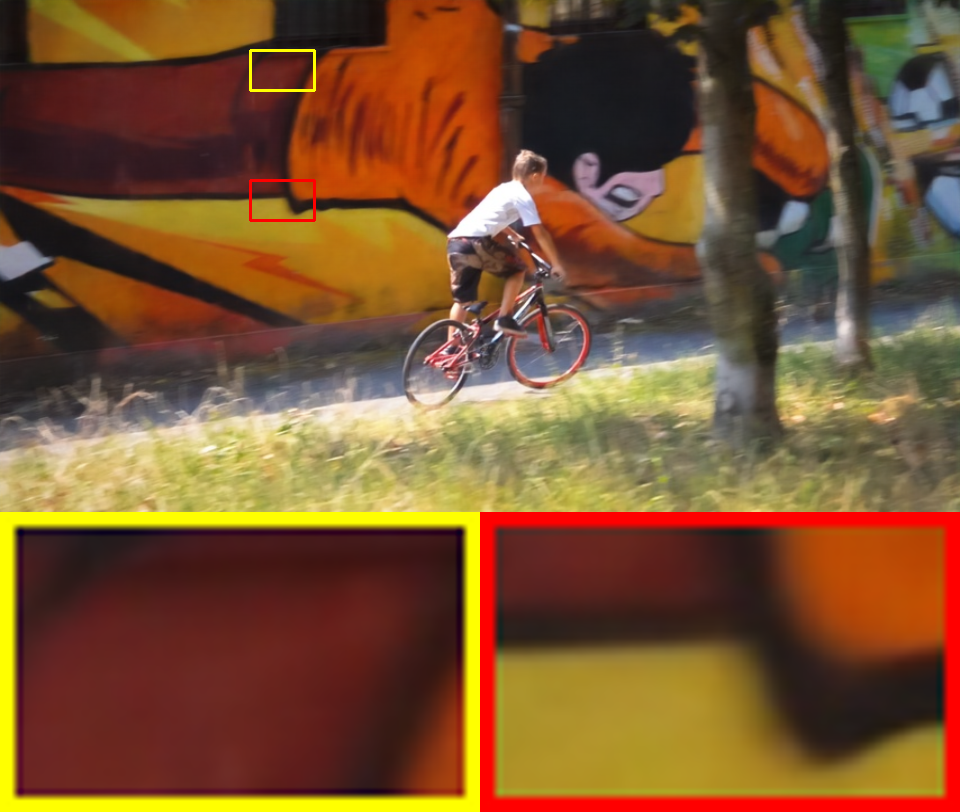}
		\caption*{VLLVE}
	\end{subfigure}
	\begin{subfigure}[c]{0.16\textwidth}
		\centering
		\includegraphics[width=1.15in]{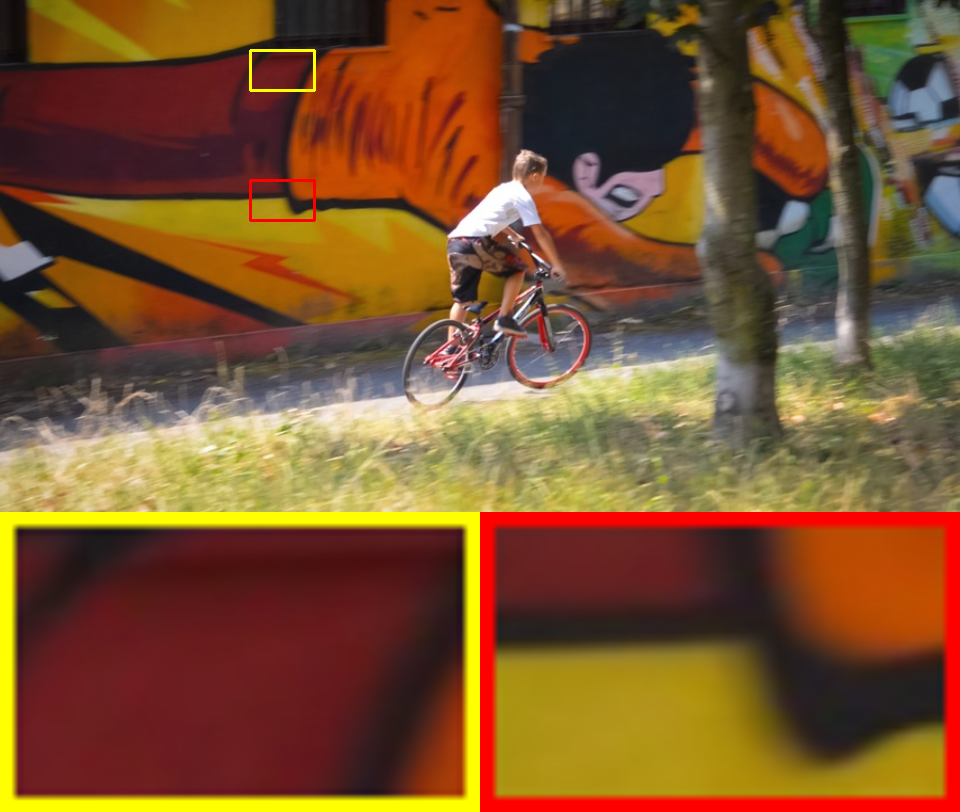}
		\caption*{VLLVE++}
	\end{subfigure} 
	\begin{subfigure}[c]{0.16\textwidth}
		\centering
		\includegraphics[width=1.15in]{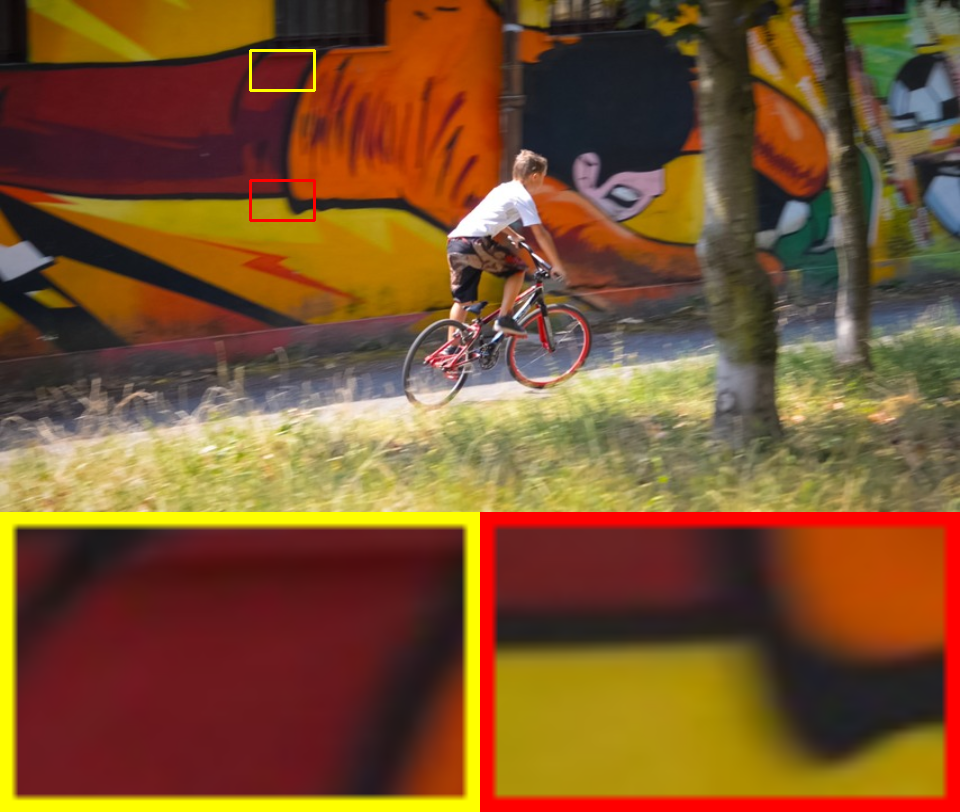}
		\caption*{GT}
	\end{subfigure} \vspace{0.5em} \\

	\begin{subfigure}[c]{0.16\textwidth}
		\centering
		\includegraphics[width=1.15in]{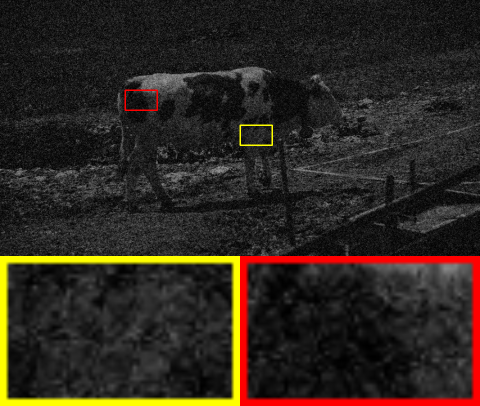}
		\caption*{Input}
	\end{subfigure}
	\begin{subfigure}[c]{0.16\textwidth}
		\centering
		\includegraphics[width=1.15in]{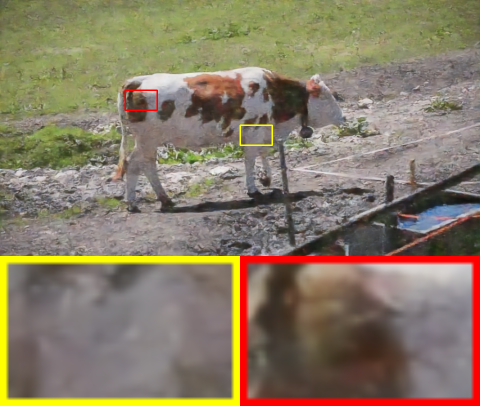}
		\caption*{RetinexFormer}
	\end{subfigure}
	\begin{subfigure}[c]{0.16\textwidth}
		\centering
		\includegraphics[width=1.15in]{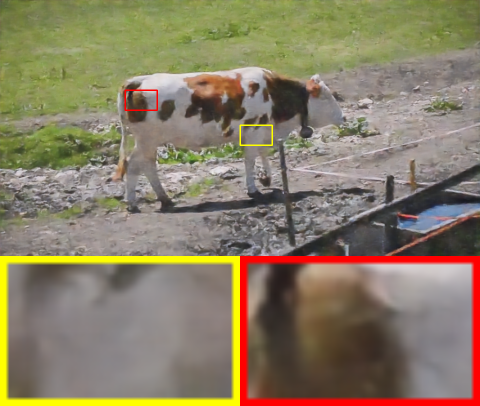}
		\caption*{DP3DF}
	\end{subfigure}
    	\begin{subfigure}[c]{0.16\textwidth}
		\centering
		\includegraphics[width=1.15in]{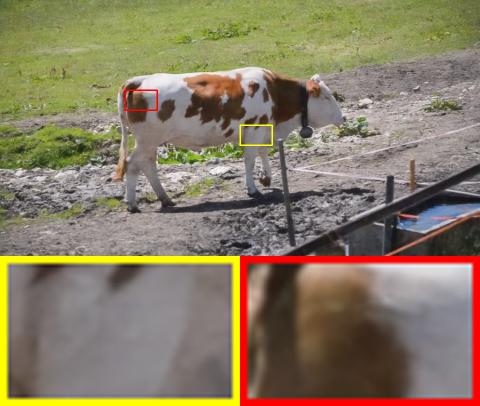}
		\caption*{VLLVE}
	\end{subfigure}
	\begin{subfigure}[c]{0.16\textwidth}
		\centering
		\includegraphics[width=1.15in]{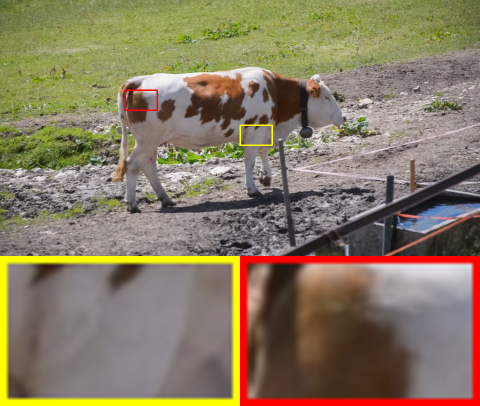}
		\caption*{VLLVE++}
	\end{subfigure} 
	\begin{subfigure}[c]{0.16\textwidth}
		\centering
		\includegraphics[width=1.15in]{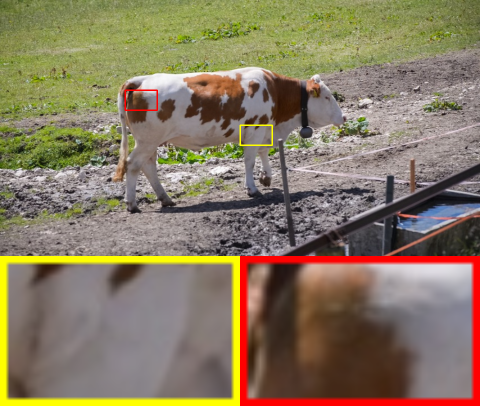}
		\caption*{GT}
	\end{subfigure} \\
	    \vspace{-0.1in}
	\caption{Comparisons on DID (top two rows) and DAVIS (bottom two rows). Our method has less noise, better visibility and details.}
	\label{fig:cmp2}
\end{figure*}

\subsection{Comparison and Baselines}
We focus on enhancement in the sRGB domain, which aligns with common real-world requirements, such as improving videos from online platforms.
We conducted a comprehensive comparison with SOTA LLVE methods, including MBLLEN~\cite{lv2018mbllen}, SMID~\cite{chen2019seeing}, SMOID~\cite{jiang2019learning}, SDSDNet~\cite{wang2021sdsd}, DP3DF~\cite{xu2023deep}, StableLLVE~\cite{zhang2021learning}, LLVE-SEG~\cite{liu2023low},  LLVE-CFA~\cite{chhirolya2022low}, BLLRVE~\cite{zhang2024binarized}, VRGSS~\cite{li2023simple}, ZRLLE-PQP~\cite{wang2024zero}, JDAE~\cite{shi2024zero}, and VLLVE~\cite{xu2025low}.
We also compared our strategy with SOTA LLIE methods for individual frames, including SNR~\cite{xu2022snr}, SMG~\cite{xu2023low}, PairLLE~\cite{fu2023learning}, RetinexFormer~\cite{cai2023retinexformer}. All these methods were trained on each dataset employing their respective released code and hyper-parameters (e.g., the training epoch that is set to guarantee convergence).
\textit{Note that all baselines are trained on our unified data split for fair comparisons. The splits slightly differ from those in the baselines' original papers, so the scores of baselines might vary from the original papers}.

\begin{table}[tb!]
    \scriptsize
	\centering
	\caption{Quantitative comparison on the DAVIS dataset.}
        \vspace{-0.1in}
	\resizebox{1.02\linewidth}{!}
    {
		\begin{tabular}{|l|cccccc|}
            \hline
            &SNR&SMG&PairLLE&RetinexFormer&SMOID&SMID\\
			\hline \hline
			PSNR &20.69 &20.18 & 19.57&21.79 & 20.66&20.51\\
			SSIM&0.710 & 0.672&0.667 &0.723 &0.697 & 0.686\\
			\hline \hline
			&SDSDNet&DP3DF&StableLLVE&LLVE-SEG&VLLVE&VLLVE++\\
			\hline
			PSNR & 21.22&22.04 &21.48 &21.45 &\underline{23.38}&\textbf{24.09}\\
			SSIM &0.716 & 0.749&0.732 &0.715 &\underline{0.782}&\textbf{0.791} \\
            \hline
	\end{tabular}}
	\label{comparison2}
\end{table}

\begin{table}[tb!]
    \large
	\centering
	\caption{The comparison between our framework and current SOTA methods on SMID, SDSD indoor, SDSD outdoor, and DID datasets, in terms of short-term (``Short'') and long-term (``Long'') temporal loss. 
    }
        \vspace{-0.1in}
    \renewcommand{\arraystretch}{1.06}
	\resizebox{1.02\linewidth}{!}
    {
		\begin{tabular}{|l|cc|cc|cc|cc|}
            \hline
			& \multicolumn{2}{c|}{DAVIS} &\multicolumn{2}{c|}{SDSD-Indoor}&\multicolumn{2}{c|}{SDSD-Outdoor}& \multicolumn{2}{c|}{DID}\\
			\hline 
			Methods & Short & Long& Short & Long& Short & Long& Short & Long \\
			\hline \hline
			SNR &0.027 & 0.070& 0.017&0.060 &0.022 &0.046 & 0.025&0.068 \\
			RetinexFormer &0.029 &0.072 &0.018 &0.061 &0.024 &0.043 & 0.024&0.065 \\
			SDSDNet &0.024 &0.068 & 0.012&0.053 &0.011 &0.038 &0.017 &0.059 \\
			DP3DF & 0.021 &0.065 &\textbf{0.011} &\textbf{0.050} &0.013 & 0.036&0.019 &0.062 \\
			StableLLVE &0.023 &0.067 &0.016 &0.055 &0.019 &0.042 &0.023 &0.066 \\
			LLVE-SEG &0.025 &0.071 &0.014 & 0.057&0.016 &0.040 & 0.021&0.064 \\
			\hline \hline
            VLLVE&\underline{0.020} &\underline{0.063} &\underline{0.012}&\underline{0.051}&\underline{0.010}&\underline{0.034} &\underline{0.015} &\underline{0.054}  \\
            VLLVE++ &\textbf{0.014} &\textbf{0.051} & \textbf{0.008}&\textbf{0.045} &\textbf{0.006} &\textbf{0.027} &\textbf{0.009} &\textbf{0.046}\\
            \hline
	\end{tabular}}
	\label{comparison5-temporal}
\end{table}

\minisection{Quantitative Result}
In \Cref{comparison5}, we display the comparisons with the selected baseline methods across the SDSD, SMID, and DID datasets.
The table reveals that our \textbf{VLLVE} and \textbf{VLLVE++} {\em consistently\/} outperform all other methods.
Notably, our scores exhibit a significant lead over all others, particularly on DID, which is a large-scale dynamic video dataset. This superiority underscores the stable capability of our method in enhancing real-world videos.
Meanwhile, VLLVE++ consistently outperforms VLLVE, owing to its enhanced decomposition strategy with residual terms and the newly introduced bidirectional refinement strategy for correspondences.
The enhanced decomposition strategy can handle a wider range of difficult-to-model degradations, while the correspondence refinement strategy further improves decomposition learning.

Furthermore, \Cref{comparison2} provides the comparative results on DAVIS. Compared to the degradation synthesis as adopted in \cite{zhang2021learning}, we further include the degradation of a low-light noise term. 
As illustrated in \Cref{comparison2}, our method VLLVE consistently yields the highest PSNR and SSIM values among the comparisons with baselines. 
On the other hand, VLLVE++ continues to outperform VLLVE, demonstrating the effectiveness of its new components.

\begin{figure*}[t]
	\centering
	\begin{subfigure}[c]{0.16\textwidth}
		\centering
		\includegraphics[width=1.15in]{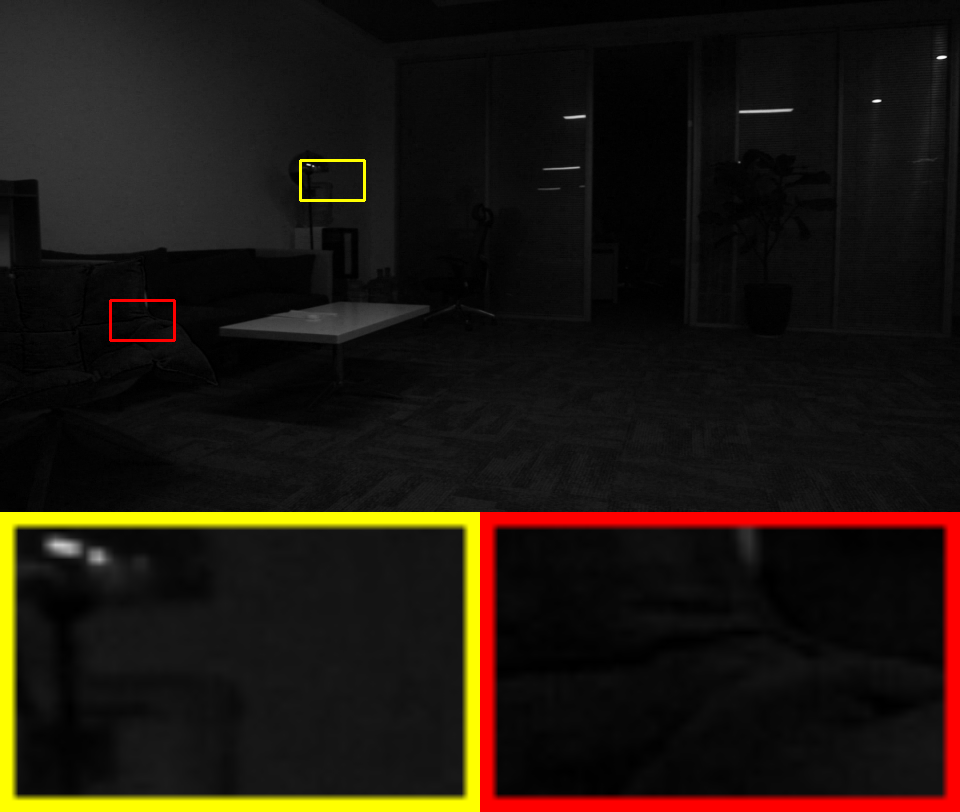}
		\caption*{Input}
	\end{subfigure}
	\begin{subfigure}[c]{0.16\textwidth}
		\centering
		\includegraphics[width=1.15in]{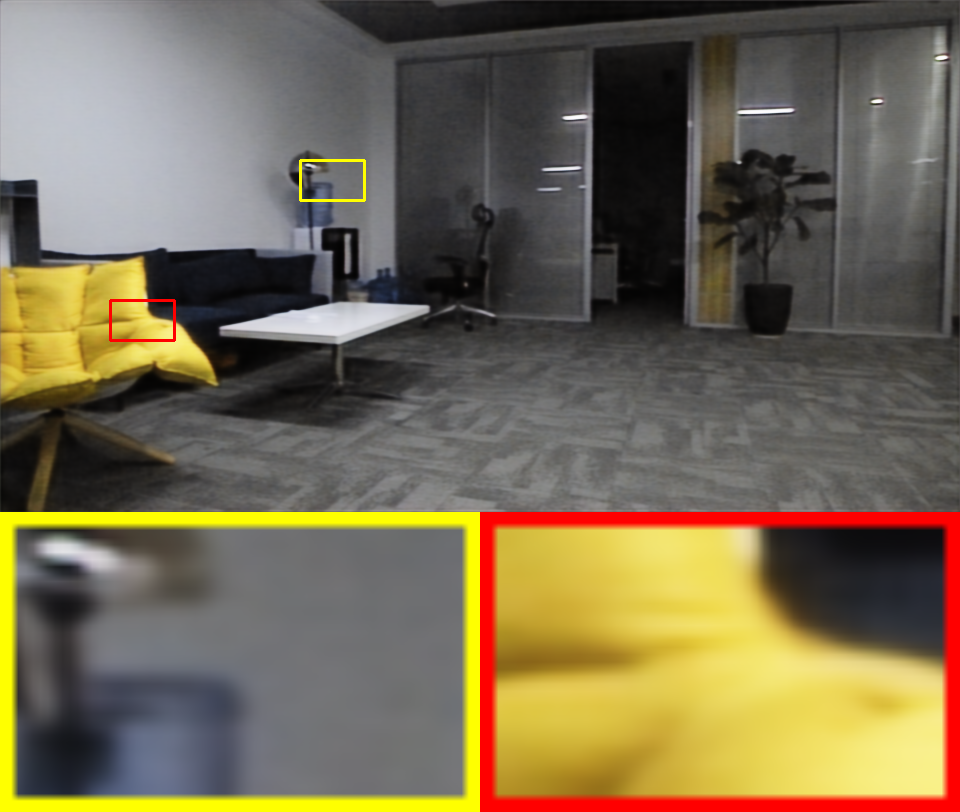}
		\caption*{RetinexFormer}
	\end{subfigure}
	\begin{subfigure}[c]{0.16\textwidth}
		\centering
		\includegraphics[width=1.15in]{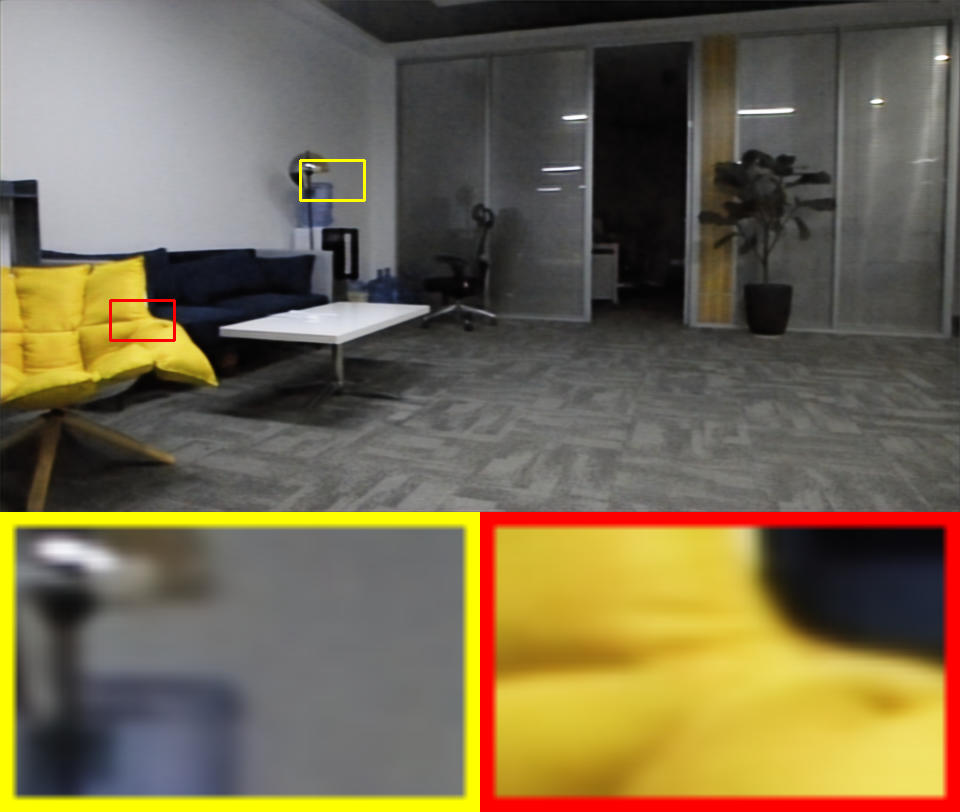}
		\caption*{DP3DF}
	\end{subfigure}
    \begin{subfigure}[c]{0.16\textwidth}
		\centering
		\includegraphics[width=1.15in]{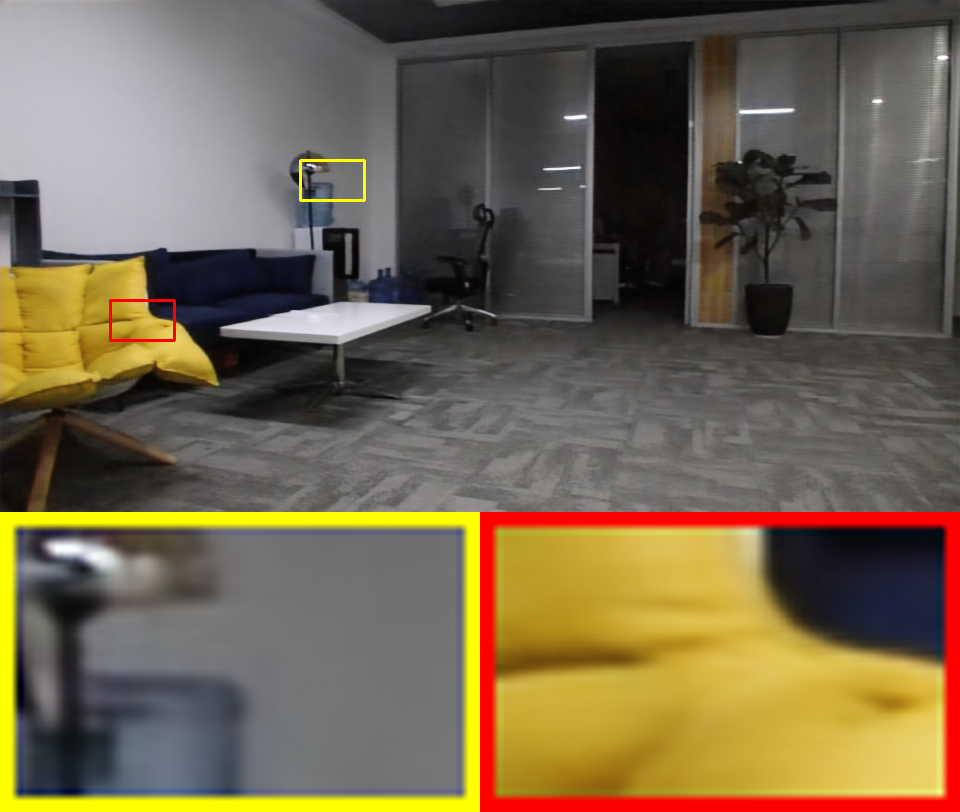}
		\caption*{VLLVE}
	\end{subfigure}
	\begin{subfigure}[c]{0.16\textwidth}
		\centering
		\includegraphics[width=1.15in]{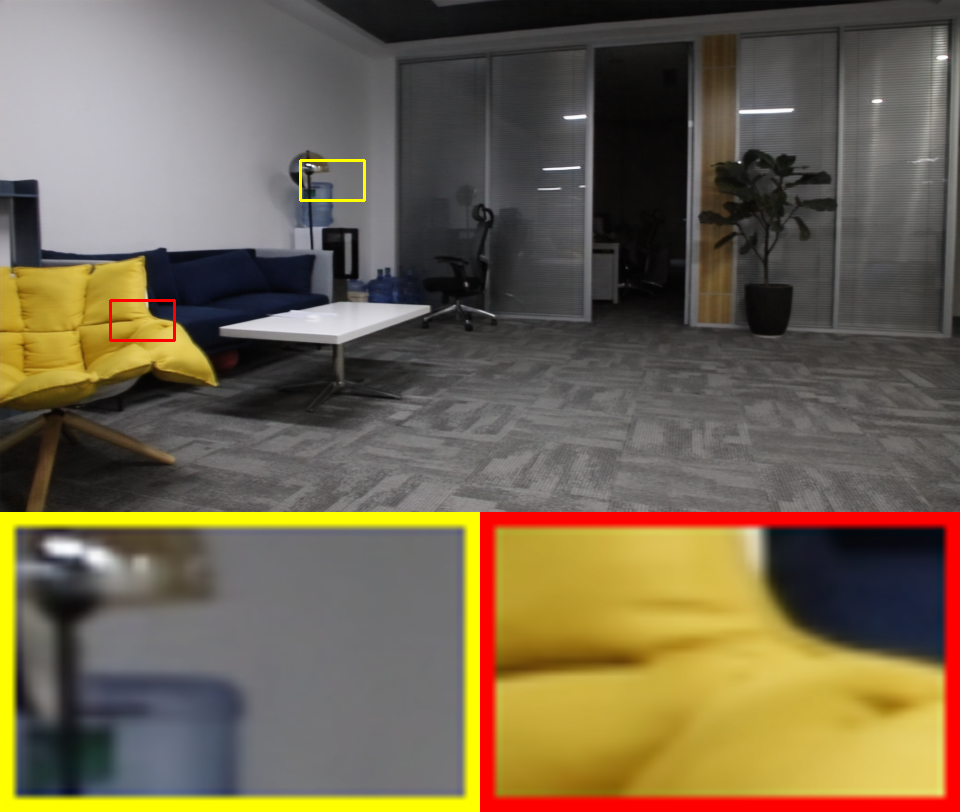}
		\caption*{VLLVE++}
	\end{subfigure} 
	\begin{subfigure}[c]{0.16\textwidth}
		\centering
		\includegraphics[width=1.15in]{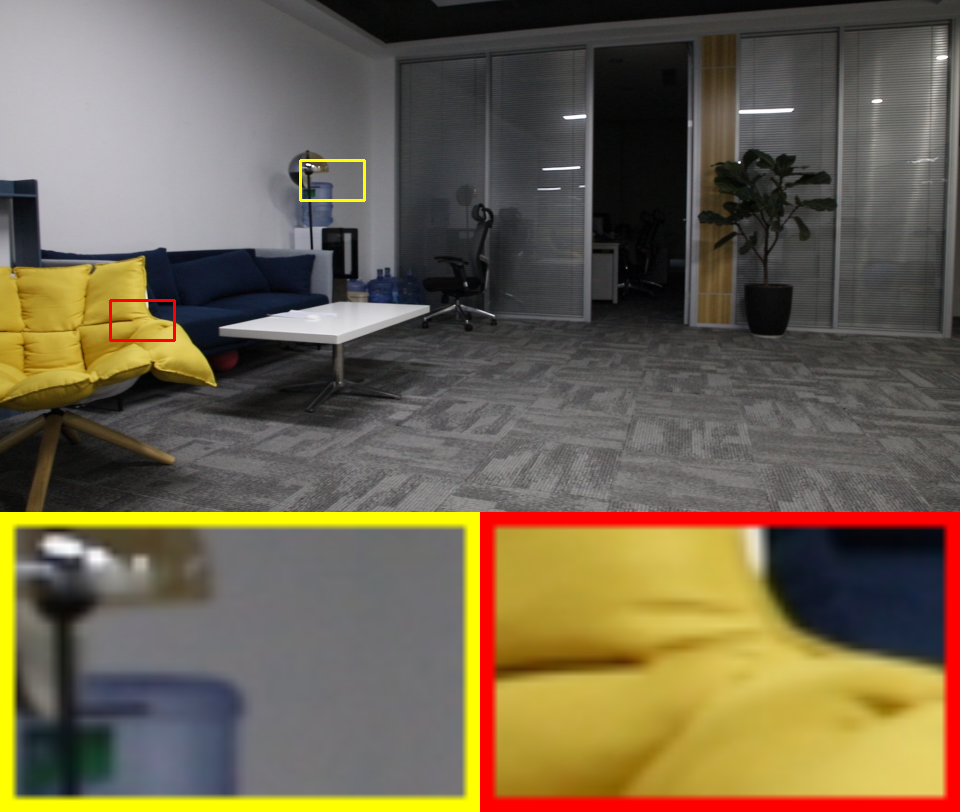}
		\caption*{GT}
	\end{subfigure} \vspace{0.4em}  \\

			\begin{subfigure}[c]{0.16\textwidth}
			\centering
			\includegraphics[width=1.15in]{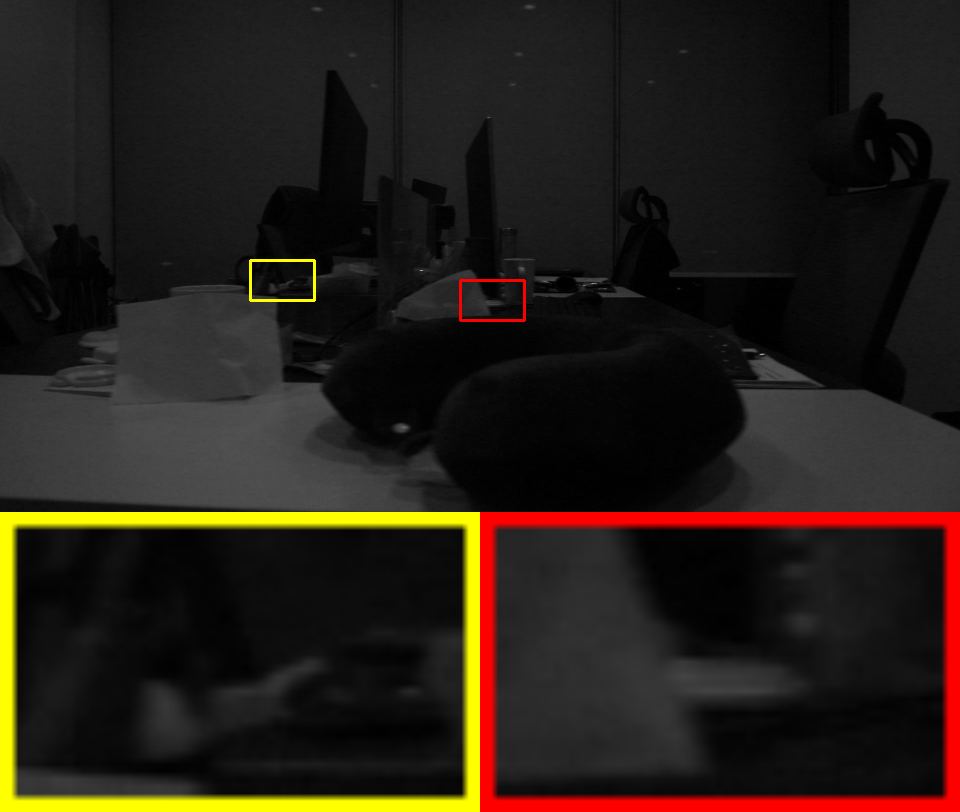}
			\caption*{Input}
		\end{subfigure}
		\begin{subfigure}[c]{0.16\textwidth}
			\centering
			\includegraphics[width=1.15in]{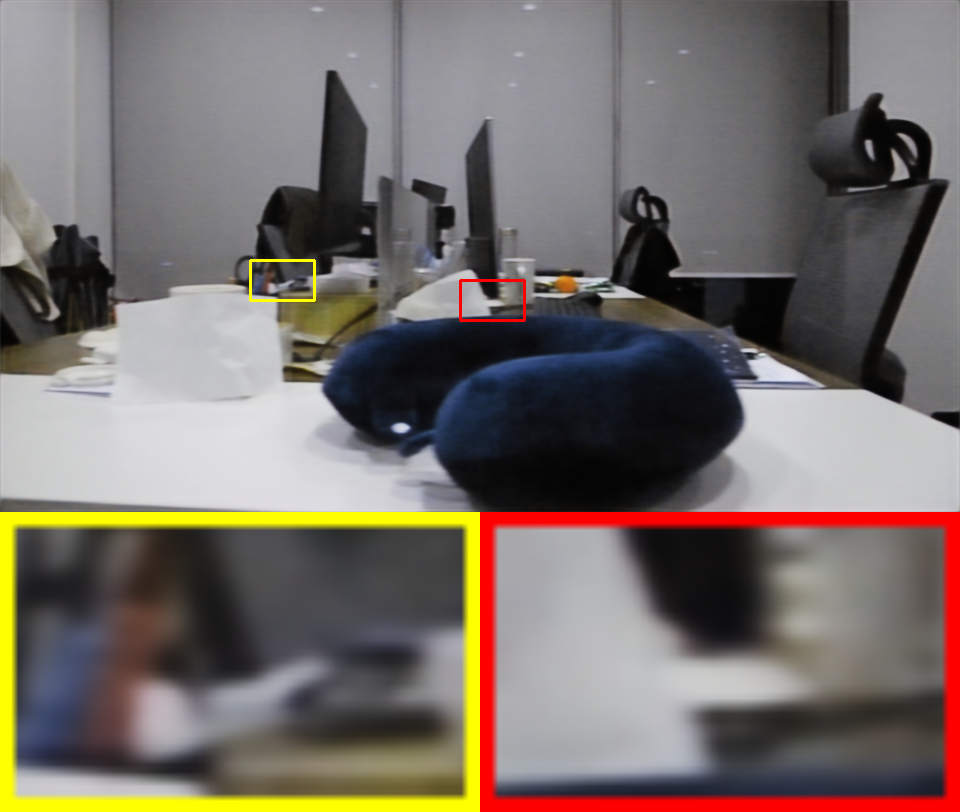}
			\caption*{RetinexFormer}
		\end{subfigure}
		\begin{subfigure}[c]{0.16\textwidth}
			\centering
			\includegraphics[width=1.15in]{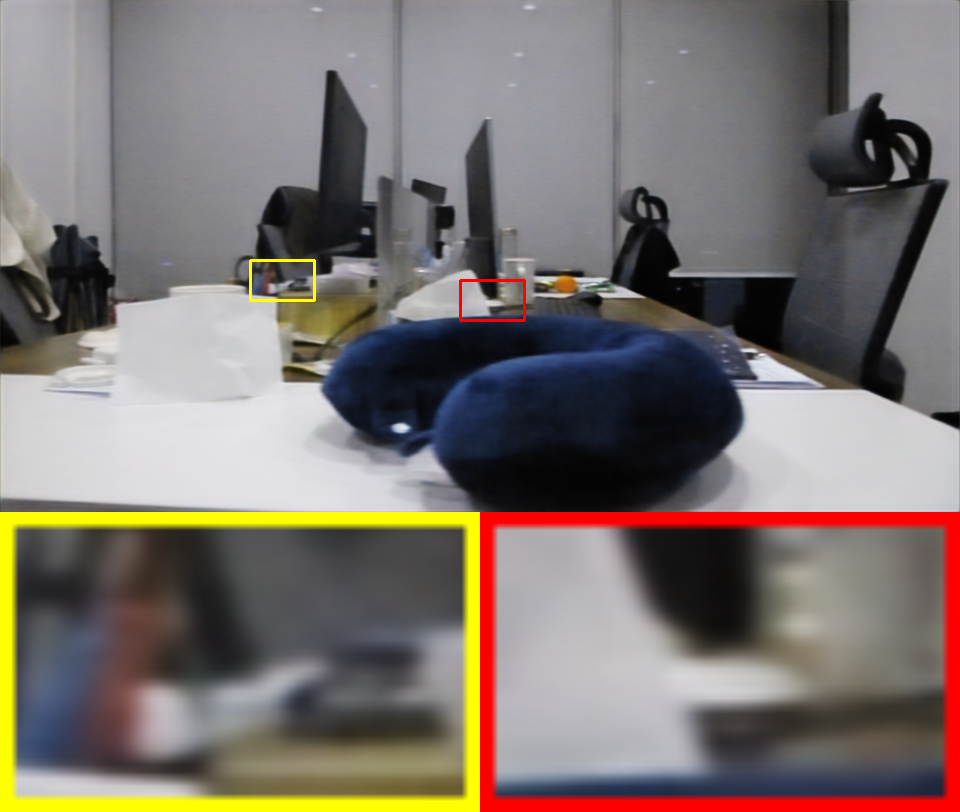}
			\caption*{DP3DF}
		\end{subfigure}
        		\begin{subfigure}[c]{0.16\textwidth}
			\centering
			\includegraphics[width=1.15in]{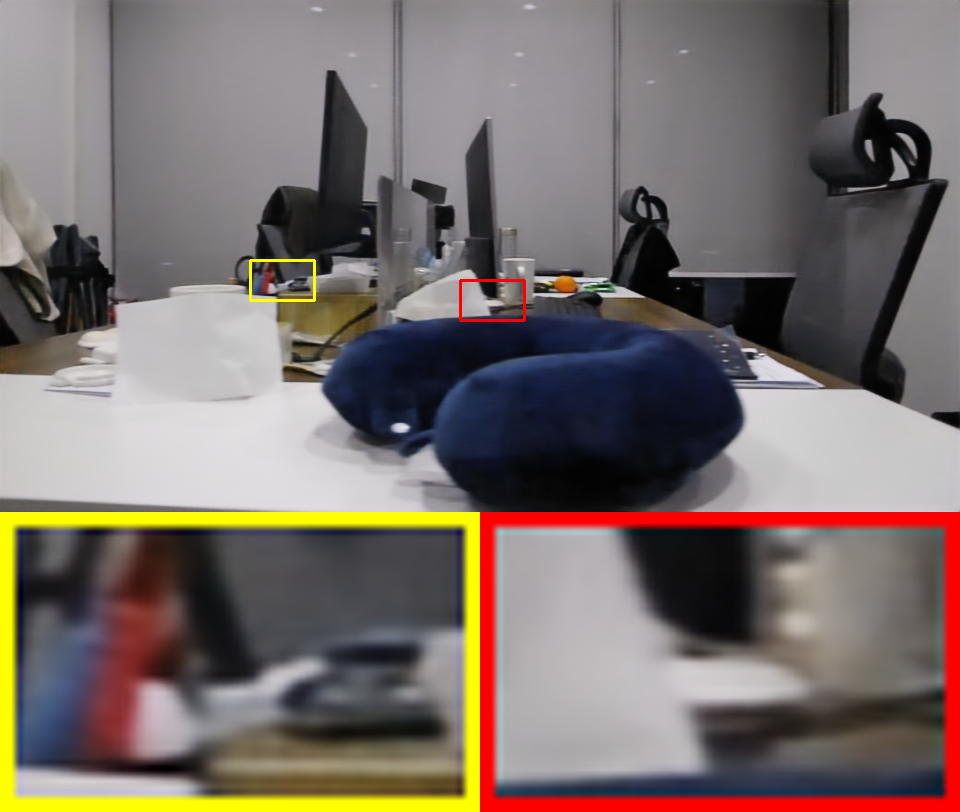}
			\caption*{VLLVE}
		\end{subfigure}
		\begin{subfigure}[c]{0.16\textwidth}
			\centering
			\includegraphics[width=1.15in]{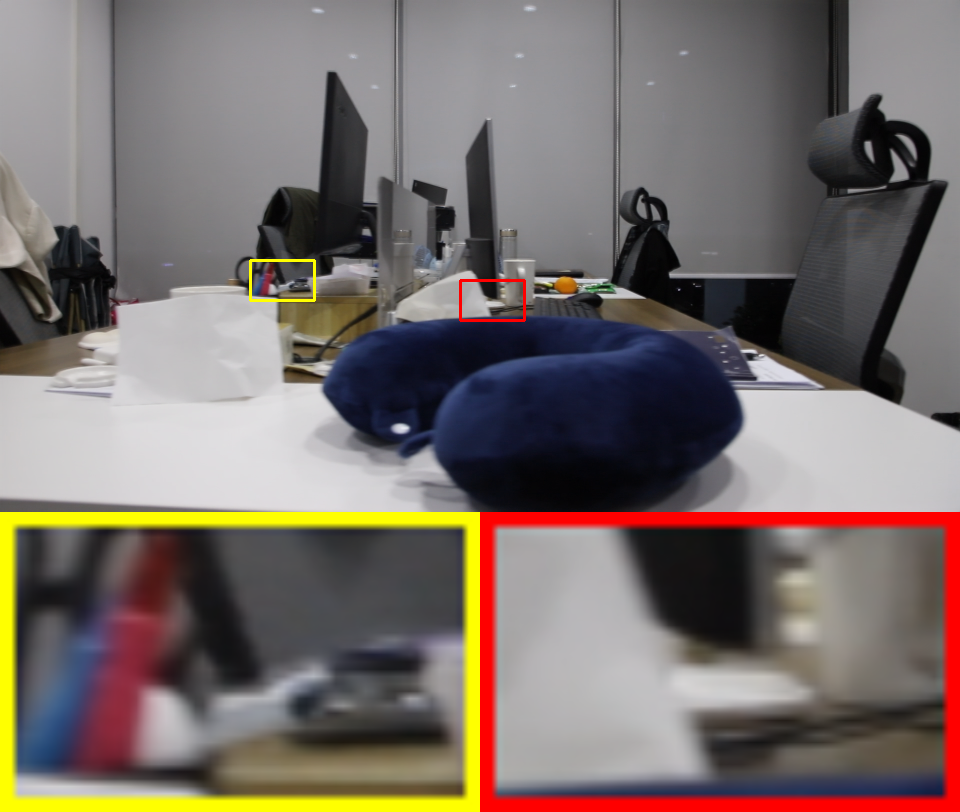}
			\caption*{VLLVE++}
		\end{subfigure} 
		\begin{subfigure}[c]{0.16\textwidth}
			\centering
			\includegraphics[width=1.15in]{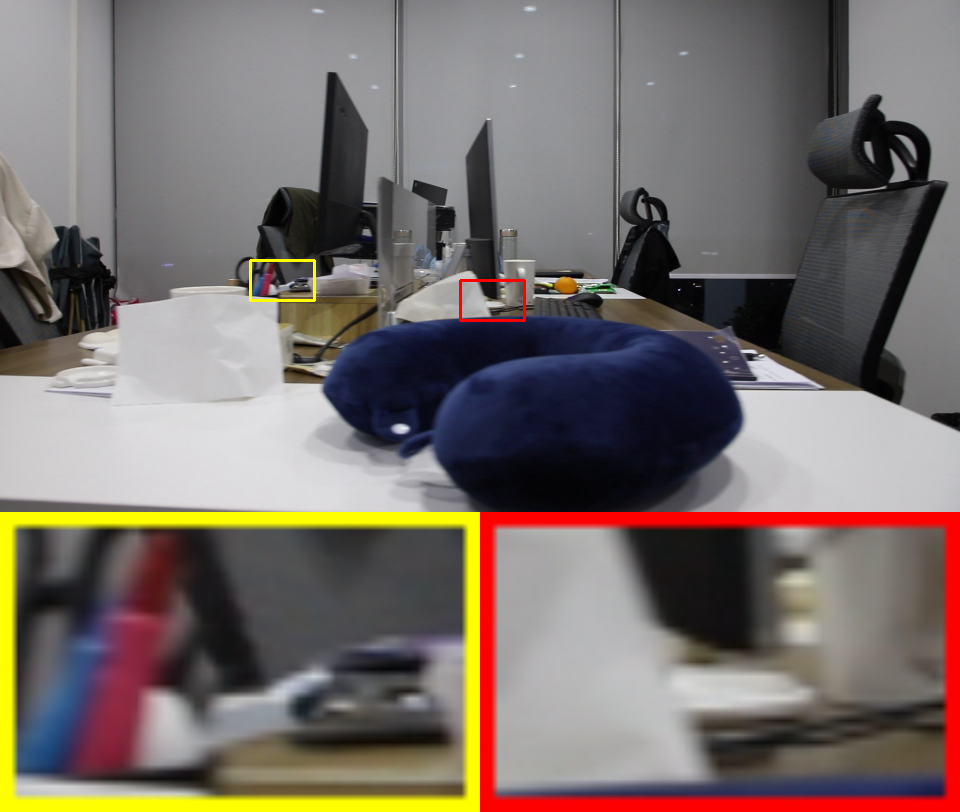}
			\caption*{GT}
		\end{subfigure} \vspace{0.4em} \\
	
	\begin{subfigure}[c]{0.16\textwidth}
		\centering
		\includegraphics[width=1.15in]{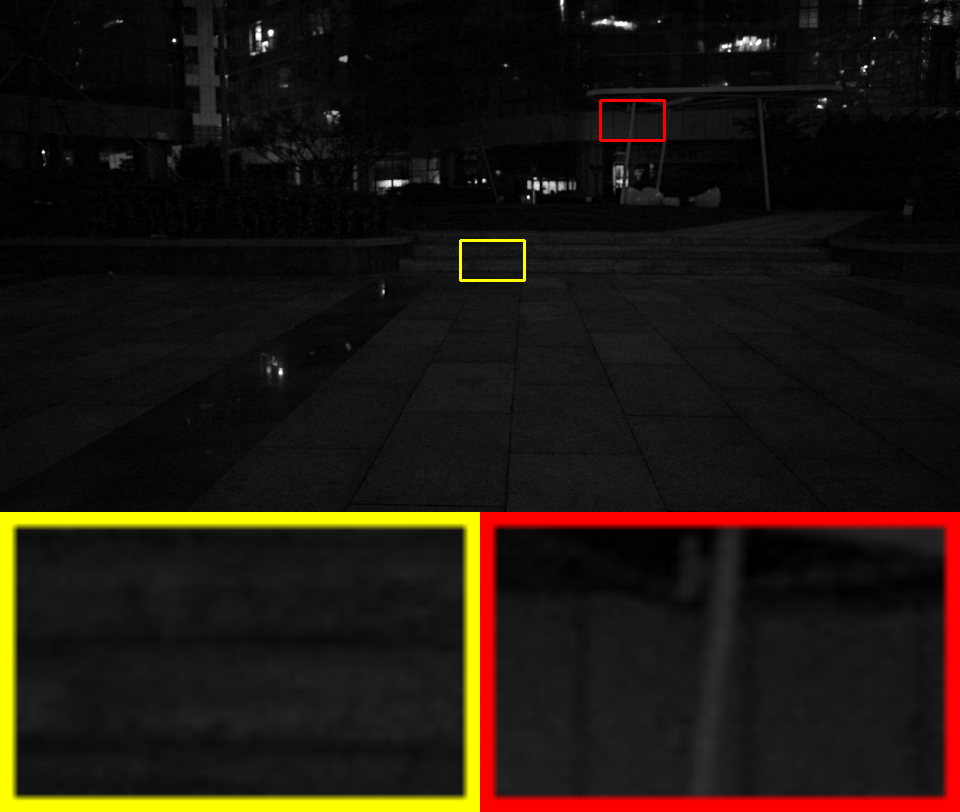}
		\caption*{Input}
	\end{subfigure}
	\begin{subfigure}[c]{0.16\textwidth}
		\centering
		\includegraphics[width=1.15in]{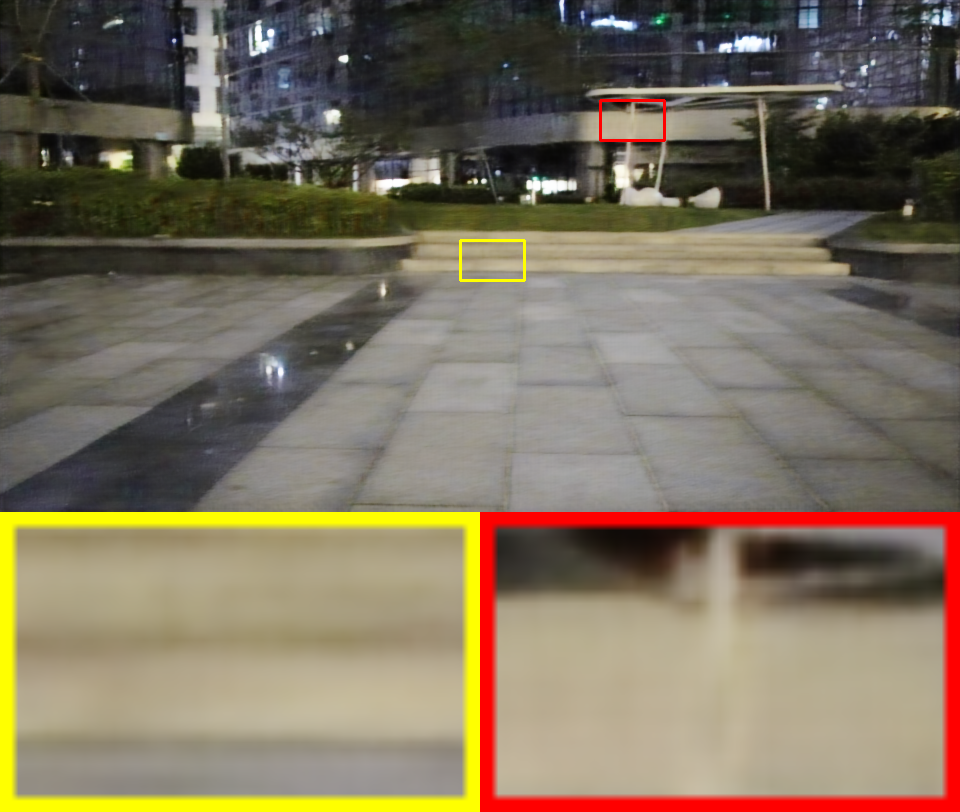}
		\caption*{RetinexFormer}
	\end{subfigure}
	\begin{subfigure}[c]{0.16\textwidth}
		\centering
		\includegraphics[width=1.15in]{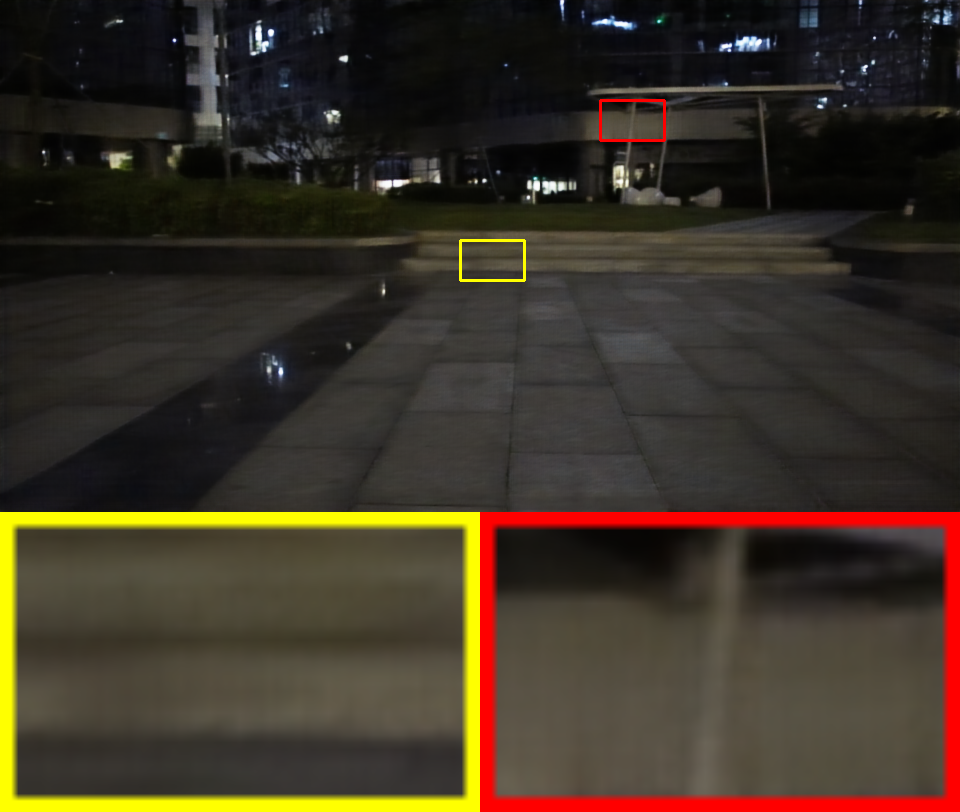}
		\caption*{DP3DF}
	\end{subfigure}
    	\begin{subfigure}[c]{0.16\textwidth}
		\centering
		\includegraphics[width=1.15in]{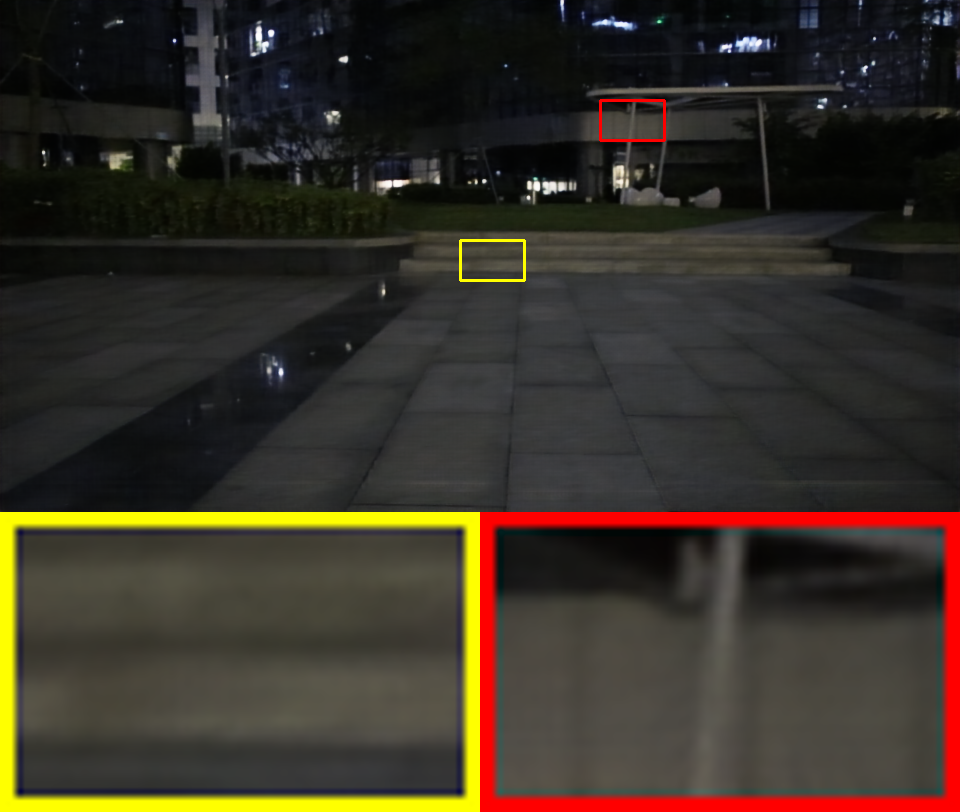}
		\caption*{VLLVE}
	\end{subfigure}
	\begin{subfigure}[c]{0.16\textwidth}
		\centering
		\includegraphics[width=1.15in]{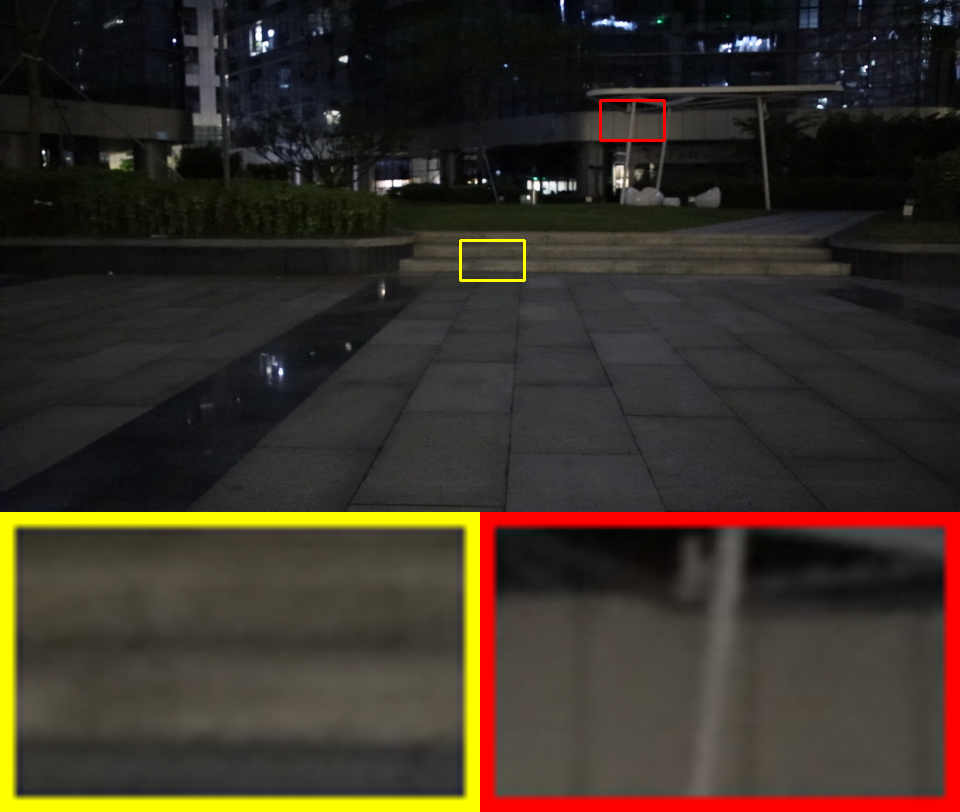}
		\caption*{VLLVE++}
	\end{subfigure} 
	\begin{subfigure}[c]{0.15\textwidth}
		\centering
		\includegraphics[width=1.15in]{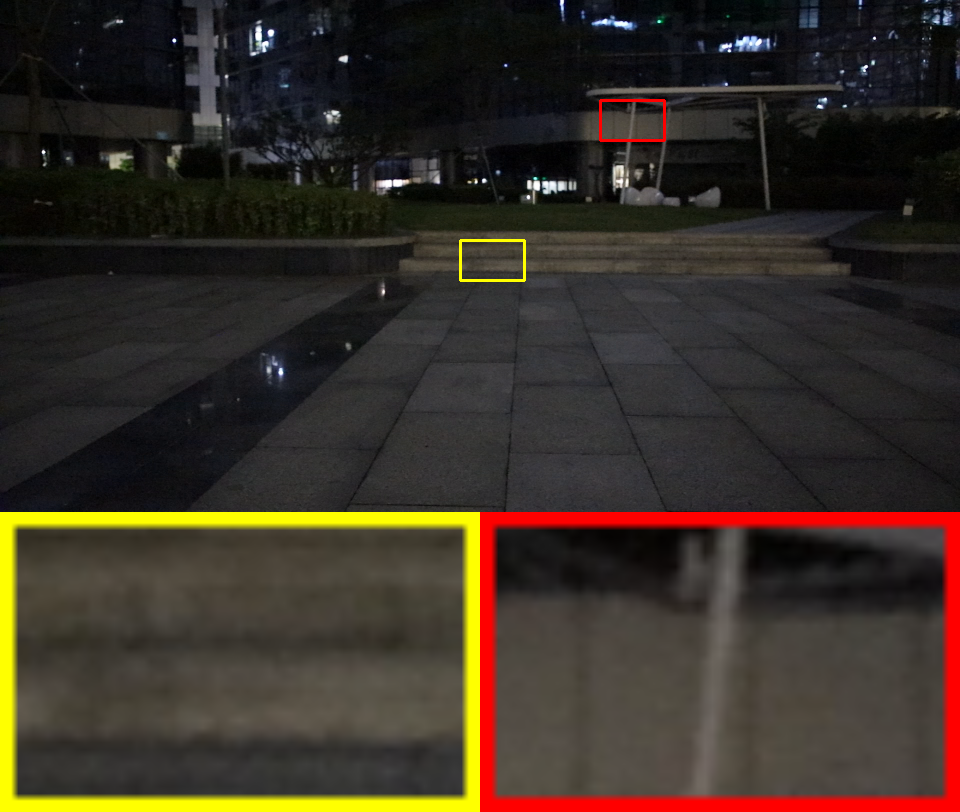}
		\caption*{GT}
	\end{subfigure} \vspace{0.4em} \\

\begin{subfigure}[c]{0.16\textwidth}
	\centering
	\includegraphics[width=1.15in]{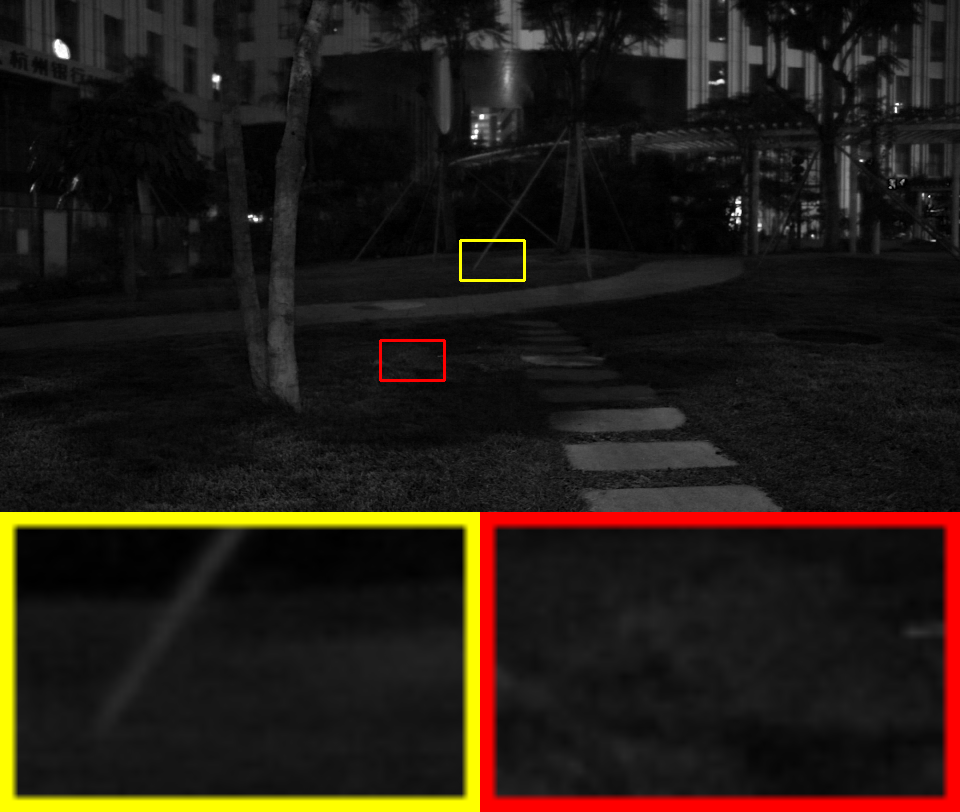}
	\caption*{Input}
\end{subfigure}
\begin{subfigure}[c]{0.16\textwidth}
	\centering
	\includegraphics[width=1.15in]{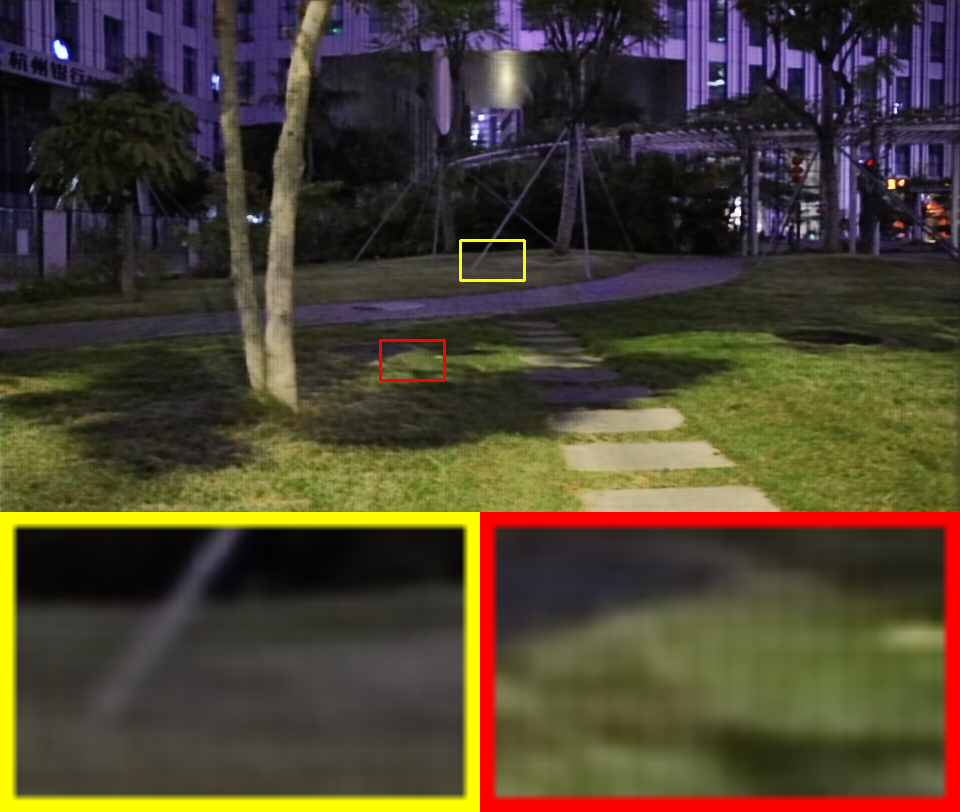}
	\caption*{RetinexFormer}
\end{subfigure}
\begin{subfigure}[c]{0.16\textwidth}
	\centering
	\includegraphics[width=1.15in]{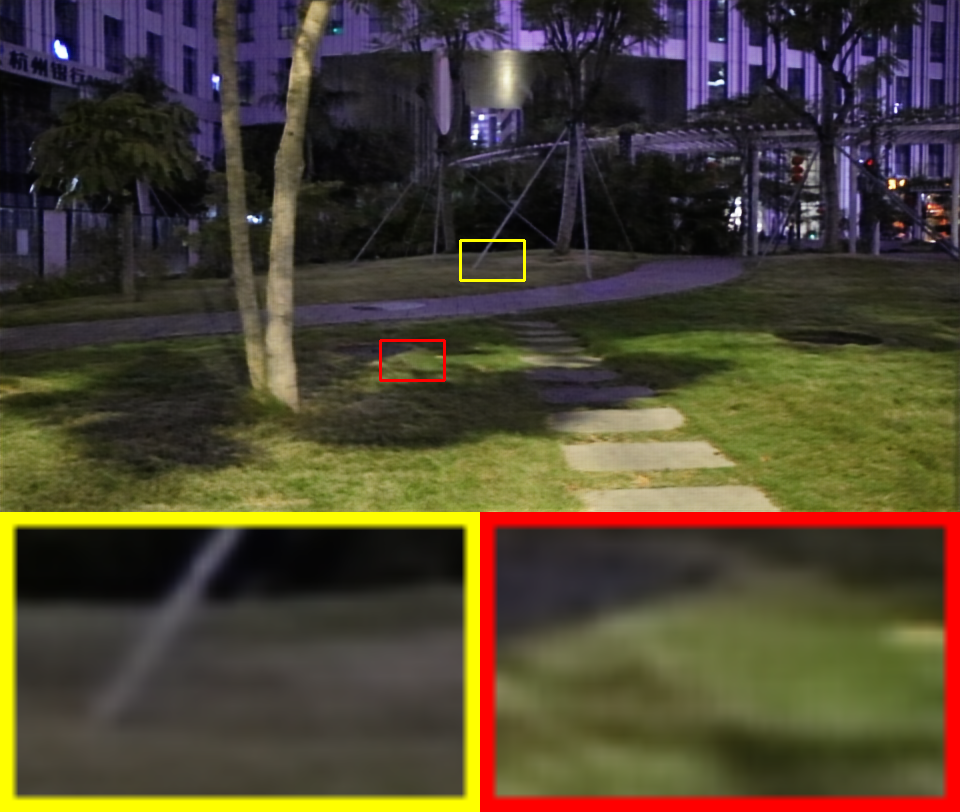}
	\caption*{DP3DF}
\end{subfigure}
\begin{subfigure}[c]{0.16\textwidth}
	\centering
	\includegraphics[width=1.15in]{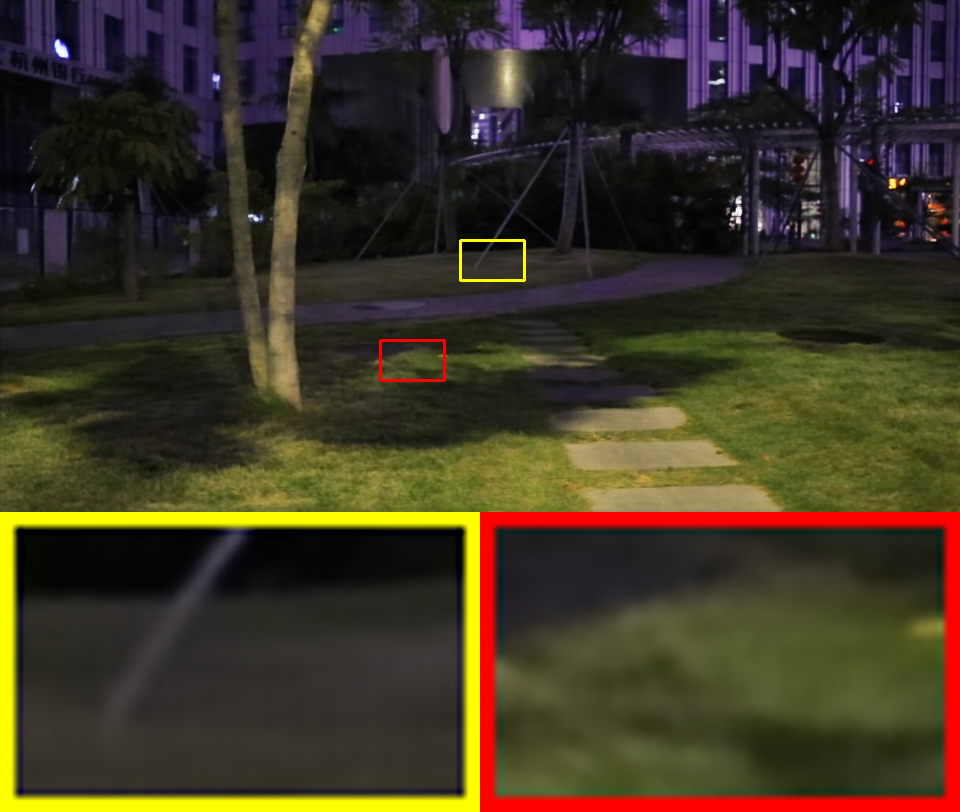}
	\caption*{VLLVE}
\end{subfigure}
\begin{subfigure}[c]{0.16\textwidth}
	\centering
	\includegraphics[width=1.15in]{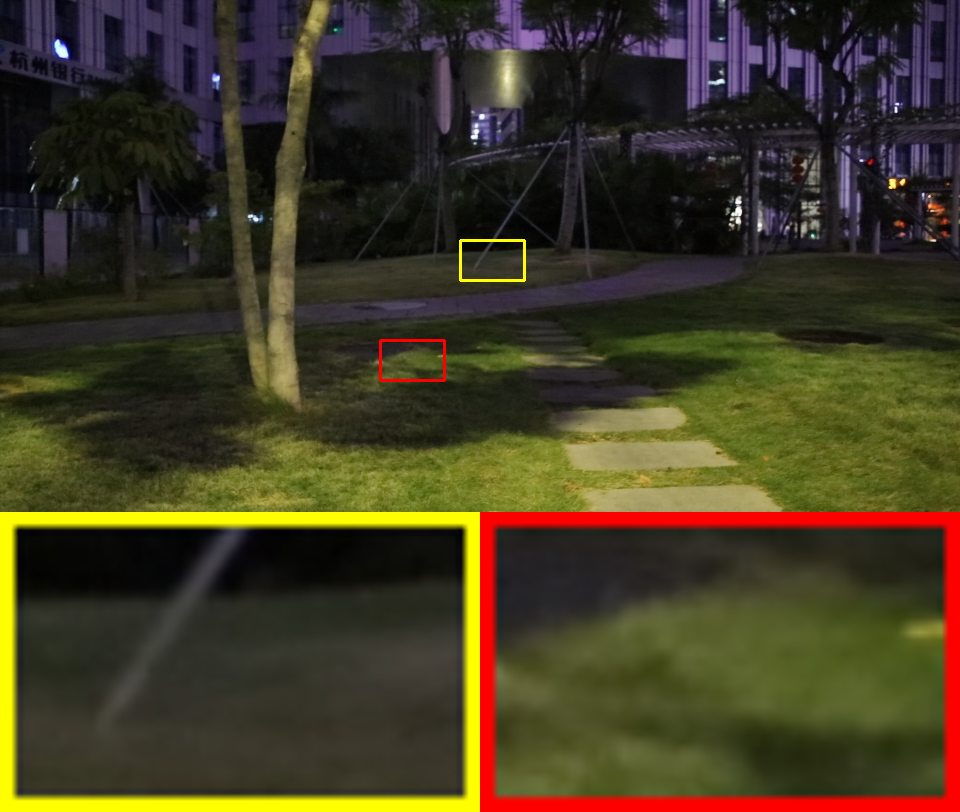}
	\caption*{VLLVE++}
\end{subfigure} 
\begin{subfigure}[c]{0.16\textwidth}
	\centering
	\includegraphics[width=1.15in]{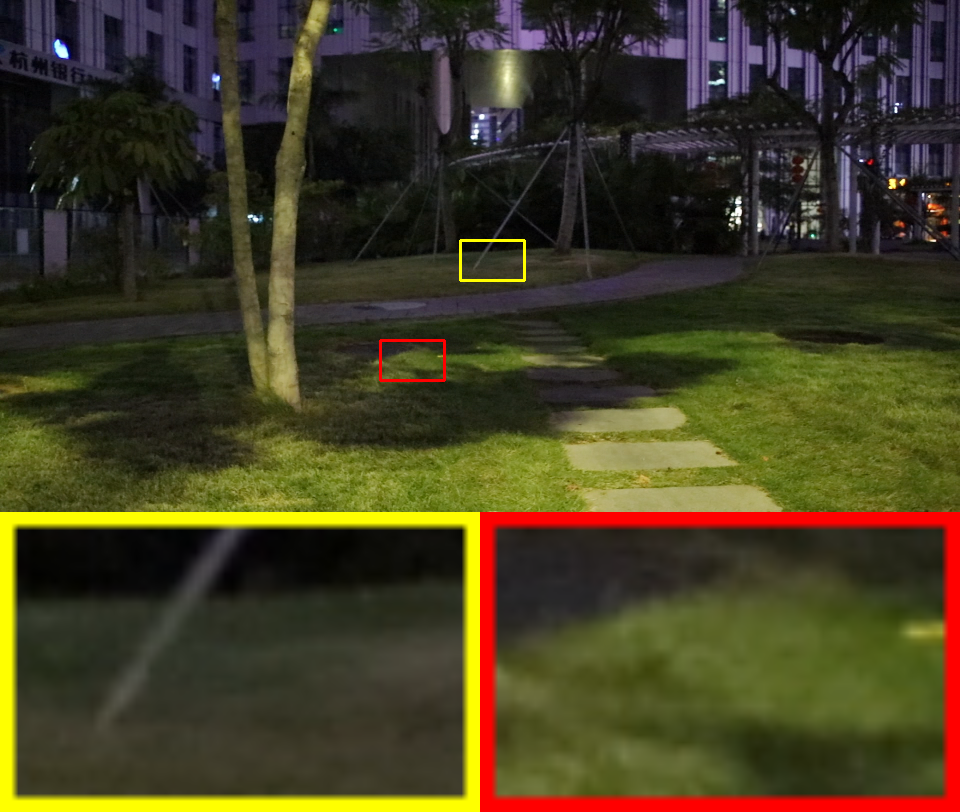}
	\caption*{GT}
\end{subfigure} \vspace{0.4em}  \\

	\begin{subfigure}[c]{0.16\textwidth}
		\centering
		\includegraphics[width=1.15in]{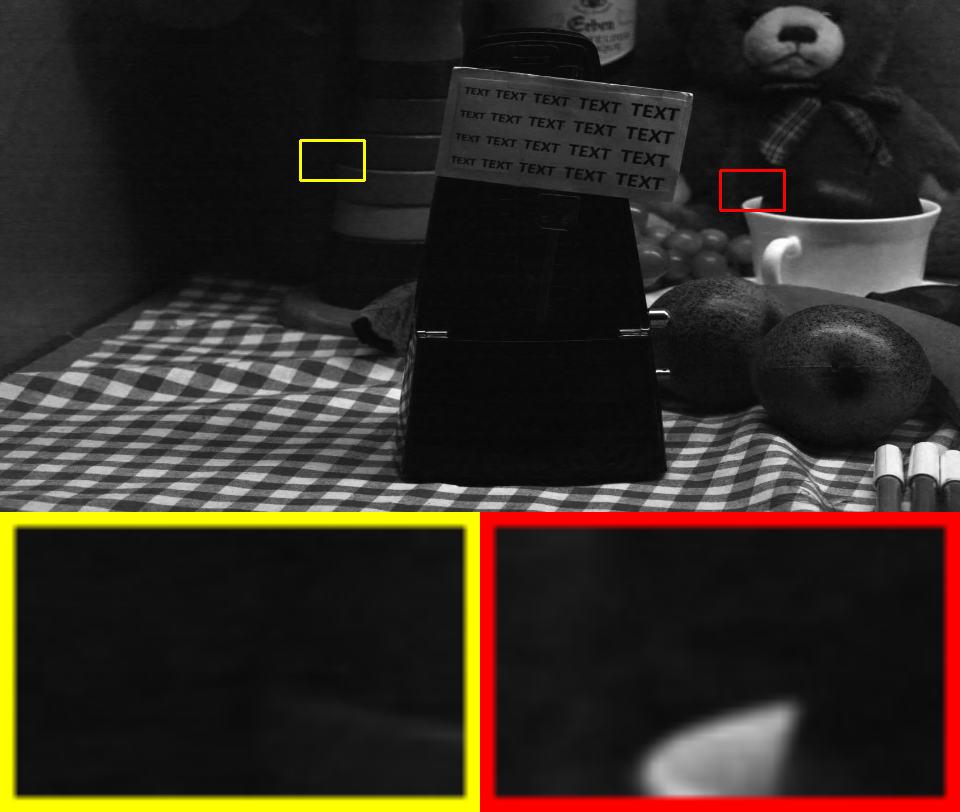}
		\caption*{Input}
	\end{subfigure}
	\begin{subfigure}[c]{0.16\textwidth}
		\centering
		\includegraphics[width=1.15in]{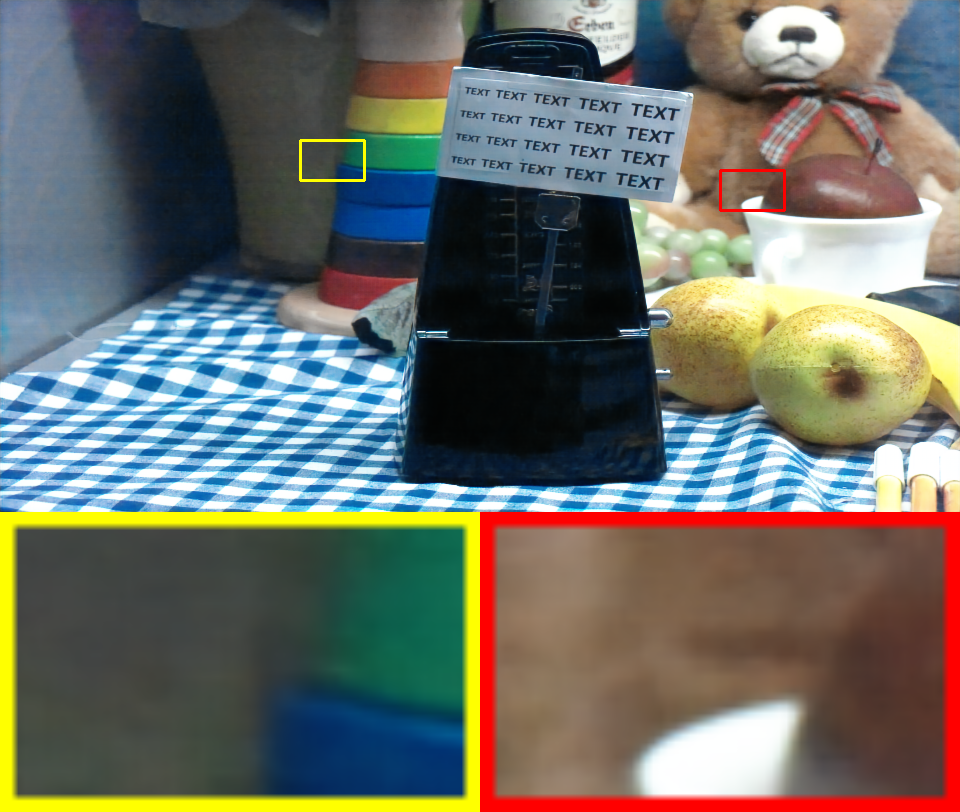}
		\caption*{RetinexFormer}
	\end{subfigure}
	\begin{subfigure}[c]{0.16\textwidth}
		\centering
		\includegraphics[width=1.15in]{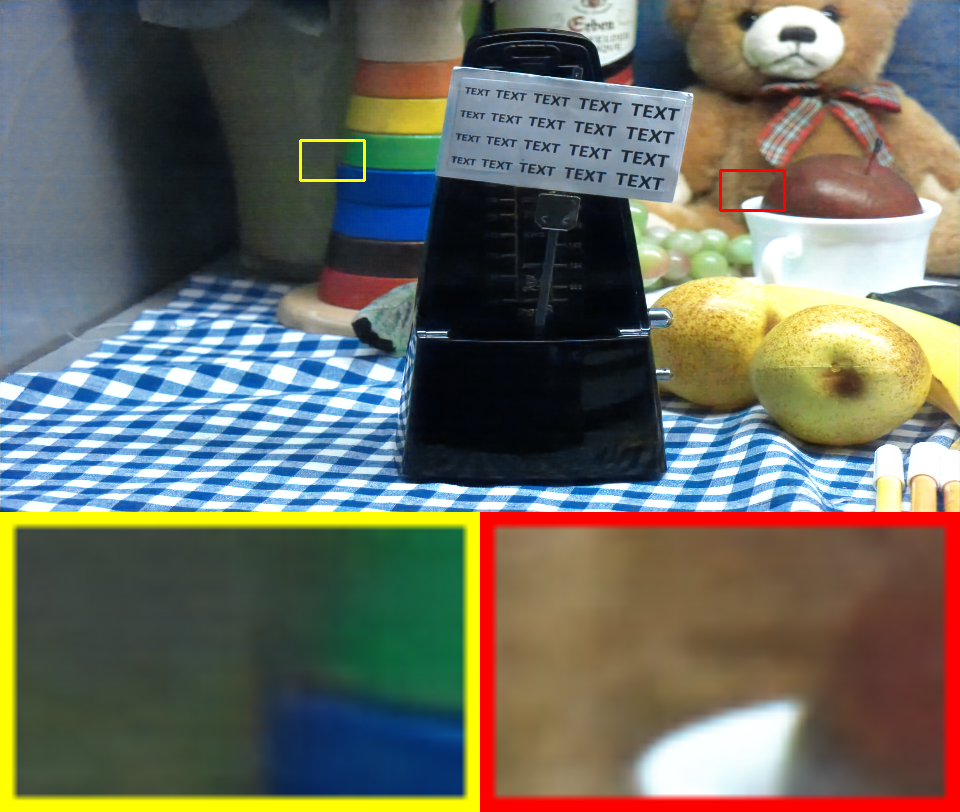}
		\caption*{DP3DF}
	\end{subfigure}
    	\begin{subfigure}[c]{0.16\textwidth}
		\centering
		\includegraphics[width=1.15in]{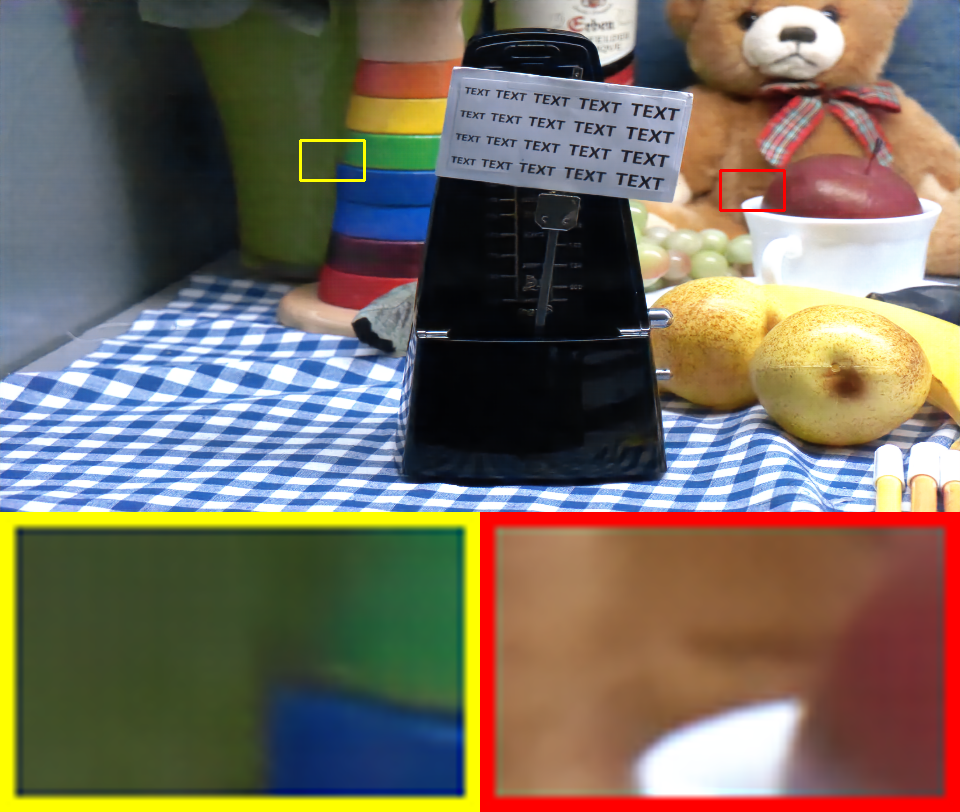}
		\caption*{VLLVE}
	\end{subfigure}
	\begin{subfigure}[c]{0.16\textwidth}
		\centering
		\includegraphics[width=1.15in]{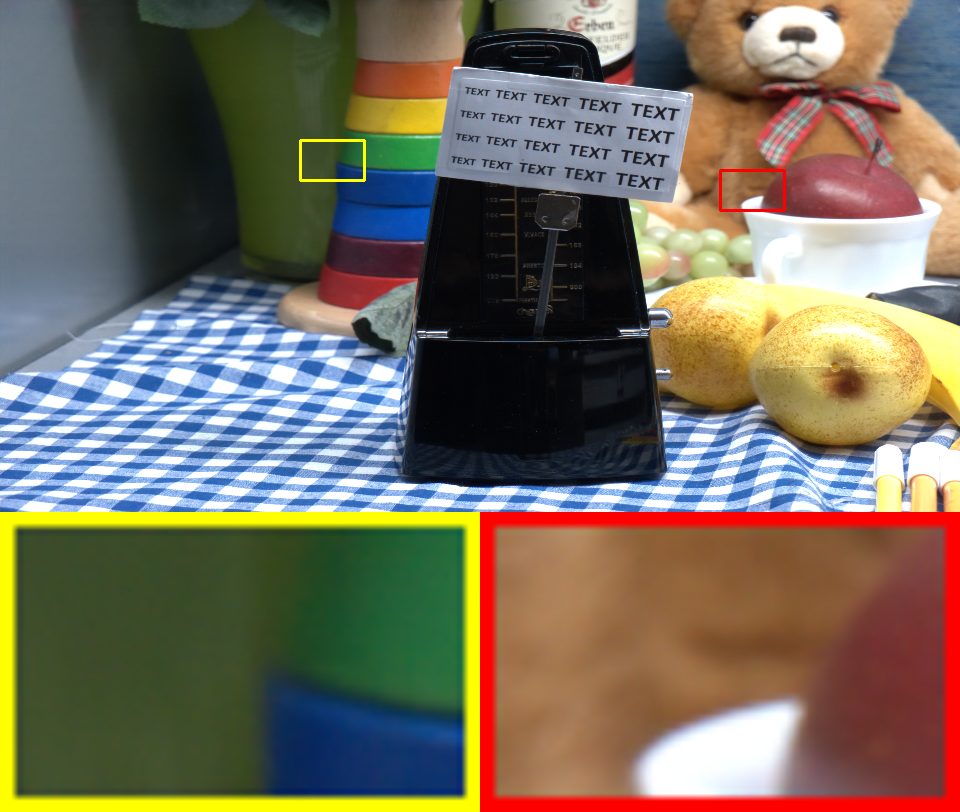}
		\caption*{VLLVE++}
	\end{subfigure} 
	\begin{subfigure}[c]{0.16\textwidth}
		\centering
		\includegraphics[width=1.15in]{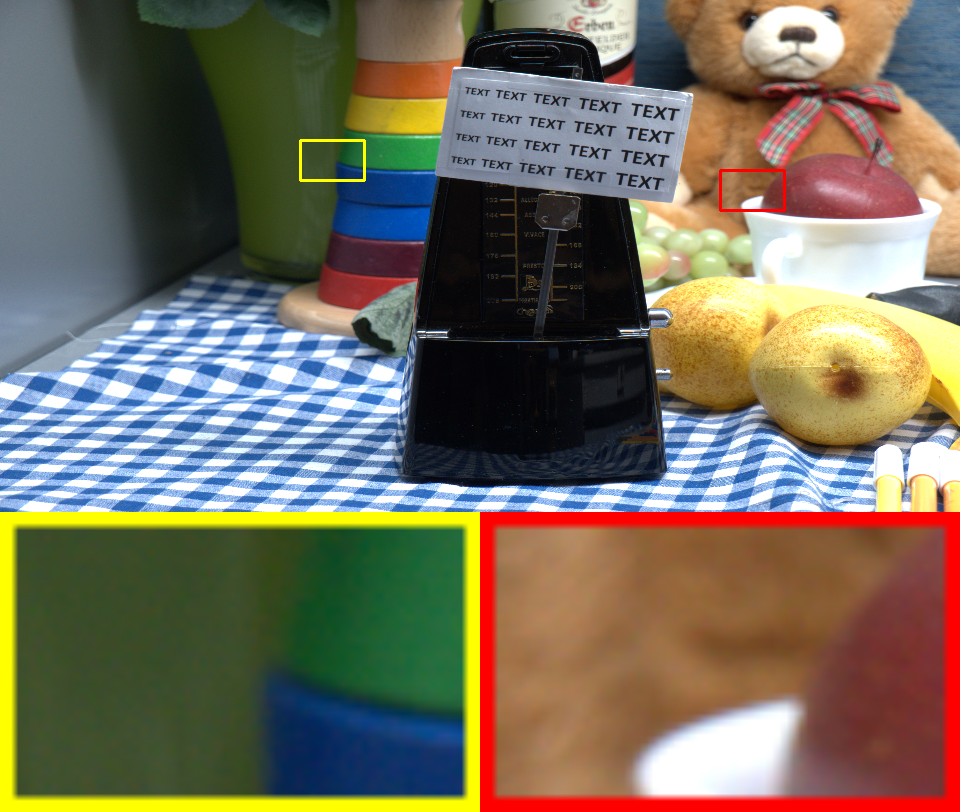}
		\caption*{GT}
	\end{subfigure} \vspace{0.4em} \\

\begin{subfigure}[c]{0.16\textwidth}
	\centering
	\includegraphics[width=1.15in]{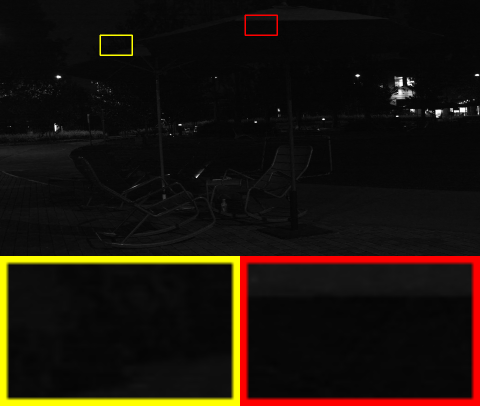}
	\caption*{Input}
\end{subfigure}
\begin{subfigure}[c]{0.16\textwidth}
	\centering
	\includegraphics[width=1.15in]{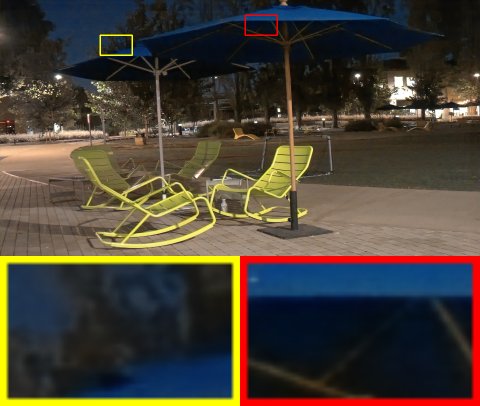}
	\caption*{RetinexFormer}
\end{subfigure}
\begin{subfigure}[c]{0.16\textwidth}
	\centering
	\includegraphics[width=1.15in]{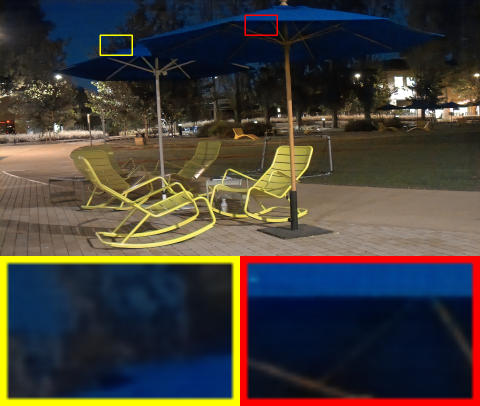}
	\caption*{DP3DF}
\end{subfigure}
\begin{subfigure}[c]{0.16\textwidth}
	\centering
	\includegraphics[width=1.15in]{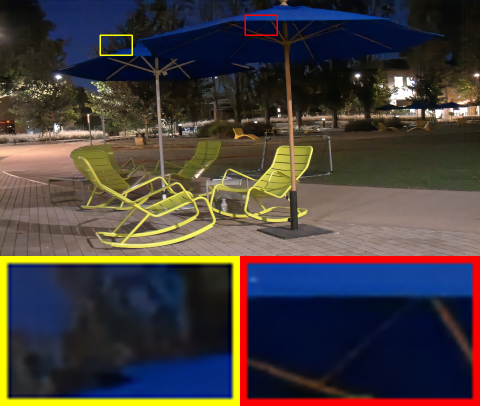}
	\caption*{VLLVE}
\end{subfigure}
\begin{subfigure}[c]{0.16\textwidth}
	\centering
	\includegraphics[width=1.15in]{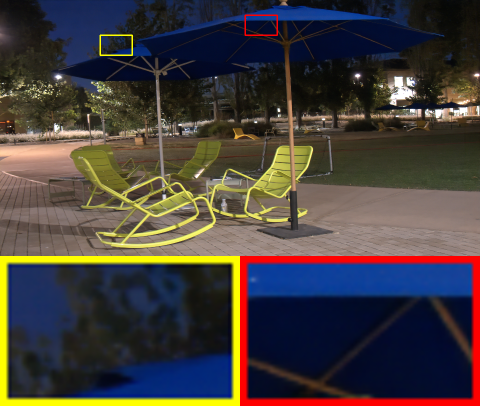}
	\caption*{VLLVE++}
\end{subfigure} 
\begin{subfigure}[c]{0.16\textwidth}
	\centering
	\includegraphics[width=1.15in]{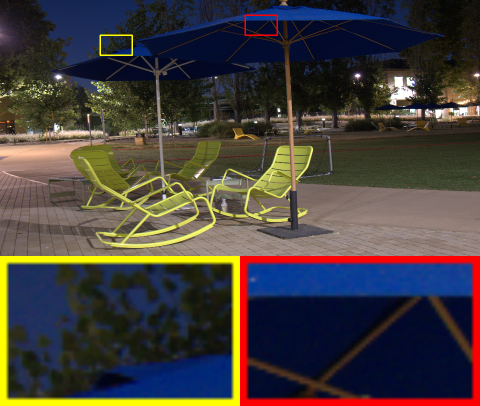}
	\caption*{GT}
\end{subfigure} \\

    \vspace{-0.1in}
	\caption{Visual comparisons on SDSD-indoor (top two rows), SDSD-outdoor (middle two rows), and SMID (bottom two rows). The results of our proposed frameworks demonstrate better visual perception with less noise, clearer visibility, and more enhanced details.}
	
	\label{fig:cmp1}
\end{figure*}

\vspace{-0.1in}
\minisection{Evaluation for Temporal Consistency}
Temporal consistency and stability should also be evaluated for LLVE.
Thus, we employ the short-term and long-term temporal loss proposed in \cite{lai2018learning} for such temporal evaluation on different datasets.
The wrapping operations among frames are computed on normal-light frames.
Moreover, the long-term loss is computed every 10 frames.
The results are shown in \Cref{comparison5-temporal} (we normalize the frame values into [0, 1]). 

VLLVE has lower temporal loss than baselines, demonstrating that videos produced from our approach are temporally stable.
Furthermore, VLLVE++ achieves superior temporal consistency, with reduced flickering artifacts compared to VLLVE. This improvement stems from our enhanced decomposition and refinement strategy, which more effectively models degradations and reconstructs individual frames.

\vspace{-0.1in}
\minisection{Qualitative Result}
Besides the quantitative comparisons, we provide visual comparisons.
\Cref{fig:cmp1} showcases the visual comparisons of SDSD and SMID, while \Cref{fig:cmp2} displays visual cases from DID and DAVIS. In general, the results enhanced by our approach exhibit a more natural appearance, including accurate color and brightness, enhanced contrast, and precise details.
Furthermore, our results show fewer artifacts in regions with complex textures, and they are closer to the ground truth than the results produced by other methods.
Meanwhile, VLLVE++ is better than VLLVE.

\begin{table}[tb!]
	\centering
        \caption{The comparisons with current representative efficient low-light enhancement methods in terms of efficiency. We present comprehensive details regarding model complexities, including parameters, GFLOPS, and inference time. We assess GFLOPS and inference time using an input size of 256 $\times$ 256. We use each model to process a video with 100 frames and compute the per-frame cost.} 
    \label{comparison-computation}
    \vspace{-0.1in}
    \renewcommand{\arraystretch}{1.06}
	\resizebox{1.0\linewidth}{!}
    {
		\begin{tabular}{|l|cccc|}
            \hline
            & Restormer~\cite{zamir2022restormer} & LLFlow~\cite{wang2022low} & SNR~\cite{xu2022snr} &Retinexformer~\cite{cai2023retinexformer} \\ \hline
			Param.(M) & 26.13 &37.68 &39.12& 1.61  \\ \hline
            GFLOPS  &144.25& 287& 26.35 & 15.57\\  \hline
            & Diff-L~\cite{jiang2023low} &SMOID~\cite{jiang2019learning}&VLLVE~\cite{xu2025low} & VLLVE++ \\ \hline
            Param.(M) &22.08&12.21&3.16 & 3.87 \\ \hline
            GFLOPS &88.92&15.11&17.34 & 20.15\\ 
            \hline
	\end{tabular}}
\end{table}

\minisection{Computation Cost Analysis}
Here, we evaluate the computational cost. As shown in Table~\ref{comparison-computation}, the efficiency of our method is comparable to the most efficient methods, such as Retinexformer, while our performance is clearly superior. This is because our method is built upon a simple U-Net architecture, and the additional cost from the CFIM is not large. These experiments demonstrate that the performance of our method is achieved without sacrificing model efficiency.

\begin{table}[tb!]
	\centering
	\large
	\caption{Ablation studies for VLLVE on SDSD, SMID, and DID.} 
    \label{comparison-abla-supp}
    \vspace{-0.1in}
    \renewcommand{\arraystretch}{1.06}
    \resizebox{1.0\linewidth}{!}
    {
        \begin{tabular}{|l|cc|cc|cc|cc|}
            \hline
            & \multicolumn{2}{c|}{SDSD-indoor} &\multicolumn{2}{c|}{SDSD-outdoor}& \multicolumn{2}{c|}{SMID} &\multicolumn{2}{c|}{DID}  \\
            \hline
            Methods & PSNR & SSIM& PSNR & SSIM & PSNR & SSIM& PSNR & SSIM\\
            \hline \hline

            w/o LR & 25.78 &0.77  &23.49  & 0.75 &26.37 &0.76  & 27.54 &0.86 \\
			w/o LL & 26.89 & 0.80 & 24.19 &0.78  &27.74 & 0.79 & 27.72 & 0.89\\
			w/o C.F. & 25.18 & 0.82 & 24.22 & 0.77 &26.75 &0.77  &25.60  &0.85 \\
			with M.I. & 26.21 & 0.84 & 25.36 & 0.80 &28.42 &0.79  &28.33  &0.88 \\
			w/o Dual & 27.53 & 0.84 &24.83  &0.79  &27.43 & 0.78 & 26.91 &0.86 \\ \hline
            
            with 2 down & 28.07 &0.86  &25.89  &0.81  & 29.14&0.80  & 29.48&0.92 \\ 
            with 4 down & 27.81 & 0.85 &25.61  &0.78  &28.76 &0.78  &29.13 &0.90 \\ 
            \hline
            with LoFTR & 28.21 &0.87  & 25.86 & 0.80 &29.29 & 0.80 &29.35 &0.91 \\ 
            with PDC-Net+ &29.03  & 0.88 & 26.22 & 0.81 &29.85 &0.82  &29.57 &0.91 \\ 
            \hline
            with F.L. & 26.55 &0.82  & 23.80 & 0.74 &27.67 & 0.78 &28.09 &0.88 \\ 
            \hline
            with N3 & 29.12 &0.90  & 27.67 &0.83  &30.18 &0.84  &30.51 &0.94 \\ 
            with 2T & 28.52 & 0.86 &26.51  & 0.82 &29.95 &0.83  &30.54 &0.93 \\
            with T/2 &28.84  & 0.87 &26.95  &0.84  &29.10 &0.81  &29.86 &0.90 \\
            with same & 28.87 &0.88  &26.39  &0.82  &29.50 &0.82  & 30.08&0.93 \\
            \hline
            with Restormer & 29.44 &0.90  &27.16  &0.85  &30.02 &0.83  &30.15 & 0.92 \\
            with MIRNet & 29.16 &0.89  & 26.85 &0.83  &29.86 &0.81 & 29.80 &0.89\\ 
            with SNR & 30.09 &0.92  & 27.78 &0.85  &30.11 &0.84  &30.87 & 0.94\\ 
            \hline 
            VLLVE& 28.93&0.88 &26.32 & 0.82& 29.60 &0.82 & 30.10&0.93 \\
            \hline
        \end{tabular}
        }
\end{table}

\subsection{Ablation Study}
\label{sec:ablation-study}

We first analyze the various components of the base VLLVE framework, including \textbf{investigations not contained in our original conference version of IJCAI2025}. Furthermore, we conduct additional ablation experiments to evaluate the new components introduced in VLLVE++, specifically the enhanced decomposition strategy and bidirectional correspondence refinement module.

\subsubsection{Extensive Ablation Studies for Components in VLLVE}

\vspace{-0.1in}
\minisection{Ablation Study for VLLVE's Components}
(1) ``w/o LR'': remove the constraint on learning the view-independent part. (2) ``w/o LL'': the constraint on learning the view-dependent part is deleted. (3) ``w/o C.F.'': we eliminate the cross-frame attention and fusion in the network, i.e., each frame being enhanced individually. (4) ``with M.I.'': use multiple neighboring frames as input and employ the temporal alignment via deformable convolution. (5) ``w/o Dual'': the loss is applied to only one frame in each iteration.
The results are summarized in \Cref{comparison-abla-supp}. 

By comparing ``w/o LR'' and ``w/o LL'' with the full setting (``VLLVE''), we highlight the significance of the proposed constraints.
Furthermore, we validate the importance of mutually propagating features among different frames
by comparing ``w/o C.F.'' with ``VLLVE''. 
Our propagation strategy via the dual network proves to be more effective than the common strategy of using multi-frame inputs for propagation (``w/o M.I.'' v.s. ``VLLVE'').
The dual learning (simultaneously supervise the two outputs of the dual network) also proves to be effective (``w/o Dual'' v.s. ``VLLVE'').

\vspace{-0.1in}
\minisection{Compute Correspondences with Varying Image Resolution}
We can downscale $I_{n,t_1}$ and $I_{n,t_2}$ and compute correspondences at a faster speed. These ablation settings employ downsampling factors of 2 and 4 and are referred to as ``with 2 down'' and ``with 4 down'', respectively. The results are presented in Table~\ref{comparison-abla-supp}. While their performance is lower than when using the full resolution, their results still outperform baselines in Table~\ref{comparison5}.
This is because downsampling does not compromise the ability to capture correspondences among salient objects.

\vspace{-0.1in}
\minisection{Compute Correspondences with Other Models Besides DKM}
In the original implementation, we utilized the DKM model~\cite{edstedt2023dkm} to compute correspondences.
To assess our performance using alternative correspondence computation models, we conduct experiments using LoFTR~\cite{sun2021loftr} and PDC-Net+~\cite{truong2023pdc}. These settings are referred to as ``with LoFTR'' and ``with PDC-Net+'', respectively. The results are presented in Table~\ref{comparison-abla-supp}, and it is evident that our framework's performance with LoFTR and PDC-Net+ is comparable to the original setting, with some cases even yielding higher scores. 
Thus, ours performance is not significantly influenced by the choice of correspondence computation models.

\vspace{-0.1in}
\minisection{Apply Correspondence Constraints at Feature Level}
In original VLLVE, the constraint for the view-independent term is computed at the pixel level, using RGB values. 
As an alternative approach, we explored conducting the cross-frame correspondence constraint at the feature level within the corresponding decoder branch in our framework. This ablation setting is denoted as ``with F.L.''.
The results are presented in \Cref{comparison-abla-supp}, and it's apparent that this approach's performance is inferior to the original pixel-level setting.

\vspace{-0.1in}
\minisection{Varying Frames for Propagation}
Our VLLVE framework offers the flexibility to extend feature propagation among a variable number of frames.
To investigate this extension, we introduced an ablation setting where we load a different number of frames, denoted as $N$ ($N=2$ in the original implementation), for propagation and simultaneous supervision during training. 
During training, $N$ frames are randomly chosen from the neighboring frames and designated as the closest frames during inference. 
We conducted experiments with $N$ set to 3, referred to as ``with N3''. 
The results in \Cref{comparison-abla-supp} indicate that their performance surpasses that of the original setting with $N=2$. This improvement can be attributed to the fact that more extensive propagation leads to a more comprehensive understanding of the target scene.

\vspace{-0.1in}
\minisection{Selection of Reference Frames}
In the original setting, we select the reference frame from among the $T$ neighboring frames for the target frame. To further analyze this selection process, we conducted ablation experiments using both $2T$ and $\frac{T}{2}$ neighboring frames for reference frame selection. These results are labeled ``with 2T'' and ``with T/2'' in \Cref{comparison-abla-supp}.
Their performance closely matches that of the original setting, underscoring the robustness of our VLLVE to the choice of reference frame.

Furthermore, we explored an additional ablation setting where the target frame and the reference frame are chosen to be the same (feature propagation occurs during training).
This result is labeled ``with same'' in \Cref{comparison-abla-supp}, and it's worth noting that its performance closely resembles that of the original setting where the closest frame is used as the reference frame.

Note that the performance across ``with 2T'', ``with T/2'', ``with Same'', and the original setting is similar, as their differences arise from the choice of reference frame only. It is important to note that ``with 2T'', ``with T/2'', and ``with Same'' all use the same training model but vary the reference frame during inference. Additionally, since the temporal size T is not very large, it further narrows the performance differences among these settings.

\vspace{-0.1in}
\minisection{Effects with Different Encoder-decoder Backbones}
In the original implementation, our VLLVE framework is implemented with the simple UNet as the encoder-decoder backbone. In this section, we extend our analysis by employing more complex backbones, such as MIRNet~\cite{zamir2020learning} (a typical CNN model), Restormer~\cite{zamir2022restormer} (a representative transformer model), and SNR~\cite{xu2022snr} (the mixture of CNN and transformer model). We substitute the encoder–decoder architecture in Fig.~\ref{fig:vllve++} with the corresponding modules from these networks.
The results are shown in \Cref{comparison-abla-supp}, labeled as ``with MIRNet'', ``with Restormer'', and ``with SNR'', respectively. Notably, their performance surpasses that of our original UNet-based setting.
This ablation aims to demonstrate that while our method achieves SOTA performance with simple networks, its results can be further enhanced using stronger encoder and decoders.

\begin{table}[tb!]
	\centering
	\caption{We conduct ablation studies on SDSD, SMID, and DID for VLLVE++, which is built upon VLLVE by incorporating residual term modeling in the decomposition (R.M.) and employing a correspondence refinement strategy (C.R.). VLLVE++ = VLLVE+R.M.+C.R.} 
    \vspace{-0.1in}
	\label{comparison-abla-++}
    \renewcommand{\arraystretch}{1.06}
	\resizebox{1.0\linewidth}{!}
    {
		\begin{tabular}{|c|cc|cc|cc|cc|}
            \hline
			& \multicolumn{2}{c|}{SDSD-indoor} &\multicolumn{2}{c|}{SDSD-outdoor}& \multicolumn{2}{c|}{SMID} &\multicolumn{2}{c|}{DID}  \\ \hline
			Methods & PSNR & SSIM& PSNR & SSIM & PSNR & SSIM& PSNR & SSIM\\
			\hline \hline
			VLLVE& 28.93& 0.88 &26.32 & 0.82& 29.60 &0.82 & 30.10&0.93 \\
            VLLVE+R.M. & 29.41& 0.88 & 26.85 & 0.83 & 30.17 & 0.82 & 30.56 & 0.94 \\
            VLLVE+C.R. & 29.52 & 0.89 & 27.06 & 0.84& 30.25 & 0.83 & 30.71 & 0.93 \\
			\hline   
            VLLVE+R.M.+C.R.& \textbf{29.78}&\textbf{0.90} &\textbf{27.47} & \textbf{0.85}& \textbf{30.71} &\textbf{0.84} & \textbf{31.06}&\textbf{0.95} \\
            \hline
	\end{tabular}}
\end{table}

\begin{figure*}[t]
	\begin{center} 
		\includegraphics[width=1.0\linewidth]{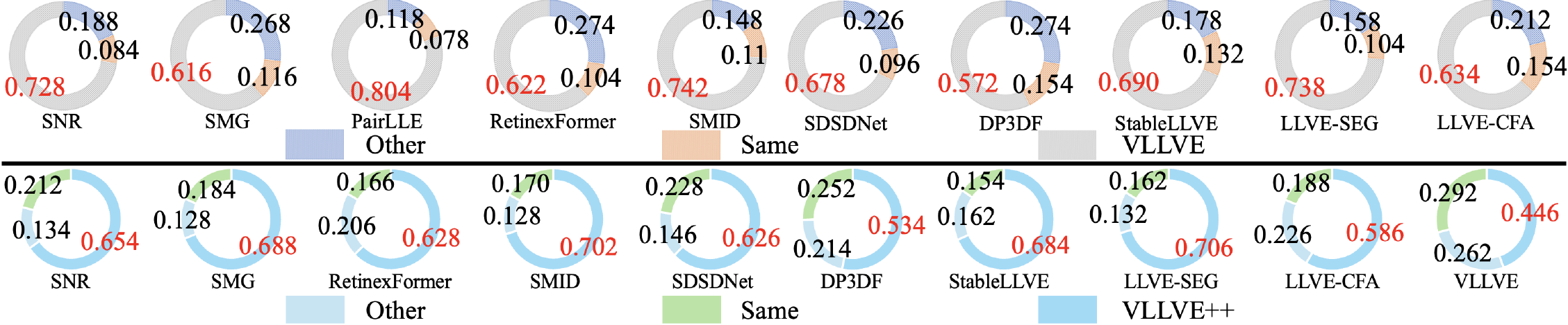}
	\end{center}
    \vspace{-0.1in}
	\caption{
		The above pie charts summarize the results of our user study, and ours is preferred by participants.
	}
	\label{us_tbl}
\end{figure*}

\subsubsection{More Ablation Studies for Components in VLLVE++}

\vspace{-0.1in}
\minisection{Effects of New Decomposition with Residual Terms}
Compared to the decomposition strategy in VLLVE, VLLVE++ introduces a residual term to account for minor degradations that cannot be well approximated by the product of two terms. The residual term prediction shares the same backbone as the view-independent and view-dependent branches.
To evaluate the impact of this modification, we conduct an ablation study comparing the original VLLVE decomposition with our improved strategy (VLLVE+R.M.). As shown in \Cref{comparison-abla-++}, upgrading the decomposition strategy alone (proposed in VLLVE++) still leads to a significant performance improvement compared with VLLVE, validating the effectiveness of our proposed new decomposition strategy.

\vspace{-0.1in}
\minisection{Effects of Bidirection Refinement with Correspondences}
In addition to the new decomposition strategy, we further refine the correspondences and leverage the updated correspondences to improve decomposition accuracy. This creates a positive feedback loop, addressing the limitation where pre-trained correspondence estimation networks may underperform on target scenes in the dataset (particularly since these scenes were not part of the networks' training data).
Although our framework exhibits strong robustness to correspondence errors (analyzed in Sec.~\ref{sec:robustness}), more accurate correspondences generally lead to better decomposition and improved final performance. As shown in \Cref{comparison-abla-++}, even the bidirectional refinement alone (VLLVE vs. VLLVE+C.R.) yields noticeable gains. When combined with our new decomposition strategy, VLLVE++ achieves further improvements, demonstrating that two components are complementary rather than contradictory.

\vspace{-0.1in}
\subsection{User Study}
To validate the effectiveness of our framework through human subjective evaluation, we conducted a large-scale user study with 100 participants of diverse ages, educational backgrounds, and a balanced gender distribution.
For the user study, we randomly selected 30 in-the-wild videos captured with a Sony A7R3 and an iPhone 13. All models included in the study were trained on DID, a large-scale dataset covering diverse scenes and devices. Following common practice in low-light enhancement research~\cite{xu2023low,wang2023lighting}, we adopted the AB-test protocol. For each video, the output of our method was labeled as ``Video A'' and that of a baseline method as ``Video B''. 
For each baseline comparison, five videos were randomly sampled for evaluation.
During the study, participants simultaneously viewed Videos A and B, with their left–right positions randomized to avoid positional bias. They were asked to choose: Video A is better, Video B is better, or Same, based on criteria including natural brightness, contrast, color fidelity, detail preservation, artifact reduction, and temporal consistency.
Each participant evaluated 10 methods $\times$ 5 videos $=50$ comparisons, taking on average about 50 minutes to complete.

\Cref{us_tbl} summarizes the user study's results on VLLVE, and we can see that ours gets more selections from participants over all the baselines. This demonstrates that our method's results are more preferred by the human subjective perception. 

Furthermore, \Cref{us_tbl} presents the user study results comparing VLLVE++ with baseline methods, including VLLVE. The results demonstrate that VLLVE++ is consistently preferred over the baselines, confirming its effectiveness in subjective assessment.

\subsection{The Effects of CFIM Structure}
As discussed earlier, CFIM offers two key benefits: it improves the consistency of image enhancement and enhances backbone robustness by supplying the cross-attention mechanism with information from different time steps.
To isolate the effect of CFIM, we establish baselines by removing the decomposition component from our strategy. The baseline model processes a single frame, while the comparative setup integrates CFIM, applying cross-attention between randomly selected frame pairs during training.
As shown in Table~\ref{comparison-abla-supp-cfim}, models with CFIM substantially outperform those without it, demonstrating the effectiveness of the proposed approach.
Besides, the ablation results of ``w/o C.F.'' in Sec.~\ref{sec:ablation-study} also support the effects of CFIM in the decomposition.

\begin{table}[tb!]
    \scriptsize
	\centering
	\huge
	\caption{The experiments to show the effects of CFIM. ``R.F.'' denotes RetinexFormer.} 
    \label{comparison-abla-supp-cfim}
    \vspace{-0.1in}
    \renewcommand{\arraystretch}{1.06}
    \resizebox{1.0\linewidth}{!}
    {
        \begin{tabular}{|l|cc|cc|cc|cc|}
            \hline
            & \multicolumn{2}{c|}{SDSD-indoor} &\multicolumn{2}{c|}{SDSD-outdoor}& \multicolumn{2}{c|}{SMID} &\multicolumn{2}{c|}{DID}  \\
            \hline
            Methods & PSNR & SSIM& PSNR & SSIM & PSNR & SSIM& PSNR & SSIM\\
            \hline \hline
            SNR &27.30 &0.84 &23.23 & 0.82& 28.49 &0.81 &24.85 &0.90\\
            SNR+CFIM &\textbf{28.15} &\textbf{0.86} &\textbf{24.57} &\textbf{0.83} &\textbf{29.24} &\textbf{0.82} &\textbf{26.03} &\textbf{0.91}\\ \hline
            R.F.& 26.56& 0.79 &22.80 &0.77 &29.15 &0.82 & 25.40&0.89\\
            R.F.+CFIM &\textbf{27.42} &\textbf{0.80} &\textbf{23.75} &\textbf{0.79} &\textbf{29.87} &\textbf{0.83} &\textbf{26.38} &\textbf{0.90}\\ 
            \hline
        \end{tabular}
        }
\end{table}

\vspace{-0.1in}
\subsection{Robustness Towards Correspondences}
\label{sec:robustness}
We have shown that our method is robust to variations in the choice of correspondence computation networks. Here, we further investigate the resilience to incorrect or incomplete correspondences.

In the first experiment, we introduce random noise to the computed correspondences by perturbing their location values. As shown in Table~\ref{comparison-abla-supp-robust}, the perturbed models (``VLLVE with per.'' and ``VLLVE++ with per.'') exhibit a slight performance drop compared to the unperturbed case. Nonetheless, they still outperform most SOTA baselines, demonstrating that our framework is robust to a certain level of erroneous correspondences, which commonly occur in regions with complex textures or other challenging scenarios.

In the second experiment, we examine the impact of insufficient correspondences, such as those occurring in occluded regions. To simulate this, we randomly reduce the number of correspondences to 10\% of the original set. As shown in Table~\ref{comparison-abla-supp-robust}, the models with reduced correspondences (``VLLVE with reduce'' and ``VLLVE++ with reduce'') still outperform the baselines, though with a smaller margin. This indicates that even a limited set of correspondences provides valuable information that aligns with other constraints, further demonstrating the robustness of our framework.

\begin{table}[tb!]
	\centering
	\caption{The experiments to show the robustness of our framework towards correspondences.} 
    \label{comparison-abla-supp-robust}
    \vspace{-0.1in}
    \renewcommand{\arraystretch}{1.06}
    \resizebox{1.0\linewidth}{!}
    {
        \begin{tabular}{|l|cc|cc|cc|cc|}
            \hline
            & \multicolumn{2}{c|}{SDSD-indoor} &\multicolumn{2}{c|}{SDSD-outdoor}& \multicolumn{2}{c|}{SMID} &\multicolumn{2}{c|}{DID}  \\
            \hline
            Methods & PSNR & SSIM& PSNR & SSIM & PSNR & SSIM& PSNR & SSIM\\
            \hline
            VLLVE with reduce &27.97 &0.84 &25.73 & 0.80&29.12 &0.79 & 28.86&0.89\\
            VLLVE with per. &28.10 &0.86 &26.04 &0.79 &29.37 &0.81 &29.08 &0.91\\
            VLLVE      & 28.93&0.88 &26.32 & 0.82& 29.60 &0.82 & 30.10&0.93 \\
            \hline
            \hline 
            VLLVE++ with reduce &28.94 &0.87 &26.80 &0.81 &29.88 &0.80 &29.95 &0.91\\
            VLLVE++ with per. &29.05 &0.88 &27.06 &0.83 &30.14 &0.82 &30.21 &0.93\\
            VLLVE++ & 29.78&0.90 &27.47 & 0.85& 30.71 &0.84 & 31.06&0.95 \\
            \hline
        \end{tabular}
        }
\end{table}

\begin{table}[t]
    \scriptsize
	\centering
	\caption{
		Quantitative comparison on the SMID and SDSD datasets involving degradations of both low light and low resolution.
        ``F'' denotes FastDVDnet~\cite{tassano2020fastdvdnet} and ``T'' means StableLLVE~\cite{zhang2021learning}.
	}
	\label{comparison1}
    \vspace{-0.1in}
    \renewcommand{\arraystretch}{1.06}
    {
		\begin{tabular}{|l|cc|cc|cc|}
            \hline
			& \multicolumn{2}{c|}{SMID} &\multicolumn{2}{c|}{SDSD Indoor}&\multicolumn{2}{c|}{SDSD Outdoor}\\
			\hline \hline
			Methods & PSNR & SSIM& PSNR & SSIM& PSNR & SSIM \\
			F+T+Zooming&23.97&0.74&26.54&0.81&24.41&0.73 \\
			F+T+TGA &24.57 &\textbf{0.76}&25.31&0.78&25.01&0.75 \\
			F+T+TDAN &24.00 &0.70&25.89&0.87&23.81&0.71 \\
			T+F+Zooming&23.73 &0.70&26.01&0.79&23.69&0.74 \\
			T+F+TGA &24.66 &0.68&24.70&0.77&24.88&0.72 \\
			T+F+TDAN &24.21 &0.71&24.88&0.81&23.35&0.70 \\
			\hline
			BasicVSR~\cite{chan2021basicvsr} &21.78&0.62&20.72&0.71&20.91&0.70\\
			BasicVSR++~\cite{chan2021basicvsr++}&22.48&0.65&21.02&0.75&21.31&0.72\\
			IconVSR~\cite{chan2021basicvsr} &21.99&0.63&20.94&0.73&20.89&0.71\\
			Zooming~\cite{xiang2020zooming}&24.89 &0.71&26.32&0.84&22.05&0.72 \\
			TGA~\cite{isobe2020video} &23.40 &0.67&23.92&0.76&23.83&0.74 \\
			TDAN~\cite{tian2020tdan}&24.65 &0.70&24.00&0.80&22.57&0.74 \\
			EDVR~\cite{wang2019edvr} & 24.50&0.70&25.00&0.83&23.37&0.75 \\
			DP3DF~\cite{xu2023deep} &25.73 &0.73&27.11&0.85&25.80&0.77  \\
			\hline 
            VLLVE~\cite{xu2025low}&\underline{26.85} &\underline{0.74} &\underline{29.41}&\underline{0.87}&\underline{27.18}&\underline{0.79}  \\
            VLLVE++ &\textbf{27.71} &\textbf{0.76} &\textbf{30.25} &\textbf{0.88} &\textbf{28.13} &\textbf{0.80} \\
            \hline
	\end{tabular}}
\end{table}

\section{VLLVE++ in More Challenging Scenarios}

\subsection{Comparison with More Degradations}
\label{sec:more-deg}

We also evaluate our framework under more severe degradations in dark environments.
Following~\cite{xu2023deep}, we conduct the experiments of mapping a low-resolution, low-light, and noisy video to a high-resolution, normal-light, and noise-free video.
The experimental settings are adopted as the same as them in~\cite{xu2023deep}. The baselines contain SOTA video SR methods, including BasicVSR~\cite{chan2021basicvsr}, IconVSR~\cite{chan2021basicvsr}, BasicVSR++~\cite{chan2021basicvsr++}, Zooming~\cite{xiang2020zooming}, TGA~\cite{isobe2020video}, TDAN~\cite{tian2020tdan}, EDVR~\cite{wang2019edvr}, and DP3DF~\cite{xu2023deep}.
Moreover, the combined approaches are also considered as the competitive methods by involving the video denoising method FastDVDnet~\cite{tassano2020fastdvdnet} and video illumination enhancement StableLLVE~\cite{zhang2021learning}.
To achieve SR, we add two extra up-sampling layers for decoder branches.

The results are presented in \Cref{comparison1}, demonstrating that our method VLLVE and VLLVE++ consistently achieves superior PSNR and SSIM values compared to baselines. 
Moreover, our proposed VLLVE++ consistently outperforms VLLVE even in this more challenging scenario with additional degradations. This superior performance further validates the effectiveness of (1) our novel decomposition strategy and (2) the bidirectional refinement mechanism for correspondences and decomposition results.

\begin{table}[tb!]
    \scriptsize
	\centering
	\caption{Quantitative comparison on YouTube.}
	\label{comparison2-youtube}
    \vspace{-0.1in}
    \renewcommand{\arraystretch}{1.06}
	\resizebox{1.02\linewidth}{!}
    {
		\begin{tabular}{|l|cccccc|}
            \hline
            &SNR&SMG&PairLLE&RetinexFormer&SMOID&SMID\\
			\hline
			PSNR &22.65 &21.82 & 20.19&23.14 & 20.83&20.60\\
			SSIM&0.758 & 0.719&0.661&0.801 &0.689 & 0.694\\
			\hline \hline
            &SDSDNet&DP3DF&StableLLVE&LLVE-SEG&VLLVE&VLLVE++\\
			\hline
			PSNR  & 21.11&22.96 &21.57 &22.54 &\underline{24.83} &\textbf{25.75}\\
			SSIM  &0.767 & 0.784&0.753 &0.746 &\underline{0.835} &\textbf{0.842}\\
            \hline
	\end{tabular}}
\end{table}

\begin{table}[tb!]
    \large
	\centering
	\caption{Quantitative comparison on Real-world BVI-RLV.}
	\label{comparison2-BVI-RLV}
    \vspace{-0.1in}
    \renewcommand{\arraystretch}{1.06}
	\resizebox{1.0\linewidth}{!}
    {
		\begin{tabular}{|l|ccccccc|}
            \hline
            &SMOID&SDSDNet&PCDUNet&STA-SUNet&BVI-CDM&VLLVE&VLLVE++\\
			\hline
			PSNR &17.98 &18.05&19.50 &20.64&22.20 &\underline{23.61} &\textbf{24.47} \\
			SSIM&0.621 &0.690 &0.757 &0.765& 0.773&\underline{0.786} & \textbf{0.794}\\
            \hline
	\end{tabular}}
\end{table}

\subsection{Experiments on More Real-world LLVE Datasets}
\label{sec:youtube}

\begin{figure*}[t]
	\centering
	\begin{subfigure}[c]{0.16\textwidth}
		\centering
		\includegraphics[width=1.15in]{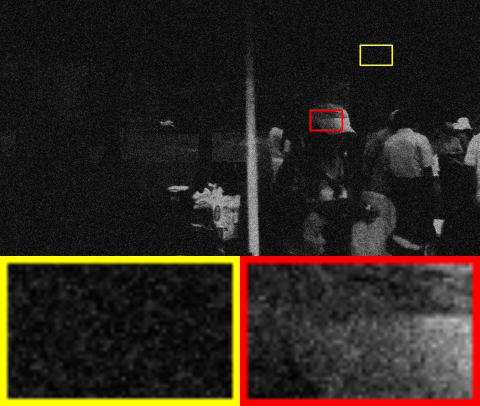}
		\caption*{Input}
	\end{subfigure}
	\begin{subfigure}[c]{0.16\textwidth}
		\centering
		\includegraphics[width=1.15in]{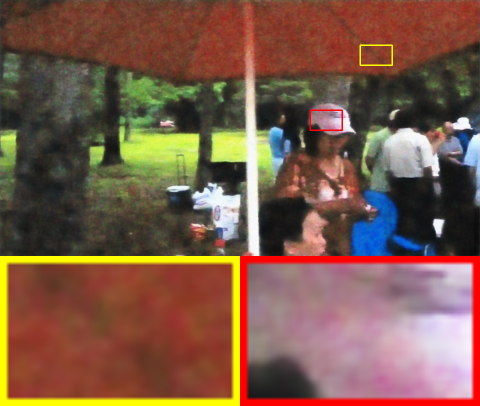}
		\caption*{RetinexFormer}
	\end{subfigure}
	\begin{subfigure}[c]{0.16\textwidth}
		\centering
		\includegraphics[width=1.15in]{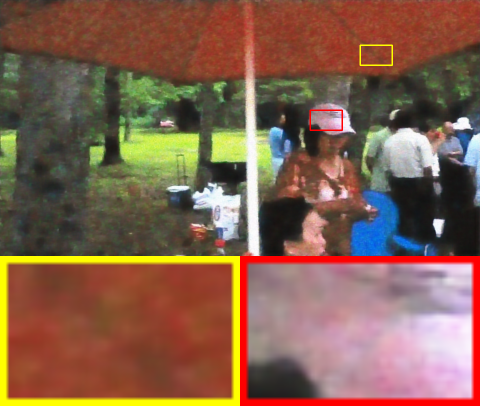}
		\caption*{DP3DF}
	\end{subfigure}
    	\begin{subfigure}[c]{0.16\textwidth}
		\centering
		\includegraphics[width=1.15in]{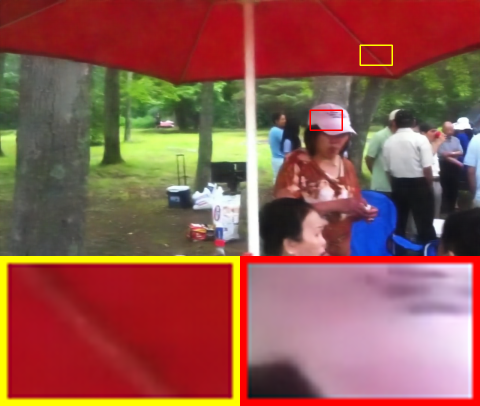}
		\caption*{VLLVE}
	\end{subfigure}
	\begin{subfigure}[c]{0.16\textwidth}
		\centering
		\includegraphics[width=1.15in]{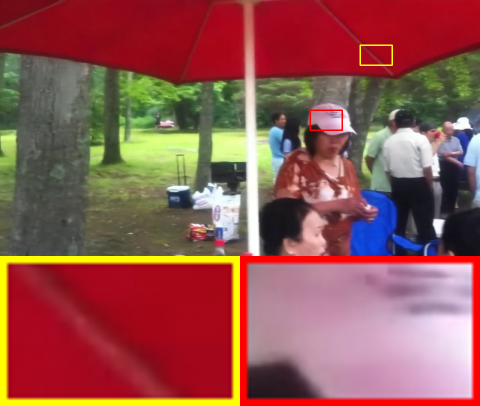}
		\caption*{VLLVE++}
	\end{subfigure} 
	\begin{subfigure}[c]{0.16\textwidth}
		\centering
		\includegraphics[width=1.15in]{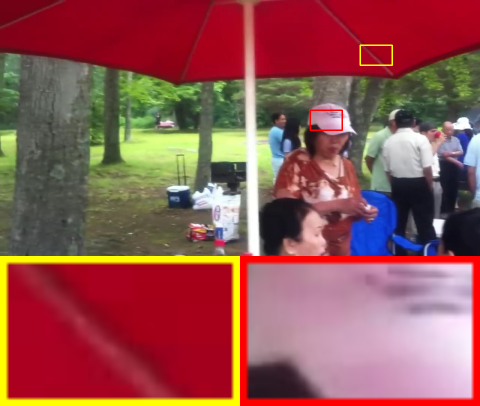}
		\caption*{GT}
	\end{subfigure}

	\begin{subfigure}[c]{0.16\textwidth}
		\centering
		\includegraphics[width=1.15in]{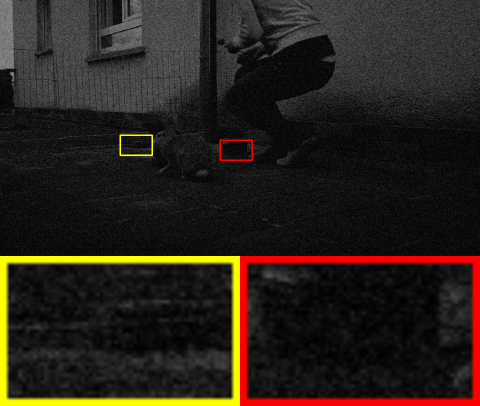}
		\caption*{Input}
	\end{subfigure}
	\begin{subfigure}[c]{0.16\textwidth}
		\centering
		\includegraphics[width=1.15in]{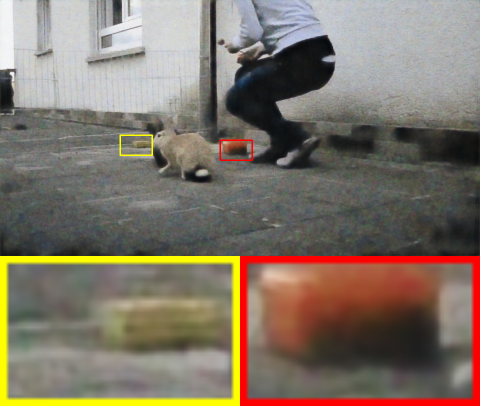}
		\caption*{RetinexFormer}
	\end{subfigure}
	\begin{subfigure}[c]{0.16\textwidth}
		\centering
		\includegraphics[width=1.15in]{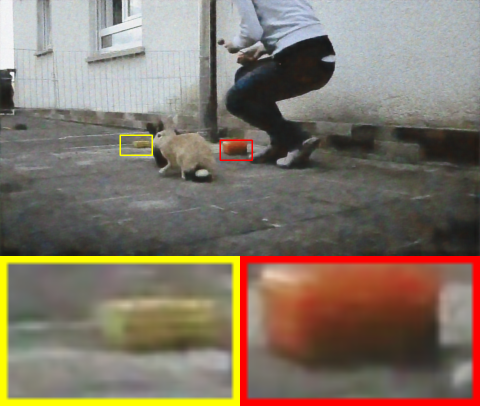}
		\caption*{DP3DF}
	\end{subfigure}
    	\begin{subfigure}[c]{0.16\textwidth}
		\centering
		\includegraphics[width=1.15in]{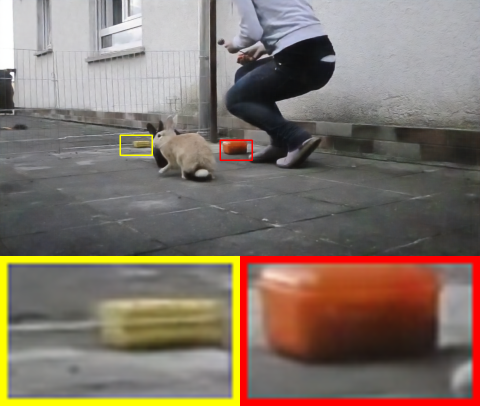}
		\caption*{VLLVE}
	\end{subfigure}
	\begin{subfigure}[c]{0.16\textwidth}
		\centering
		\includegraphics[width=1.15in]{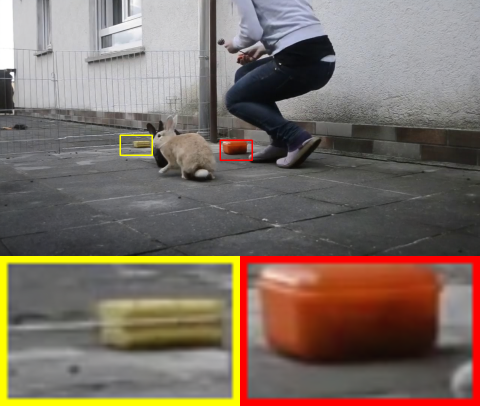}
		\caption*{VLLVE++}
	\end{subfigure} 
	\begin{subfigure}[c]{0.16\textwidth}
		\centering
		\includegraphics[width=1.15in]{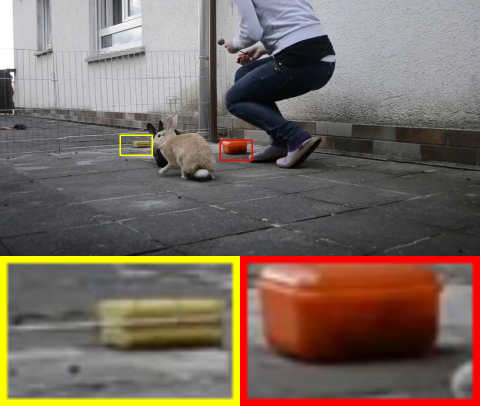}
		\caption*{GT}
	\end{subfigure}  
	
	\caption{Visual comparisons on YOUTUBE. Our method has less noise, better visibility, and richer details.}
	\label{fig:cmp-youtube1}
	
\end{figure*}

Compared to DAVIS, YouTube-VOS~\cite{xu2018youtube} is substantially larger and contains more dynamic videos. To evaluate our framework under these more challenging conditions, we conducted experiments on YouTube-VOS, applying the same synthetic degradation strategy used for DAVIS. The results, presented in \Cref{comparison2-youtube}, demonstrate that our method (VLLVE and VLLVE++) consistently achieves the best performance. Notably, VLLVE++ further outperforms VLLVE in this setting. Visual comparisons are provided in \Cref{fig:cmp-youtube1}.

We further evaluate our framework on the recently released real-world dataset BVI-RLV~\cite{anantrasirichai2024bvi,lin2024bvi}, which contains 40 low-light video scenes with diverse motion patterns under two lighting conditions. As shown in Table~\ref{comparison2-BVI-RLV}, we compare our method with three additional SOTA approaches (PCDUNet~\cite{lin2024bvi}, STA-SUNet~\cite{lin2024spatio}, and BVI-CDM~\cite{lin2024bvi}). The results consistently highlight the superiority of VLLVE and VLLVE++, and the strong performance of VLLVE++ further confirms the effectiveness of our proposed strategies for real-world applications.

\begin{table}[tb!]
    \large
    \centering
    \caption{Quantitative comparison on 3D low-light datasets with normal-light pairs, proposed in~\cite{wang2023lighting}. ``R.F.'' denotes RetinexFormer, and ``+N.'' means conducting NeRF training on the enhanced images.}
    \label{comparison-3d}
    \vspace{-0.1in}
    \renewcommand{\arraystretch}{1.06}
    \resizebox{1\linewidth}{!}
    {
        \begin{tabular}{|l|cccccc|}
            \hline
            &PairLLE&SNR&R.F.& LLNeRF& VLLVE & VLLVE++\\
            \hline
            PSNR &18.35 & 18.87 &19.49 &20.50 &\underline{21.09} &\textbf{21.58}\\
            SSIM&0.604  &0.650 &0.702 & 0.758& \underline{0.763} & \textbf{0.775}\\
            \hline \hline
            &PairLLE+N.&SNR+N.&R.F.+N.& SCI+N. & VLLVE+N. & VLLVE++ + N.\\
            \hline
            PSNR &18.72 & 19.24 &19.98 & 13.08 &\underline{22.35} &\textbf{23.51}\\
            SSIM&0.681 &0.693 &0.736 &0.505 & \underline{0.782} &\textbf{0.794}\\
            \hline
        \end{tabular}}
\end{table}

\begin{table}[tb!]
    \scriptsize
	\centering
	\caption{Quantitative comparison on the RAW dataset, CRVD.}
	\label{comparison22}
    \vspace{-0.1in}
    \renewcommand{\arraystretch}{1.06}
	\resizebox{1.0\linewidth}{!}
    {
        \begin{tabular}{|l|ccccc|}
            \hline
			CRVD&SNR&SDSDNet&DP3DF&VLLVE&VLLVE++\\
			\hline \hline
			PSNR &29.69  &29.86  &30.05 &\underline{30.28} &\textbf{31.03} \\
			SSIM& 0.825 &0.830   &0.828 &\underline{0.837} &\textbf{0.842} \\
            \hline
	\end{tabular}}
\end{table}

\vspace{-0.1in}
\subsection{Enhancement for NeRF}
\label{sec:nerf}
Our framework's enhanced results (including both VLLVE and VLLVE++) are beneficial for NeRF reconstruction, as we have implemented a decomposition that maintains consistency across different views.
To evaluate the impact of our enhanced results on NeRF, we conducted experiments on the multi-view low-light dataset from \cite{wang2023lighting}. We used the network trained on DID, which leverages a large-scale training set and thus offers strong generalization. Moreover, DID naturally includes multi-view images. For each view, we selected its nearest neighboring view as the reference input. The experimental settings follow those in \cite{wang2023lighting}.

In \Cref{comparison-3d}, we present the PSNR and SSIM results on the testing set of \cite{wang2023lighting}, where ground truth data is available. 
It's important to note that none of the low-light enhancement networks were fine-tuned on the testing set besides the NeRF training.
The superiority of VLLVE and VLLVE++ is evident.
Meanwhile, VLLVE++ achieves superior performance, attributable to both enhanced correspondence accuracy in our new decomposition.

\subsection{Experiments for RAW data}
\label{sec:raw}

We conduct a comparison for the RAW LLVE task, which involves mapping low-light and noisy RAW frames to normal-light and noise-free sRGB frames. 
In this task, our network processes a video clip in the RAW format as input. 
We employ one dynamic RAW video dataset for experiments: the CRVD dataset~\cite{yue2020supervised} featuring dynamic RAW video data (low-light inputs were synthesized by introducing invisibility and noise to the RAW data). 
The results are presented in \Cref{comparison22}, demonstrating that our approach (VLLVE and VLLVE++) continues to deliver SOTA performance on the RAW video benchmark.

\section{Conclusion}
We propose a novel low-light video enhancement framework, VLLVE, which introduces a spatiotemporal decomposition into view-independent and view-dependent components. To guide the view-independent component, we leverage correspondence priors across frames to enforce consistency, while a temporal continuity constraint is applied to the view-dependent component.
In addition, we design a dual-network architecture for mutual feature propagation, further strengthening the consistency of the decomposition. 

To further improve enhancement performance, we introduce VLLVE++, an enhanced version of VLLVE. It extends the original framework by adopting a more comprehensive decomposition strategy and integrating cooperative learning between decomposition and correspondence refinement. The decomposition is refined through the addition of a residual term guided by a spatial continuity constraint and a more accurate reconstruction target. Correspondence refinement is performed in a self-supervised manner. VLLVE++ achieves superior performance in both quantitative and qualitative evaluations, and establishes SOTA results on more challenging scenes and tasks, as demonstrated by our experiments.

\section*{Acknowledgments}
This work is supported by 1) the Zhejiang Province Science Foundation under grant No. LD24F020002; 2) the Key Project of the National Natural Science Foundation of China under grant No. 62536007; 3) the Zhejiang Province's 2025 ``Leading Goose + X'' Science and Technology Plan under grant No. 2025C02034.



\ifCLASSOPTIONcaptionsoff
\newpage
\fi



%


\bibliographystyle{IEEEtran}
\bibliography{egbib}

%

\vspace{-.4in}
\begin{IEEEbiography}
    [{\includegraphics[height=1.15in,clip,keepaspectratio]{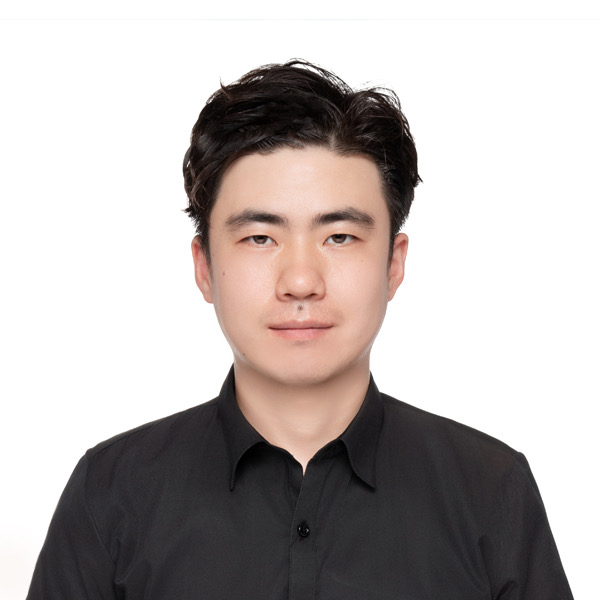}}]
    {Xiaogang Xu}
    is a postdoc research fellow in the Chinese University of Hong Kong. He received his Ph.D. degree from CUHK in 2022 and bachelor degree from Zhejiang University in 2018. In 2023, he is a research scientist in Zhejiang Lab and meanwhile a ZJU100 Young Professor at ZJU.
	He obtained the Hong Kong PhD Fellowship in 2018. He serves as a reviewer for CVPR, ICCV, ECCV, Neurips, ICLR, TPAMI, TIP, IJCV, etc. His research interest includes deep learning, computational photography, AIGC, large models.
\end{IEEEbiography}

\vspace{-.4in}
\begin{IEEEbiography}
	[{\includegraphics[height=1.05in,clip,keepaspectratio]{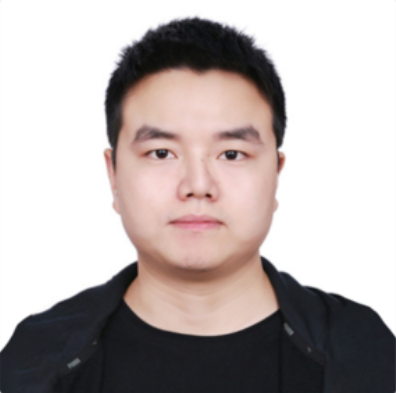}}]
	{Kun Zhou} is currently an assistant professor at Shenzhen University. He received the B.Eng. degree in Information and Computing Science from Xiamen University of Technology, Xiamen, China, in 2013; the M.S. degree in Faculty of Electrical Engineering and Computer Science from Ningbo University, Ningbo, China, in 2017; the Ph.D. degree from the Chinese University of Hong Kong (ShenZhen) in 2024. 
    His research interests include 3D computer vision, image$\&$video restoration, and video frame interpolation/extrapolation. He also serves as a reviewer for IJCV, TIP, CVPR, ICCV, NeurIPS, etc.
\end{IEEEbiography}

\vspace{-.4in}
\begin{IEEEbiography}
	[{\includegraphics[height=1.25in,clip,keepaspectratio]{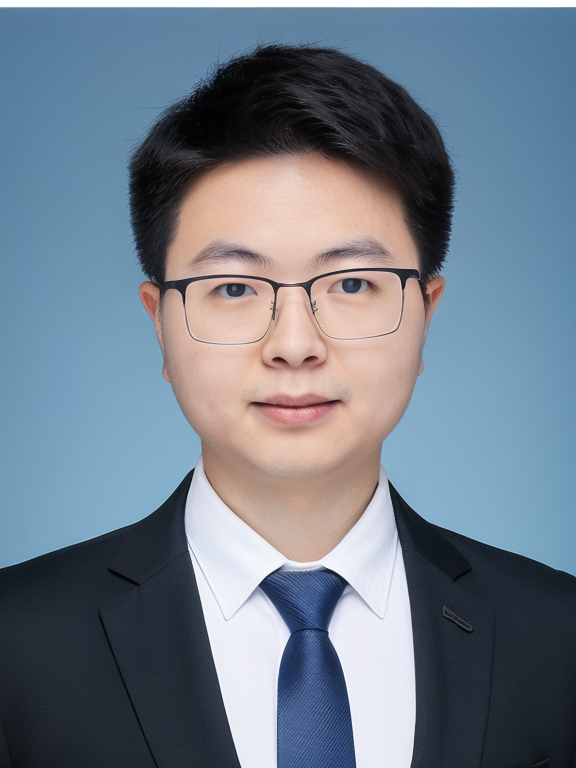}}]
	{Tao Hu} is currently a research scientist at PICO, Bytedance. Previously, he was a postdoctoral researcher at the National University of Singapore.  He received his Ph.D. in Computer Science from The Chinese University of Hong Kong. His general research interests cover the broad area of computer vision, with special emphasis on 3D representation and generation. His research has been published in top-tier conferences including CVPR, ICCV, ECCV, and NeurIPS. Notably, his paper received a Best Paper Award nomination at ICCV 2023.
\end{IEEEbiography}

\vspace{-.4in}
\begin{IEEEbiography}
	[{\includegraphics[height=1.25in,clip,keepaspectratio]{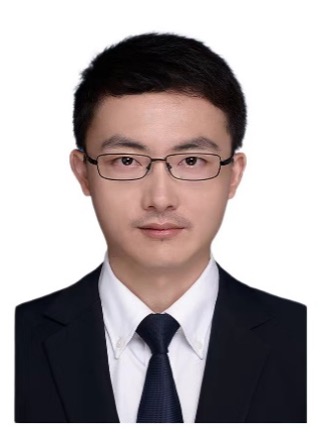}}]
	{Jiafei Wu}
	received the B.S. degree from JXUFE in 2010, the M.S. degree and Ph.D. degree from the University of Hong Kong in 2012 and 2017, respectively. He has been a senior engineer, manager and deputy director from 2018 to 2023 in SenseTime. He is currently with the Zhejiang Lab. His research interests include deep learning, trustworthy AI, embedded system, and computational intelligence.
\end{IEEEbiography}

\vspace{-.4in}
\begin{IEEEbiography}
	[{\includegraphics[height=1.15in,clip,keepaspectratio]{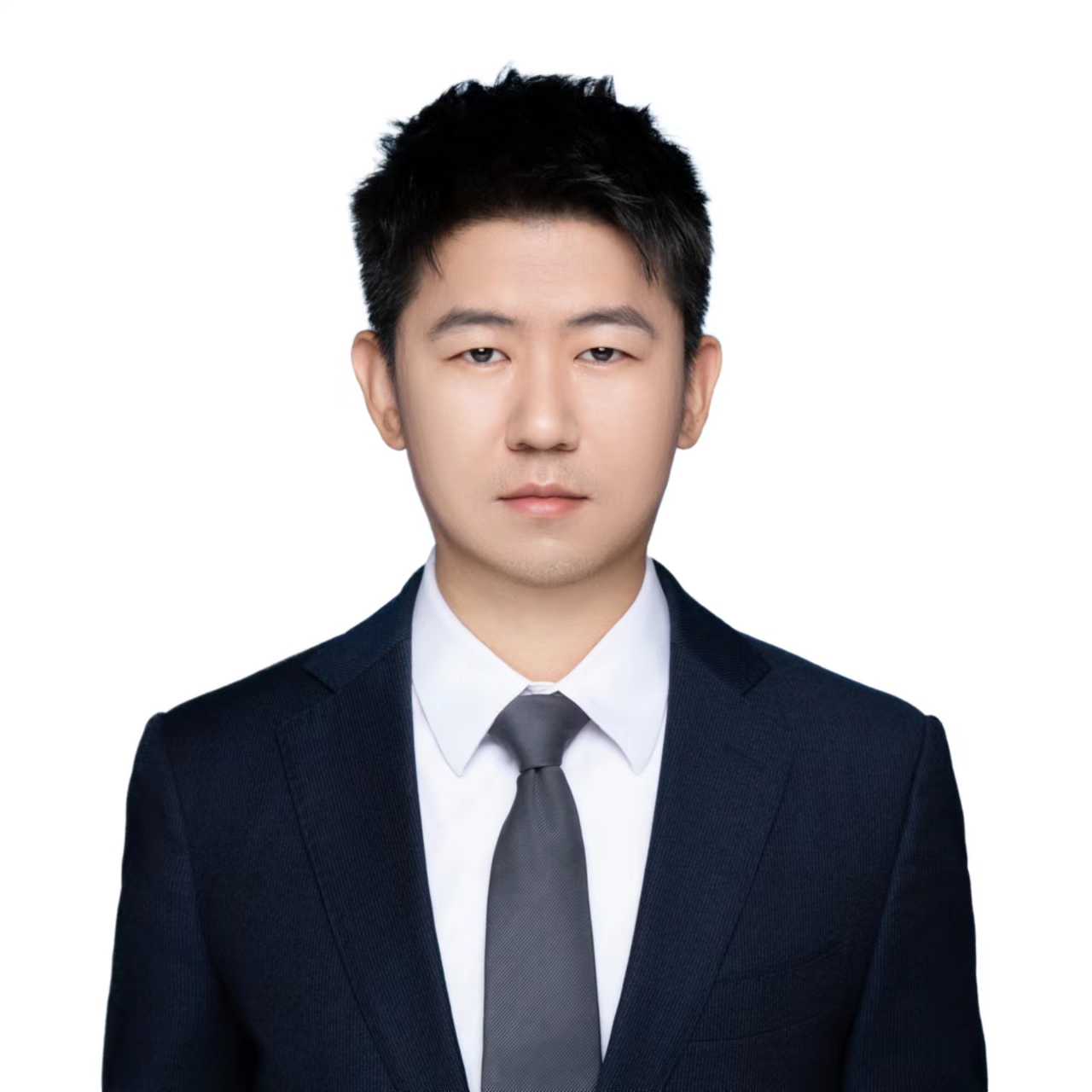}}]
	{Ruixing Wang} is currently at camera group of DJI. He received the B.S. degree from Huazhong University of Science and Technology in 2016, and the Ph. D. degree from the Chinese University of Hong Kong, in 2021. Before joining in DJI, he was a principal engineer in Honor Device Co., Ltd. He serves as a reviewer for CVPR, ICCV, ECCV, Neurips, ICML, ICLR, AAAI, WACV, ACCV, TPAMI, IJCV, etc. His research interests include computational photography and image processing. 
\end{IEEEbiography}

\vspace{-.4in}
\begin{IEEEbiography}
	[{\includegraphics[height=1.25in,clip,keepaspectratio]{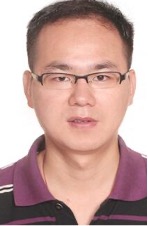}}]
	{Hao Peng}
	is currently a professor in the School of Computer Science and Technology at Zhejiang Normal University. He received his Ph.D degree in communication and information system from Shanghai Jiaotong University. He serves as vice dean of the School of Computer Science and Technology, executive dean of AI Research Academy, Director of Network Security and Optimization Research Institute. He is sponsored by Special Support Program for High-level Talents in Zhejiang Province and National Natural Science Foundation of China. His current research interests include Network Security, AI Security and Data-driven Security. He is a member of IEEE, CAAI and CCF.
\end{IEEEbiography}

\vspace{-.4in}
\begin{IEEEbiography}
    [{\includegraphics[height=1.26in,clip,keepaspectratio]{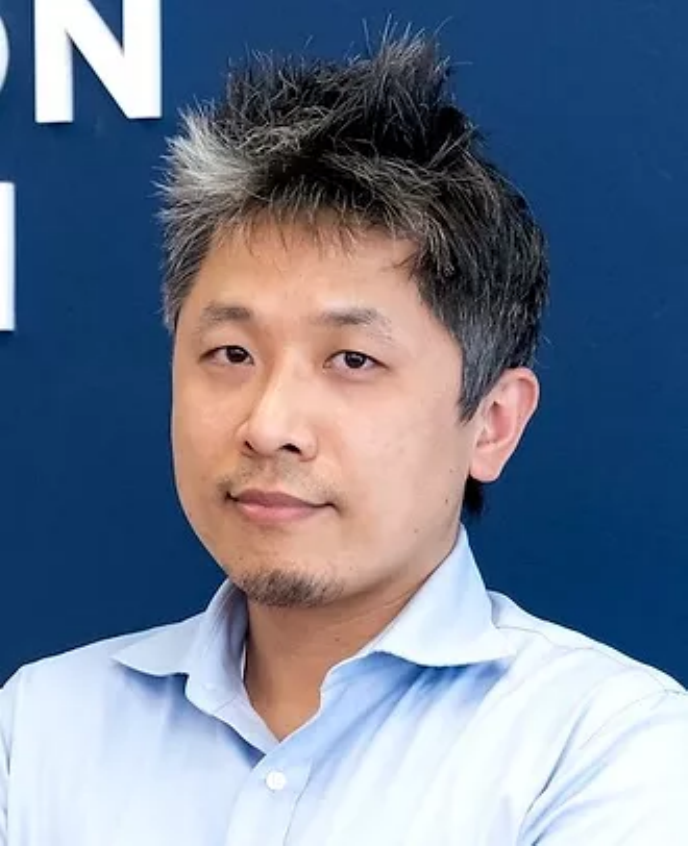}}]
    {Bei Yu}
    (M'15-SM'22)
    received the Ph.D.~degree from The University of Texas at Austin in 2014.
    He is currently a Professor in the Department of Computer Science and Engineering, The Chinese University of Hong Kong.
    He has served as TPC Chair of ACM/IEEE Workshop on Machine Learning for CAD, and in many journal editorial boards and conference committees.
    He received eleven Best Paper Awards from ICCAD 2024 \& 2021 \& 2013,
    IEEE TSM 2022, DATE 2022, ASPDAC 2021 \& 2012, ICTAI 2019, Integration, the VLSI Journal in 2018,
    ISPD 2017, SPIE Advanced Lithography Conference 2016, and six ICCAD/ISPD contest awards.
\end{IEEEbiography}

\end{document}